\documentclass{article}

\usepackage{fullpage}
\usepackage{times}

\usepackage{amsfonts}
\usepackage{amsmath} % assumes amsmath package installed
\usepackage{amssymb}  % assumes amsmath package installed
\usepackage{amsthm} %
\usepackage{thmtools}

\usepackage{graphicx}
\usepackage{graphics} % for pdf, bitmapped graphics files
\usepackage{url}
\usepackage{rotating}
\usepackage{subfigure}
\usepackage{enumitem}
\usepackage[usenames, dvipsnames]{color}
\usepackage[linesnumbered,ruled,lined]{algorithm2e}
\usepackage{algcompatible}

\def\no{\; {not} \;}

\newcommand{\stt}[1]{{\small\texttt{#1}}}

\newtheorem{execexample}{\bf Execution Example}

\declaretheoremstyle[
  headfont=\normalfont\bfseries,
  numbered=unless unique,
  bodyfont=\normalfont,
  spaceabove=1em plus 0.75em minus 0.25em,
  spacebelow=1em plus 0.75em minus 0.25em,
  qed=$\blacklozenge$,
]{example}

\declaretheorem[
  style=example,
  title=Example
]{example2}

\declaretheoremstyle[
  headfont=\normalfont\bfseries,
  numbered=unless unique,
  bodyfont=\normalfont,
  spaceabove=1em plus 0.75em minus 0.25em,
  spacebelow=1em plus 0.75em minus 0.25em,
  qed=$\blacktriangle$,
]{definition2}

\declaretheorem[
  style=definition2,
  title=Definition
]{defn2}

\declaretheoremstyle[
  headfont=\normalfont\bfseries,
  numbered=unless unique,
  bodyfont=\normalfont,
  spaceabove=1em plus 0.75em minus 0.25em,
  spacebelow=1em plus 0.75em minus 0.25em,
  qed=$\bigstar$,
]{proposition2}

\declaretheorem[
  style=proposition2,
  title=Proposition
]{prop2}

\declaretheoremstyle[
  headfont=\normalfont\bfseries,
  numbered=unless unique,
  bodyfont=\normalfont,
  spaceabove=1em plus 0.75em minus 0.25em,
  spacebelow=1em plus 0.75em minus 0.25em,
  qed=$\blacksquare$,
]{lemma2}

\declaretheorem[
  style=lemma2,
  title=Lemma
]{lemm2}

\newcommand{\rif}{\stackrel{\,\,+}{\leftarrow}}
\newenvironment{s_itemize}{\begin{list}{$\bullet$}
{\setlength{\rightmargin}{0em}
\setlength{\itemsep}{0em}
\setlength{\topsep}{0em}
\setlength{\parsep}{0em}}}{\end{list}}

\newcounter{ctr}

\begin{document}
\title{REBA: A Refinement-Based Architecture for Knowledge
  Representation and Reasoning in Robotics}

\author{
  Mohan Sridharan \\
  School of Computer Science\\
  University of Birmingham, UK\\
  \texttt{m.sridharan@bham.ac.uk}
  \and
  Michael Gelfond \\
  Department of Computer Science \\
  Texas Tech University, USA\\
  \texttt{michael.gelfond@ttu.edu}
  \and
  Shiqi Zhang \\
  Department of Computer Science \\
  SUNY at Binghamton, USA\\
  \texttt{szhang@cs.binghamton.edu}
  \and
  Jeremy Wyatt \\
  School of Computer Science \\
  University of Birmingham, UK \\
  \texttt{jlw@cs.bham.ac.uk}
}

% \texttt{szhang@cs.utexas.edu; m.sridharan@auckland.ac.nz; michael.gelfond@ttu.edu; jlw@cs.bham.ac.uk}

\maketitle

%%%%%-----------------------------------------------------------------------------
%%%%%-----------------------------------------------------------------------------
\begin{abstract}
  This paper describes an architecture for robots that combines the
  complementary strengths of probabilistic graphical models and
  declarative programming to represent and reason with logic-based and
  probabilistic descriptions of uncertainty and domain knowledge. An
  action language is extended to support non-boolean fluents and
  non-deterministic causal laws. This action language is used to
  describe tightly-coupled transition diagrams at two levels of
  granularity, with a fine-resolution transition diagram defined as a
  refinement of a coarse-resolution transition diagram of the domain.
  The coarse-resolution system description, and a history that
  includes (prioritized) defaults, are translated into an Answer Set
  Prolog (ASP) program. For any given goal, inference in the ASP
  program provides a plan of abstract actions. To implement each such
  abstract action, the robot automatically zooms to the part of the
  fine-resolution transition diagram relevant to this action. A
  probabilistic representation of the uncertainty in sensing and
  actuation is then included in this zoomed fine-resolution system
  description, and used to construct a partially observable Markov
  decision process (POMDP). The policy obtained by solving the POMDP
  is invoked repeatedly to implement the abstract action as a sequence
  of concrete actions, with the corresponding observations being
  recorded in the coarse-resolution history and used for subsequent
  reasoning. The architecture is evaluated in simulation and on a
  mobile robot moving objects in an indoor domain, to show that it
  supports reasoning with violation of defaults, noisy observations
  and unreliable actions, in complex domains.
%   on robots, there is a
%   $39\%$ reduction in execution time compared with a purely
%   probabilistic, but still hierarchical, approach.
\end{abstract}

%%%%%-----------------------------------------------------------------------------
%%%%%-----------------------------------------------------------------------------
%\vspace*{-2.25em}
\section{Introduction}
%\vspace*{-0.75em}
\label{sec:intro}
Robots\footnote{We use the terms ``robot'' and ``agent''
  interchangeably in this paper.} are increasingly being used to
assist humans in homes, offices and other complex domains. To truly
assist humans in such domains, robots need to be re-taskable and
robust. We consider a robot to be re-taskable if its reasoning system
enables it to achieve a wide range of goals in a wide range of
environments. We consider a robot to be robust if it is able to cope
with unreliable sensing, unreliable actions, changes in the
environment, and the existence of atypical environments, by
representing and reasoning with different description of knowledge and
uncertainty. While there have been many attempts, satisfying these
desiderata remains an open research problem.

Robotics and artificial intelligence researchers have developed many
approaches for robot reasoning, drawing on ideas from two very
different classes of systems for knowledge representation and
reasoning, based on logic and probability theory respectively. Systems
based on logic incorporate compositionally structured commonsense
knowledge about objects and relations, and support powerful
generalization of reasoning to new situations. Systems based on
probability reason optimally (or near optimally) about the effects of
numerically quantifiable uncertainty in sensing and action. There have
been many attempts to combine the benefits of these two classes of
systems, including work on joint (i.e., logic-based and probabilistic)
representations of state and action, and algorithms for planning and
decision-making in such formalisms.
% for factored/relational Bayes nets and Markov logic networks, and
% work on joint representations of action for various probabilistic
% extensions of PDDL. Approaches for planning and decision making in
% these formalisms include factored and relational Markov decision
% processes (MDPs), their partially observable extensions
% (factored/relational POMDPs), and probabilistic answer set
% programming.
These approaches provide significant expressive power, but they also
impose a significant computational burden. More efficient (and often
approximate) reasoning algorithms for such unified
probabilistic-logical paradigms are being developed. However,
practical robot systems that combine abstract task-level planning with
probabilistic reasoning, link, rather than unify, their logic-based
and probabilistic representations, primarily because roboticists often
need to trade expressivity or correctness guarantees for computational
speed. Information close to the sensorimotor level is often
represented probabilistically to quantitatively model and reason about
the uncertainty in sensing and actuation, with the robot's beliefs
including statements such as ``the robotics book is on the shelf with
probability $0.9$''. At the same time, logic-based systems are used to
reason with (more) abstract commonsense knowledge, which may not
necessarily be natural or easy to represent probabilistically.  This
knowledge may include hierarchically organized information about
object sorts (e.g., a cookbook is a book), and default information
that holds in all but a few exceptional situations (e.g., ``books are
typically found in the library''). These representations are linked,
in that the probabilistic reasoning system will periodically commit
particular claims about the world being true, with some residual
uncertainty, to the logical reasoning system, which then reasons about
those claims as if they were true. There are thus languages of
different expressive strengths, which are linked within an
architecture.

The existing work in architectures for robot reasoning has some key
limitations. First, many of these systems are driven by the demands of
robot systems engineering, and there is little formalization of the
corresponding architectures. Second, many systems employ a logical
language that is indefeasible, e.g., first order predicate logic, and
incorrect commitments can lead to irrecoverable failures. Our proposed
architecture addresses these limitations. It represents and reasons
about the world, and the robot's knowledge of it, at two
granularities. A fine-resolution description of the domain, close to
the data obtained from the robot's sensors and actuators, is reasoned
about probabilistically, while a coarse-resolution description of the
domain, including commonsense knowledge, is reasoned about using
non-monotonic logic. Our architecture precisely defines the coupling
between the representations at the two granularities, enabling the
robot to represent and efficiently reason about commonsense knowledge,
what the robot does not know, and how actions change the robot's
knowledge. The interplay between the two types of knowledge is viewed
as a conversation between, and the (physical and mental) actions of, a
\emph{logician} and a \emph{statistician}.  Consider, for instance,
the following exchange: \medskip
\begin{s_itemize}
\item[Logician:] \emph{the goal is to find the robotics book. I do not
    know where it is, but I know that books are typically in the
    library and I am in the library. We should first look for the
    robotics book in the library}.  

\vspace{0.5em}
\item[Logician $\to$ Statistician:] \emph{look for the robotics book
    in the library.  You only need to reason about the robotics book
    and the library}.

\vspace{0.5em}
\item[Statistician:] \emph{In my representation of the world, the
    library is a set of grid cells. I shall determine how to locate
    the book probabilistically in these cells considering the
    probabilities of movement failures and visual processing
    failures}.

\vspace{0.5em}
\item[Statistician:] \emph{I visually searched for the robotics book
    in the grid cells of the library, but did not find the book.
    Although there is a small probability that I missed the book, I am
    prepared to commit that the robotics book is not in the library}.

\vspace{0.5em}
\item[Statistician $\to$ Logician:] \emph{here are my observations
    from searching the library; the robotics book is not in the
    library}.

\vspace{0.5em}
\item[Logician:] \emph{the robotics book was not found in the library
    either because it was not there, or because it was moved to
    another location. The next default location for books is the
    bookshelf in the lab. We should go look there next}.

\vspace{0.5em}
\item[and so on...]
\end{s_itemize}
where the representations used by the logician and the statistician,
and the communication of information between them, is coordinated by a
\emph{controller}. This imaginary exchange illustrates key features of
our approach:
\begin{itemize}
\item Reasoning about the states of the domain, and the effects of
  actions, happens at different levels of granularity, e.g., the
  logician reasons about rooms, whereas the statistician reasons about
  grid cells in those rooms.
\item For any given goal, the logician computes a plan of abstract
  actions, and each abstract action is executed probabilistically as a
  sequence of concrete actions planned by the statistician.
\item The effects of the coarse-resolution (logician's) actions are
  non-deterministic, but the statistician's fine-resolution action
  effects, and thus the corresponding beliefs, have probabilities
  associated with them.
\item The coarse-resolution knowledge base (of the logician) may
  include knowledge of things that are irrelevant to the current goal.
  Probabilistic reasoning at fine resolution (by statistician) only
  considers things deemed relevant to the current coarse-resolution
  action.
\item Fine-resolution probabilistic reasoning about observations and
  actions updates probabilistic beliefs, and highly likely statements
  (e.g., probability $> 0.9$) are considered as being completely
  certain for subsequent coarse-resolution reasoning (by the
  logician).
% \item Coarse-resolution reasoning includes prioritized default
%   knowledge, knowledge effects of actions, and the fact that those
%   knowledge effects can be non-deterministic. It can also reason about
%   how things were in the past to explain the current observations (of
%   the statistician).
\end{itemize}
% This paper provides both a full, formal specification of the
% architecture and a working implementation of the architecture on a
% mobile robot platform.

%%%%%-----------------------------------------------------------------------------
\subsection{Technical Contributions}
\label{sec:intro-contrib}
The design of our architecture is based on tightly-coupled transition
diagrams at two levels of granularity. A coarse-resolution description
includes commonsense knowledge, and the fine-resolution transition
diagram is defined as a \emph{refinement} of the coarse-resolution
transition diagram. For any given goal, non-monotonic logical
reasoning with the coarse-resolution system description and the
system's recorded history, results in a sequence of \emph{abstract}
actions. Each such abstract action is implemented as a sequence of
\emph{concrete} actions by \emph{zooming} to a part of the
fine-resolution transition diagram relevant to this abstract action,
and probabilistically modeling the non-determinism in action outcomes.
The technical contributions of this architecture are summarized below.

\paragraph{Action language extensions.} An action language is a
formalism used to model action effects, and many action languages have
been developed and used in robotics, e.g., STRIPS,
PDDL~\cite{ghallab:plan04}, BC~\cite{lee:ijcai13}, and
$\mathcal{AL}_d$~\cite{Gelfond:ANCL13}. We extend $\mathcal{AL}_d$ in
two ways to make it more expressive. First, we allow fluents (domain
attributes that can change) that are non-Boolean, which allows us to
compactly model a wider range of situations.  Second, we allow
non-deterministic causal laws, which captures the non-deterministic
effects of the robot's actions, not only in probabilistic but also
qualitative terms.  This extended version of $\mathcal{AL}_d$ is used
to describe the coarse-resolution and fine-resolution transition
diagrams of the proposed architecture.

\paragraph{Defaults, histories and explanations.} Our architecture
makes three contributions related to reasoning with default knowledge
and histories. First, we expand the notion of the history of a dynamic
domain, which typically includes a record of actions executed and
observations obtained (by the robot), to support the representation of
(prioritized) default information. We can, for instance, say that a
textbook is typically found in the library and, if it is not there, it
is typically found in the auxiliary library.  Second, we define the
notion of a model of a history with defaults in the initial state,
enabling the robot to reason with such defaults.  Third, we limit
reasoning with such expanded histories to the coarse resolution, and
enable the robot to efficiently (a) use default knowledge to compute
plans to achieve the desired goal; and (b) reason with history to
generate explanations for unexpected observations. For instance, in
the absence of knowledge about the locations of a specific object, the
robot can construct a plan using the object's default location to
speed up search. Also, the robot can build a revised model of the
history to explain subsequent observations that contradict
expectations based on initial assumptions.

\paragraph{Tightly-coupled transition diagrams.} The next set of
contributions are related to the relationship between different models
of the domain used by the robot, i.e., the tight coupling between the
transition diagrams at two resolutions. First, we provide a formal
definition of one transition diagram being a \emph{refinement} of
another, and use this definition to formalize the notion of the
coarse-resolution transition diagram being refined to obtain the
fine-resolution transition diagram---the fact that both transition
diagrams are described in the same language facilitates their
construction and this formalization. A coarse-resolution state is, for
instance, magnified to provide multiple states at the
fine-resolution---the corresponding ability to reason about space at
two different resolutions is central for scaling to larger
environments. We find two resolutions to be practically sufficient for
many robot tasks, and leave extensions to other resolutions as an open
problem. Second, we define \emph{randomization} of a fine-resolution
transition diagram, replacing deterministic causal laws by
non-deterministic ones.  Third, we formally define and automate
\emph{zooming} to a part of the fine-resolution transition diagram
relevant to a specific coarse-resolution transition, allowing the
robot, while executing any given abstract action, to avoid considering
parts of the fine-resolution diagram irrelevant to this action, e.g.,
a robot moving between two rooms only considers its location in the
cells in those rooms.

\paragraph{Dynamic generation of probabilistic representations.} The
next set of innovations connect the contributions described so far to
quantitative models of action and observation uncertainty. First, we
use a semi-supervised algorithm, the randomized fine-resolution
transition diagram, prior knowledge (if any), and experimental trials,
to collect statistics and compute probabilities of fine-resolution
action outcomes and observations. Second, we provide an algorithm
that, for any given abstract action, uses these computed probabilities
and the zoomed fine-resolution description to automatically construct
the data structures for, and thus significantly limit the
computational requirements of, probabilistic reasoning. Third, based
on the coupling between transition diagrams at the two resolutions,
the outcomes of probabilistic reasoning update the coarse-resolution
history for subsequent reasoning.

\paragraph{Methodology and architecture.} The final set of
contributions are related to the overall architecture. First, for the
design of the software components of robots that are re-taskable and
robust, we articulate a methodology that is rather general, provides a
path for proving correctness of these components, and enables us to
predict the robot's behavior. Second, the proposed knowledge
representation and reasoning architecture combines the representation
and reasoning methods from action languages, declarative programming,
probabilistic state estimation and probabilistic planning, to support
reliable and efficient operation. The domain representation for
logical reasoning is translated into a program in
SPARC~\cite{balai:lpnmr13}, an extension of CR-Prolog, and the
representation for probabilistic reasoning is translated into a
partially observable Markov decision process
(POMDP)~\cite{kaelbling:AI98}. CR-Prolog~\cite{Balduccini:aaaisymp03}
(and thus SPARC) incorporates consistency-restoring rules in Answer
Set Prolog (ASP)---in this paper, the terms ASP, CR-Prolog and SPARC
are often used interchangeably---and has a close relationship with our
action language, allowing us to reason efficiently with hierarchically
organized knowledge and default knowledge, and to pose state
estimation, planning, and explanation generation within a single
framework. Also, using an efficient approximate solver to reason with
POMDPs supports a principled and quantifiable trade-off between
accuracy and computational efficiency in the presence of uncertainty,
and provides a near-optimal solution under certain
conditions~\cite{kaelbling:AI98,Ong:ijrr2010}. Third, our architecture
avoids exact, inefficient probabilistic reasoning over the entire
fine-resolution representation, while still tightly coupling the
reasoning at different resolutions. This intentional separation of
non-monotonic logical reasoning and probabilistic reasoning is at the
heart of the representational elegance, reliability and inferential
efficiency provided by our architecture.

\medskip
\noindent
The proposed architecture is evaluated in simulation and on a physical
robot finding and moving objects in an indoor domain. We show that the
architecture enables a robot to reason with violation of defaults,
noisy observations, and unreliable actions, in larger, more complex
domains, e.g., with more rooms and objects, than was possible before.

%%%%%-----------------------------------------------------------------------------
\subsection{Structure of the Paper}
\label{sec:intro-struct}
The remainder of the paper is organized as follows.
Section~\ref{sec:illus-example} introduces a domain used as an
illustrative example throughout the paper, and
Section~\ref{sec:relwork} discusses related work in knowledge
representation and reasoning for robots.
Section~\ref{sec:methodology} presents the methodology associated with
the proposed architecture, and Section~\ref{sec:arch-ald-hist}
introduces definitions of basic notions used to build mathematical
models of the domain. Section~\ref{sec:arch-ald} describes the action
language used to describe the architecture's coarse-resolution and
fine-resolution transition diagrams, and
Section~\ref{sec:arch-hl-hist} introduces histories with initial state
defaults as an additional type of record, describes models of system
histories, and reduces planning with the coarse-resolution domain
representation to computing the answer set of the corresponding ASP
program. Section~\ref{sec:arch-logician} provides the logician's
domain representation base on these definitions.  Next,
Section~\ref{sec:arch-refzoomrand} describes the (a) refinement of the
coarse-resolution transition diagram to obtain the fine-resolution
transition diagram; (b) randomization of the fine-resolution system
description; (c) collection of statistics to compute the probability
of action outcomes and observations; and (d) zooming to the part of
the randomized system description relevant to the execution of any
given abstract action.  Next, Section~\ref{sec:arch-pomdp} describes
how a POMDP is constructed and solved to obtain a policy that
implements the abstract action as a sequence of concrete actions.  The
overall control loop of the architecture is described in
Section~\ref{sec:arch-overall}.  Section~\ref{sec:exp} describes the
experimental results in simulation and on a mobile robot, followed by
conclusions in Section~\ref{sec:conclusion}. In what follows, we refer
to the functions and abstract actions of the coarse-resolution
transition diagram using $H$ as the subscript or superscript. The more
concrete functions and actions of the fine-resolution transition
diagram are referred to using $L$ as the subscript or superscript.

%%%%%-----------------------------------------------------------------------------
%%%%%-----------------------------------------------------------------------------
\section{Illustrative Example: Office Domain}
\label{sec:illus-example}
The following domain (with some variants) will be used as an
illustrative example throughout the paper.

%\medskip
\begin{example2}\label{ex:illus-example}[Office domain]
  {\rm Consider a robot that is assigned the goal of moving specific
    objects to specific places in an office domain. This domain
    contains:
    \begin{itemize}
    \item The sorts: $place$, $thing$, $robot$, and $object$, with
      $object$ and $robot$ being subsorts of $thing$. Sorts $textbook$
      and $cup$ are subsorts of the sort $object$. Sort names and
      constants are written in lower-case, while variable names are in
      uppercase.
    \item Four specific places: $of\!\!fice$, $main\_library$,
      $aux\_library$, and $kitchen$. We assume that these places are
      accessible without the need to navigate any corridors, and that
      doors between these places are open.
    \item A number of instances of subsorts of the sort $object$.
      Also, an instance of the sort $robot$, called $rob_1$; we do not
      consider other robots, but any such robots are assumed to have
      similar sensing and actuation capabilities.
    \end{itemize}
  }
\end{example2}

\begin{figure}[tb]
  \begin{center}
    \subfigure[0.6\textwidth][Domain map] {
      \includegraphics[width=0.6\textwidth]{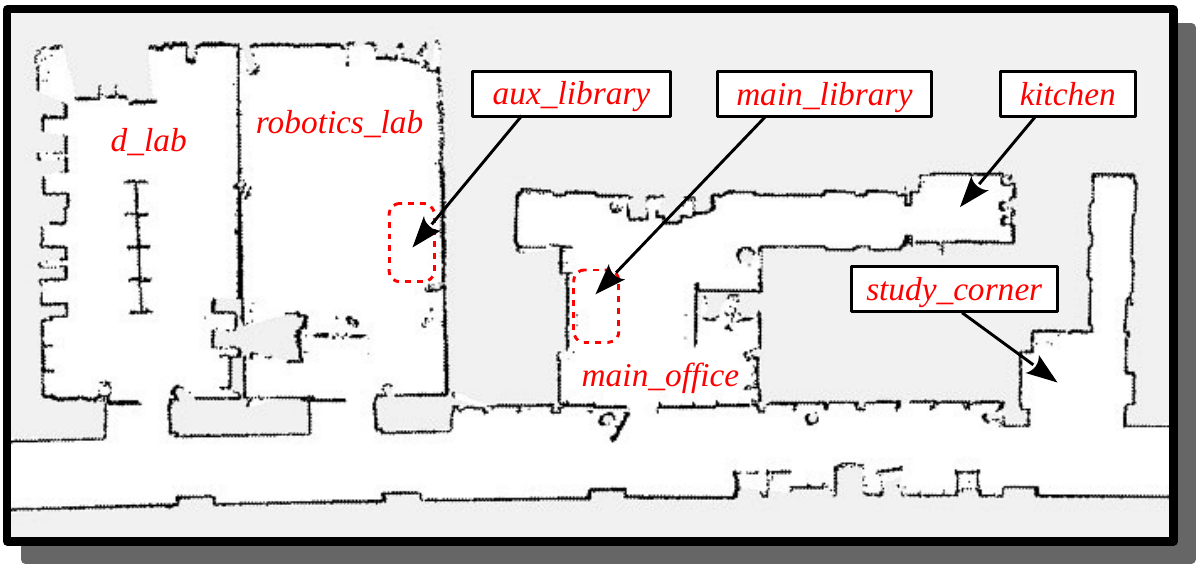}
      \label{fig:map}
    }%\hspace{0.1in}
    \subfigure[0.11\textwidth][Peoplebot] {
      \includegraphics[width=0.11\textwidth]{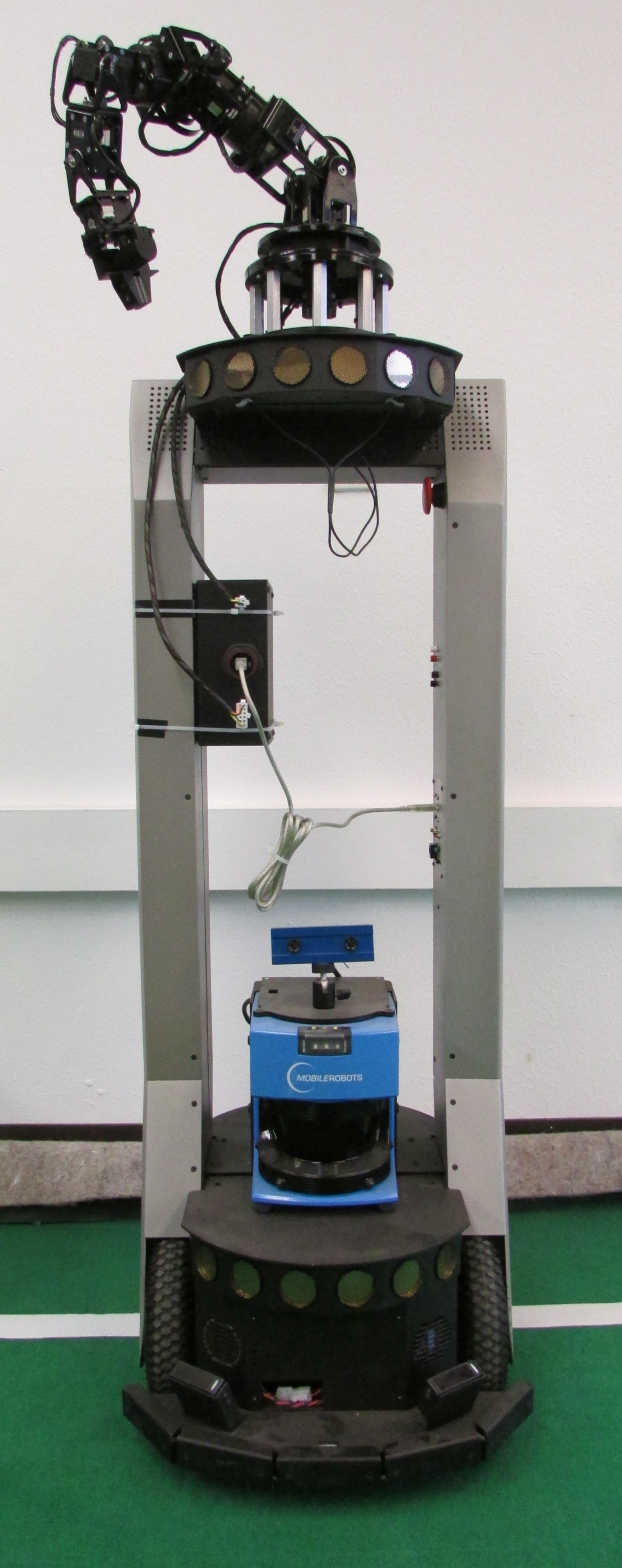}
      \label{fig:robot1}
    }
    \subfigure[0.22\textwidth][Turtlebot] {
      \includegraphics[width=0.22\textwidth]{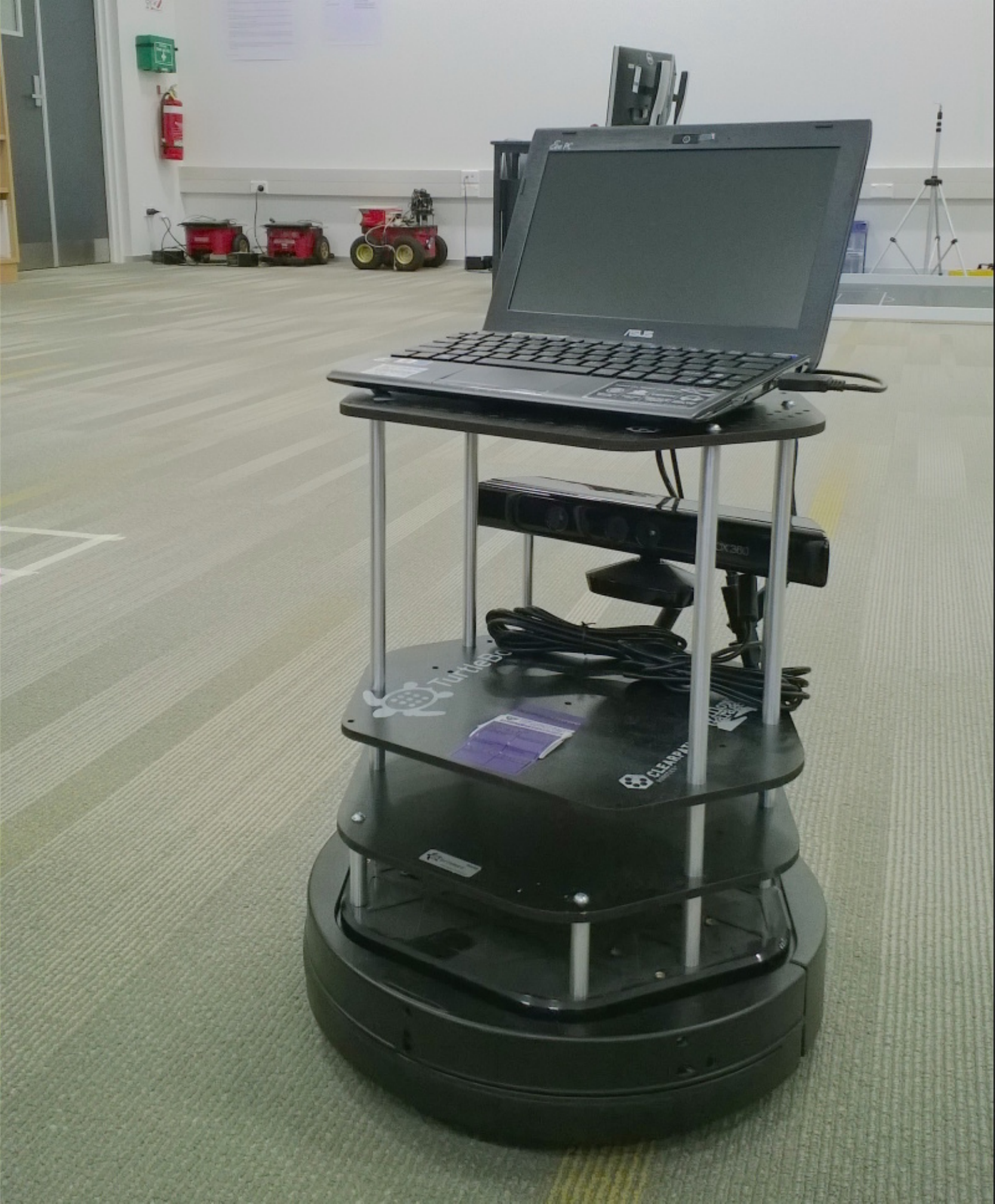}
      \label{fig:robot2}
    }
  \end{center}
  \vspace{-2em}
  \caption{(a) Subset of the map of an entire floor of a
    building---specific places are labeled as shown, and used in the
    goals assigned to the robot; (b)-(c) the ``Peoplebot'' and
    ``Turtlebot'' robot platforms used in the experimental trials.}
  \label{fig:map-robot}
\end{figure}

\noindent
As an extension of this illustrative example that will be used in the
experimental trials on physical robots, consider the robot shown in
Figure~\ref{fig:robot1} operating in an office building whose map is
shown in Figure~\ref{fig:map}.  Assume that the robot can (a) build
and revise the domain map based on laser range finder data; (b)
visually recognize objects of interest; and (c) execute actuation
commands, although neither the information extracted from sensor
inputs nor the action execution is completely reliable.  Next, assume
that the robot is in the study corner and is given the goal of
fetching the robotics textbook. Since the robot knows that books are
typically found in the main library, ASP-based reasoning provides a
plan of abstract actions that require the robot to go to the main
library, pick up the book and bring it back. For the first abstract
action, i.e., for moving to the main library, the robot can focus on
just the relevant part of the fine-resolution representation, e.g.,
the cells through which the robot must pass, but not the robotics book
that is irrelevant at this stage of reasoning.  It then creates and
solves a POMDP for this movement sub-task, and executes a sequence of
concrete movement actions until it believes that it has reached the
main library with high probability. This information is used to reason
at the coarse resolution, prompting the robot to execute the next
abstract action to pick up the robotics book. Now, assume that the
robot is unable to pick up the robotics book because it fails to find
the book in the main library despite a thorough search.  This
observation violates what the robot expects to see based on default
knowledge, but the robot explains this by understanding that the book
was not in the main library to begin with, and creates a plan to go to
the auxiliary library, the second most likely location for textbooks.
In this case, assume that the robot finds the book and completes the
task. The proposed architecture enables such robot behavior.

%%%%%-----------------------------------------------------------------------------
%%%%%-----------------------------------------------------------------------------
%\vspace*{-0.75em}
\section{Related Work}
%\vspace*{-0.75em}
\label{sec:relwork}
The objective of this paper is to enable robots to represent and
reason with logic-based and probabilistic descriptions of incomplete
domain knowledge and degrees of belief. We review some related work
below.

There are many recent examples of researchers using probabilistic
graphical models such as POMDPs to formulate tasks such as planning,
sensing, navigation, and interaction on
robots~\cite{Bai:IJRR14,Gobelbecker:aaai11,Hoey:CVIU10,rosenthal:aaai11}.
These formulations, by themselves, are not well-suited for reasoning
with commonsense knowledge, e.g., default reasoning and non-monotonic
logical reasoning, a key desired capability in robotics. In parallel,
research in classical planning and logic programming has provided many
algorithms for knowledge representation and reasoning, which have been
used on mobile robots.  These algorithms typically require a
significant amount of prior knowledge of the domain, the agent's
capabilities, and the preconditions and effects of the actions. Many
of these algorithms are based on first-order logic, and do not support
capabilities such as non-monotonic logical reasoning, default
reasoning, and the ability to merge new, unreliable information with
the current beliefs in a knowledge base. Other logic-based formalisms
address some of these limitations. This includes, for instance,
theories of reasoning about action and change, as well as Answer Set
Prolog (ASP), a non-monotonic logic programming paradigm, which is
well-suited for representing and reasoning with commonsense
knowledge~\cite{baral:book03,Gelfond:aibook14}. An international
research community has developed around ASP, with applications in
cognitive robotics~\cite{Erdem:bookchap12} and other non-robotics
domains. For instance, ASP has been used for planning and diagnostics
by one or more simulated robot housekeepers~\cite{Erdem:ISR12}, and
for representation of domain knowledge learned through natural
language processing by robots interacting with
humans~\cite{Chen:HRI12}.  ASP-based architectures have also been used
for the control of unmanned aerial vehicles in dynamic indoor
environments~\cite{Balduccini:iclp14}. Recent research has removed the
need to solve ASP programs entirely anew when the problem
specification changes, allowing new information to expand existing
programs, and supporting reuse of ground rules and conflict
information to support interactive theory
exploration~\cite{gebser:lpnmr15}. However, ASP, by itself, does not
support quantitative models of uncertainty, whereas a lot of
information available to robots is represented probabilistically to
quantitatively model the uncertainty in sensor input processing and
actuation.

Many approaches for reasoning about actions and change in robotics and
artificial intelligence (AI) are based on action languages, which are
formal models of parts of natural language used for describing
transition diagrams.  There are many different action languages such
as STRIPS, PDDL~\cite{ghallab:plan04}, BC~\cite{lee:ijcai13}, and
$\mathcal{AL}_d$~\cite{Gelfond:ANCL13}, which have been used for
different applications~\cite{brenner:JAAMAS09,khandelwal:icaps14}.  In
robotics applications, we often need to represent and reason with
recursive state constraints, non-boolean fluents and non-deterministic
causal laws. We expanded $\mathcal{AL}_d$, which already supports
recursive state constraints, to address there requirements. We also
expanded the notion of histories to include initial state defaults.
Action language BC also supports the desired capabilities but it
allows causal laws specifying default values of fluents at arbitrary
time steps, and is thus too powerful for our purposes and occasionally
poses difficulties with representing all exceptions to such defaults
when the domain is expanded.

Refinement of models or action theories has been researched in
different fields. In the field of software engineering and programming
languages, there are approaches for type and model
refinement~\cite{freeman:pldi91,lovas:phdThesis10,lovas:LMCS10,mellies:sopp15}.
These typically do not consider the theories of actions and change
that are important for robot domains.  More recent work in AI has
examined the refinement of action theories of agents in the context of
situation calculus~\cite{banihashemi:aaai17,banihashemi:ijcai18}.
These approaches assume the existence of a bisimulation relation
between the action theories for a given refinement mapping between the
theories, which often does not hold for robotics domains. They also do
not support key capabilities that are needed in robotics such as: (i)
reasoning with commonsense knowledge; (ii) automatic construction and
use of probabilistic models of sensing and actuation; and (iii)
automatic zooming to the relevant part of the refined description.
Although we do not describe it here, it is possible to introduce
simplifying assumptions and a mapping that reduces our approach to one
that is similar to the existing approach.

Robotics and AI researchers have designed algorithms and architectures
based on the understanding that robots interacting with the
environment through sensors and actuators need both logical and
probabilistic reasoning capabilities. For instance, architectures have
been developed to support hierarchical representation of knowledge and
axioms in first-order logic, and probabilistic processing of
perceptual
information~\cite{Laird:agi08,Langley:aaai06,Talamadupula:TIST10},
while deterministic and probabilistic algorithms have been combined
for task and motion planning on robots~\cite{Kaelbling:IJRR13}.
Another example is the behavior control of a robot that included
semantic maps and commonsense knowledge in a probabilistic relational
representation, and then used a continual planner to switch between
decision-theoretic and classical planning procedures based on degrees
of belief~\cite{Hanheide:ijcai11}. The performance of such
architectures can be sensitive to the choice of threshold for
switching between the different planning procedures, and the use of
first order logic in these architectures limits the expressiveness and
use of commonsense knowledge. More recent work has used a
three-layered organization of knowledge (instance, default and
diagnostic), with knowledge at the higher level modifying that at the
lower levels, and a three-layered architecture (competence layer,
belief layer and deliberative layer) for distributed control of
information flow, combining first-order logic and probabilistic
reasoning for open world planning~\cite{Hanheide:AIJ17}. Declarative
programming has also been combined with continuous-time planners for
path planning in mobile robot teams~\cite{Saribatur:iros14}. Most of
these architectures do not provide a tight coupling between the
deterministic and probabilistic components, which has a negative
effect on the computational efficiency, reliability and the ability to
trust the decisions made. More recent work has combined a
probabilistic extension of ASP with POMDPs for commonsense inference
and probabilistic planning in human-robot dialog~\cite{zhang:aaai15},
used a probabilistic extension of ASP to determine some model
parameters of POMDPs~\cite{zhang:aaai17}, and used an ASP-based
architecture to support learning of action costs on a
robot~\cite{khandelwal:icaps14}. ASP-based reasoning has also been
combined with reinforcement learning (RL), e.g., to enable an RL agent
to only explore relevant actions~\cite{leonetti:AIJ16}, or to compute
a sequence of symbolic actions that guides a hierarchical MDP
controller computing actions for interacting with the
environment~\cite{yang:ijcai18}; an architecture combining ASP-based
reasoning with relational RL has been used to interactively and
cumulatively discover domain axioms and
affordances~\cite{mohan:acs17,mohan:icaps17}.

Combining logical and probabilistic reasoning is a fundamental problem
in AI, and many principled algorithms have been developed to address
this problem. For instance, a Markov logic network combines
probabilistic graphical models and first order logic, assigning
weights to logic formulas~\cite{Richardson:ML06}. Bayesian Logic
relaxes the unique name constraint of first-order probabilistic
languages to provide a compact representation of distributions over
varying sets of objects~\cite{milch:bookchap07}.  Other examples
include independent choice logic~\cite{poole:JLP00},
PRISM~\cite{gorlin:TPLP12}, probabilistic first-order
logic~\cite{halpern:book03}, first-order relational
POMDPs~\cite{juba:JMLR16,Sanner:aaai10}, and a system (Plog) that
assigns probabilities to different possible worlds represented as
answer sets of ASP programs~\cite{baral:TPLP09,lee:aaaisss15}. Despite
significant prior research, knowledge representation and reasoning for
robots collaborating with humans continues to present open problems.
Algorithms based on first-order logic do not support non-monotonic
logical reasoning, and do not provide the desired expressiveness for
capabilities such as default reasoning---it is not always possible to
express degrees of belief and uncertainty quantitatively, e.g., by
attaching probabilities to logic statements.  Other algorithms based
on logic programming do not support one or more of the capabilities
such as reasoning about relations as in causal Bayesian networks;
incremental addition of probabilistic information; reasoning with
large probabilistic components; or dynamic addition of variables to
represent open worlds. Our prior work has developed architectures that
support different subsets of these capabilities.  For instance, we
developed an architecture that coupled planning based on a hierarchy
of POMDPs~\cite{Sridharan:AIJ10,zhang:TRO13} with ASP-based inference.
The domain knowledge included in the ASP knowledge base of this
architecture was incomplete and considered default knowledge, but did
not include a model of action effects.  In other work, ASP-based
inference provided priors for POMDP state estimation, and observations
and historical data from comparable domains were considered for
reasoning about the presence of target objects in the
domain~\cite{zhang:TRO15}. This paper builds on our more recent work
on a general, refinement-based architecture for knowledge
representation and reasoning in
robotics~\cite{mohan:knowpros16,zhang:icsr14}. The architecture
enables robots to represent and reason with tightly-coupled transition
diagrams at two different levels of granularity. In this paper, we
formalize and establish the properties of this coupling, present a
general methodology for the design of software components of robots,
provide a path for establishing correctness of these components, and
describe detailed experimental results in simulation and on physical
robot platforms.

%%%%%-----------------------------------------------------------------------------
%%%%%-----------------------------------------------------------------------------
\section{Design Methodology}
\label{sec:methodology}
Our proposed architecture is based on a design methodology. A designer
following this methodology will:
\begin{enumerate}
\item Provide a coarse-resolution description of the robot's domain in
  action language $\mathcal{AL}_d$ together with the description of the initial
  state.

\item Provide the necessary domain-specific information for, and
  construct and examine correctness of, the fine-resolution refinement
  of the coarse-resolution description.

\item Provide domain-specific information and randomize the
  fine-resolution description of the domain to capture the
  non-determinism in action execution.

\item Run experiments and collect statistics to compute probabilities
  of the outcomes of actions and the reliability of observations.

\item Provide these components, together with any desired goal, to a
  reasoning system that directs the robot towards achieving this goal.
\end{enumerate}
The reasoning system implements an action loop that can be viewed as
an interplay between a logician and statistician
(Section~\ref{sec:intro} and Section~\ref{sec:arch-overall}). In this
paper, the reasoning system uses ASP-based non-monotonic logical
reasoning, POMDP-based probabilistic reasoning, models and
descriptions constructed during the design phase, and records of
action execution and observations obtained from the robot. The
following sections describe components of the architecture, design
methodology steps, and the reasoning system.  We first define some
basic notions, specifically action description and domain history,
which are needed to build mathematical models of the domain.

%%%%%-----------------------------------------------------------------------------
%%%%%-----------------------------------------------------------------------------
\section{Action Language and Histories}
\label{sec:arch-ald-hist}
This section first describes extensions to action language
$\mathcal{AL}_d$ to support non-boolean fluents and non-deterministic
causal laws (Section~\ref{sec:arch-ald}).  Next,
Section~\ref{sec:arch-hl-hist} expands the notion of the history of a
dynamic domain to include initial state defaults, defines models of
such histories, and describes how these models can be computed.
Section~\ref{sec:arch-hl-reason} describes how these models can be
used for reasoning.  The subsequent sections describe the use of these
models (of action description and history) to provide the
coarse-resolution description of the domain, and to build more refined
fine-resolution models of the domain.

% The models of action description and history are used in the
% coarse-resolution domain description, and to build more refined
% fine-resolution models of the domain.

%%%%%-----------------------------------------------------------------------------
\subsection{$\mathcal{AL}_d$ with non-boolean functions and non-determinism}
\label{sec:arch-ald}
Action languages are formal models of parts of natural language used
for describing transition diagrams.  In this paper, we extend action
language $\mathcal{AL}_d$~\cite{Gelfond:seasp09,Gelfond:ANCL13,Gelfond:aibook14}
(we preserve the old name for simplicity) to allow functions (fluents
and statics)with non-boolean values, and non-deterministic causal
laws.

%%%--------------------------------------------------------------------------------
\subsubsection{Syntax and informal semantics of $\mathcal{AL}_d$}
\label{sec:arch-ald-informal}
The description of the syntax of $\mathcal{AL}_d$ will require some preliminary
definitions.  

\medskip
\noindent 
\textbf{\underline{Sorted Signature:}} By \emph{sorted signature} we
mean a tuple:
\begin{align*}
  \Sigma=\langle \mathcal{C},\mathcal{S}, \mathcal{F}\rangle
\end{align*}
where $\mathcal{C}$ and $\mathcal{F}$ are sets of strings, over some
fixed alphabet, which are used to name ``user-defined'' \emph{sorts}
and \emph{functions} respectively. $\mathcal{S}$ is a \emph{sort
  hierarchy}, a directed acyclic graph whose nodes are labeled by sort
names from $\mathcal{C}$.  A link $\langle c_1, c_2 \rangle$ of
$\mathcal{S}$ indicates that $c_1$ is a subsort of $c_2$.  A pair
$\langle \mathcal{C},\mathcal{S} \rangle$ will occasionally be
referred to as an \emph{ontology}.  Each function symbol $f \in
\mathcal{F}$ is assigned a non-negative integer $n$ (called $f$'s
arity), sorts $c_0,\dots,c_n$ for its parameters, and sort $c$ for its
values.  We refer to $c_0\times c_2\dots \times c_n$ as the
\emph{domain} of $f$, written as $dom(f)$, and to $c$ as the
\emph{range} of $f$, written as $range(f)$. If $n > 0$ we use the
standard mathematical notation $f : c_0\times\dots\times c_{n}
\rightarrow c$ for this assignment. We refer to a vector
$c_0,\dots,c_n,c$ as the \emph{signature} of $f$. For function of
arity $0$ (called \emph{object constants}), the notation turns into $f
: c$.  We say that $o : c$ is \emph{compatible} with sort $c^\prime$
if $\mathcal{S}$ contains a path from $c$ to $c^\prime$.  A sort
denoted by a sort name $c$ is the collection $\{o_1,\dots,o_n\}$ of
all object constants compatible with $c$; this will be written as $c =
\{o_1,\dots,o_n\}$.

In addition to all these ``user-defined'' sorts and functions, sorted
signatures often contain standard arithmetic symbols such as $0, 1,
2,\dots$ of sort $N$ of natural numbers, and relations and functions
such as $\geq$ and $+$, which are interpreted in the usual way.

\medskip 
\noindent
\emph{Terms} of a sorted signature $\Sigma$ are constructed from
variables and function symbols as follows:
\begin{itemize}
\item A variable is a term. 
\item An object constant $o:c$ is a term of sort $c$. 
\item If $f : c_0\times\dots\times c_{n} \rightarrow c$ where $n > 0$
  is a function symbol and $o_i$ is a variable or a constant
  compatible with sort $c_i$ for $0 \leq i \leq n$, then
  $f(o_0,\dots,o_n)$ is a term of sort $c$.
\end{itemize}
\emph{Atoms} of $\Sigma$ are of the form: 
\begin{align*}
  f(\bar{x})= y 
\end{align*}
where $y$ and elements of $\bar{x}$ are variables or properly typed
object constants, or standard arithmetic atoms formed by $\geq$, $>$,
etc.  If $f$ is boolean, we use the standard notation $f(\bar{x})$ and
$\neg f(\bar{x})$.  \emph{Literals} are expressions of the form
$f(\bar{x})= y$ and $f(\bar{x})\not= y$.  Terms and literals not
containing variables are called \emph{ground}.

\medskip
\noindent 
\textbf{\underline{Action Signature:}} Signatures used by action
languages are often referred to as \emph{action signatures}. They are
sorted signatures with some special features that include various
classifications of functions from $\mathcal{F}$ and the requirements
for inclusion of a number of special sorts and functions. In what
follows, we describe the special features of the action signatures
that we use in this paper.

To distinguish between actions and attributes of the domain
$\mathcal{F}$ is divided into two disjoint parts: $\mathcal{A}$ and
$\mathcal{DA}$. Functions from $\mathcal{A}$ are always boolean. Terms
formed by function symbols from $\mathcal{A}$ and $\mathcal{DA}$ will
be referred to as \emph{actions} and \emph{domain attributes}
respectively. $\mathcal{DA}$ is further partitioned into
$\mathcal{DA}_s$ and $\mathcal{DA}_f$. Terms formed by functions from
$\mathcal{DA}_s$ are referred to as \emph{statics}, and denote domain
attributes whose truth values cannot be changed by actions (e.g.,
locations of walls and doors).  Terms formed by functions from
$\mathcal{DA}_f$ are referred to as \emph{fluents}. $\mathcal{DA}_f$
is further divided into $\mathcal{DA}_{bf}$ and $\mathcal{DA}_{df}$.
Terms formed by symbols from $\mathcal{DA}_{bf}$ are called
\emph{basic fluents} and those formed by symbols from
$\mathcal{DA}_{df}$ are called \emph{defined fluents}. The defined
fluents are always boolean---they do not obey laws of inertia, and are
defined in terms of other fluents. Basic fluents, on the other hand,
obey laws of inertia (thus often called \emph{inertial fluents} in the
knowledge representation literature) and are directly changed by
actions.  Distinction between basic fluents and defined fluents, as
introduced in~\cite{Gelfond:ANCL13}, was the key difference between
the previous version of $\mathcal{AL}_d$ and its predecessor
$\mathcal{AL}$.

% THE CONFUSION STARTS HERE! WE ARE NOT PARTITIONING BASIC FLUENTS WHICH
% ARE TERMS. WE ARE PARTITIONING FUNCTIONS WHICH FORM BASIC FLUENTS. 
% I DO NOT YET NOW WHICH APPROACH IS BETTER.

The new version of $\mathcal{AL}_d$ described in this paper introduces
an additional partition of basic fluents into \emph{basic physical
  fluents} ($\mathcal{DA}_{pbf}$) describing physical attributes of
the domain, and \emph{basic knowledge fluent} ($\mathcal{DA}_{kbf}$)
describing the agent's knowledge.  There is a similar partition of
$\mathcal{A}$ into \emph{physical actions} ($\mathcal{A}_p$) that can
change the physical state of the world (i.e., the value of physical
fluents), and \emph{knowledge producing actions} that are only capable
of changing the agent's knowledge (i.e., the value of knowledge
fluents). Since robots observe their world through sensors, we also
introduce \emph{observable fluents} ($\mathcal{DA}_{obsf}$) to
represent the fluents whose values can be checked by the robot by
processing sensor inputs, or inferred based on the values of other
fluents. The set $\mathcal{DA}_{obsf}$ can be divided into two parts:
the set $\mathcal{DA}_{dobsf}$ of \emph{directly observable fluents},
i.e.  fluents whose values can be observed directly through sensors,
and the set $\mathcal{DA}_{indobsf}$ of \emph{indirectly observable
  fluents} i.e., fluents whose values are not observed directly but
are (instead) inferred from the values of other directly or indirectly
observed fluents. For instance, in Example~\ref{ex:illus-example}, the
robot in any given grid cell can directly observe if a cup is in that
grid cell. The observation of the cup in a particular cell can be used
to infer the room location of the cup. Our classification of functions
is also expanded to literals of the language. If $f$ is static then
$f(\bar{x})=y$ is a static literal, if $f$ is a basic fluent then
$f(\bar{x})=y$ is a basic fluent literal.

\medskip
\noindent
In addition to the classifications of functions, action signatures
considered in this paper also include a collection of special sorts
like $robot$, $place$, etc., and fluents intrinsic to reasoning about
observations. We will refer to the latter as \emph{observation related
  fluents}.  A typical example is a collection of defined fluents:
\begin{align}
  \label{eqn:coarse-observable}
  observable_{f}: robot\times dom(f) \times range(f) \to boolean
\end{align}
where $f$ is an observable function. These fluents are used to specify
domain-specific conditions under which a particular robot can observe
particular values of particular observable fluents. For instance, in
the domain in Example~\ref{ex:illus-example}, we may need to say that
robot $rob_1$ can only observe the place location of an object if it
is also in the same place:
\begin{align*}
  observable_{loc}(rob_1, O, Pl)~~\mathbf{if}~~loc(rob_1)=Pl
\end{align*}
For readability, we will slightly abuse the notation and write the
above statements as:
\begin{align*}
  observable : robot \times obs\_fluent\times value \to
  boolean
\end{align*}
where $obs\_fluent$ stands for ``observable fluent'', and:
\begin{align*}
  observable(rob_1, loc(O), Pl)~~\mathbf{if}~~loc(rob_1)=Pl
\end{align*}
In Section~\ref{sec:arch-refine-observe}, we describe the use of these
(and other such) observation-related fluents for describing a theory
of observations. Then, in Section~\ref{sec:exp-setup}, we describe the
processing of inputs from sensors to observe the values of fluents.

\medskip
\noindent 
\textbf{\underline{Statements of $\mathcal{AL}_d$:}} Action language
$\mathcal{AL}_d$ allows five types of statements: \emph{deterministic
  causal laws}, \emph{non-deterministic causal laws}, \emph{state
  constraints}, \emph{definitions}, and \emph{executability
  conditions}. With the exception of non-deterministic causal law
(Statement~\ref{eqn:causal-law-1}), these statements are built from
ground literals.
\begin{itemize}
\item Deterministic causal laws are of the form:
  \begin{align}
    \label{eqn:causal-law-0}
    a \ \mathbf{causes}\ f(\bar{x})=y~~ \mathbf{if}\
    ~body 
  \end{align}
  where $a$ is an action literal, $f$ is a basic fluent literal, and
  $body$ is a collection of fluent and static literals. If $a$ is
  formed by a knowledge producing action, $f$ must be a knowledge
  fluent.  Intuitively, Statement~\ref{eqn:causal-law-0} says that if
  $a$ is executed in a state satisfying $body$, the value of $f$ in
  any resulting state would be $y$. Non-deterministic causal laws are
  of the form:
  \begin{align}
    \label{eqn:causal-law-1}
    a \ \mathbf{causes}\ f(\bar{x})\ =\ \{Y : p(Y)\} ~\mathbf{if}\
    ~body
  \end{align}
  where $p$ is a unary boolean function symbol from $\mathcal{DA}$,
  or:
  \begin{align}
    \label{eqn:causal-law-2}
    a \ \mathbf{causes}\ f(\bar{x})\ :\ sort\_name ~\mathbf{if}\ ~body
  \end{align}
  Statement~\ref{eqn:causal-law-1} says that if $a$ is executed in a
  state satisfying $body$, $f$ may take on any value from the set
  $\{Y: p(Y)\} \cap range(f)$ in the resulting state.
  Statement~\ref{eqn:causal-law-2} says that $f$ may take any value
  from $\{sort\_name \cap range(f)\}$. If the $body$ of a causal law
  is empty, the $\mathbf{if}$ part of the statement is omitted. Note
  that these axioms are formed from terms and literals that are
  ground, and (possibly) from the expression $\{Y : p(Y)\}$ that is
  sometimes referred to as a \emph{set term}. Occurrences of $Y$ in a
  set term are called \emph{bound}. A statement of $\mathcal{AL}_d$ is
  \emph{ground} if every variable occurring in it is bound.  Even
  though the syntax of $\mathcal{AL}_d$ only allows ground sentences,
  we often remove this limitation in practice. For instance, in the
  context of Example~\ref{ex:illus-example}, we may have the
  deterministic causal law:
  \begin{align*}
    &move(R, Pl)~~\mathbf{causes}~~loc(R) = Pl \nonumber 
  \end{align*}
  which says that for every robot $R$ moving to place $Pl$ will end up
  in $Pl$. In action languages, each such statement is usually
  understood as shorthand for a collection of its ground instances,
  i.e., statements obtained by replacing its variables by object
  constants of the corresponding sorts. We use a modified version of
  this approach in which only non-bound variables are eliminated in
  this way.

\item State constraints are of the form:
  \begin{equation}
    \label{eqn:constraint}
    f(\bar{x}) = y\ ~\mathbf{if}\ ~body
  \end{equation}
  where $f$ is a basic fluent or static. The state constraint says
  that $f(\bar{x}) = y$ must be true in every state satisfying $body$.
  For instance, the constraint:
  \begin{align*}
    loc(Ob) = Pl~~\mathbf{if}~~loc(R) = Pl,~~in\_hand(R, Ob) 
  \end{align*}
  guarantees that the object grasped by a robot shares the robot's
  location.

\item The definition of the value of a defined fluent $f$ on $\bar{x}$
  is a collection of statements of the form:
  \begin{align}
    \label{eqn:definition}
    f(\bar{x})\ ~\mathbf{if}\ ~body
  \end{align}
  where $f(\bar{x})$ is true if it follows from the truth of at least
  one of its defining rules.  Otherwise, $f(\bar{x})$ is false.

\item Executability conditions are statements of the form:
  \begin{align}
    \label{eqn:exec}
    \mathbf{impossible}\ a_0,\ldots,a_k \ \mathbf{if}\ ~body
  \end{align}
  which implies that in a state satisfying $body$, actions $a_0,\ldots
  a_k$ cannot occur simultaneously. For instance, the following
  executability condition:
  \begin{align*}
    \mathbf{impossible}~~move(R, Pl)~~\mathbf{if}~~loc(R)=Pl  
  \end{align*} 
  implies that a robot cannot move to a location if it is already
  there; and
  \begin{align*}
    \mathbf{impossible}~~grasp(R_1, Th),
    grasp(R_2,Th)~~\mathbf{if}~~R_1 \not= R_2
  \end{align*}
  prohibits two robots from simultaneously grasping the same thing.
\end{itemize}
We can now define the notion of a system description.  
\begin{defn2}\label{def:ald-sd}[System Description]\\
  A \emph{system description} of $\mathcal{AL}_d$ is a collection of
  $\mathcal{AL}_d$ statements over some action signature $\Sigma$.
\end{defn2}
\noindent
Next, we discuss the formal semantics of $\mathcal{AL}_d$.

%%%--------------------------------------------------------------------------------
\subsubsection{Formal semantics of $\mathcal{AL}_d$}
\label{sec:arch-ald-formal}
The semantics of system description $\mathcal{D}$ of the new
$\mathcal{AL}_d$ is similar to that of the old one. In fact, the old
language can be viewed as the subset of $\mathcal{AL}_d$ in which all
functions are boolean, causal laws are deterministic, and no
distinction is made between physical and knowledge related actions and
fluents. The semantics of $\mathcal{D}$ is given by a transition
diagram ${\tau(\mathcal{D})}$ whose nodes correspond to possible
states of the system. The diagram contains an arc $\langle
\sigma_1,a,\sigma_2 \rangle$ if, after the execution of action $a$ in
state $\sigma_1$, the system may move into state $\sigma_2$. We define
the states and transitions of ${\tau(\mathcal{D})}$ in terms of answer
sets of logic programs, as described
below---see~\cite{Gelfond:seasp09,Gelfond:aibook14} for more details.

In what follows, unless otherwise stated, by ``atom'' and ``term'' we
refer to ``ground atom'' and ``ground term'' respectively.  Recall
that an \emph{interpretation} of the signature of $\mathcal{D}$ is an
assignment of a value to each term $f(\bar{x})$ in the signature. An
interpretation can be represented by the collection of atoms of the
form $f(\bar{x})= y$, where $y$ is the value of $f(\bar{x})$.  For any
interpretation $\sigma$, let $\sigma^{nd}$ be the collection of atoms
of $\sigma$ formed by basic fluents and statics---${nd}$ stands for
\emph{non-defined}. Let $\Pi_c(\mathcal{D})$, where $c$ stands for
constraints, denote the logic program defined as:
\begin{enumerate}
\item For every state constraint (Statement~\ref{eqn:constraint}) and
  definition (Statement~\ref{eqn:definition}), program
  $\Pi_c(\mathcal{D})$ contains:
  \begin{align}
    \label{eqn:pic-state-constraint}
    f(\bar{x})=y~~ \leftarrow body
  \end{align}
\item For every defined fluent $f$, $\Pi_c(\mathcal{D})$ contains the
  closed world assumption (CWA):
  \begin{align}
    \label{eqn:pic-defined-fluent}
    \neg f(\bar{x})\leftarrow \mbox{ not } f(\bar{x})
  \end{align}
  where, unlike classical negation ``$\lnot~a$'' that implies ``a is
  believed to be false'', \emph{default negation} ``$not~a$'' only
  implies that ``a is not believed to be true'', i.e., $a$ can be
  true, false or just unknown.
\end{enumerate}

%\medskip    
\noindent
We can now define states of ${\tau(\mathcal{D})}$.
\begin{defn2}\label{def:ald-state}[State of $\tau(\mathcal{D})$]\\
  {\rm An interpretation $\sigma$ is a {\bf state} of the transition
    diagram ${\tau(\mathcal{D})}$ if it is the unique answer set of
    program $\Pi_c(\mathcal{D}) \cup \sigma^{nd}$.  }
\end{defn2}
\noindent
As an example, consider a system description $\mathcal{D}_s$
from~\cite{Gelfond:aibook14} with two defined fluents $f$ and $g$
defined by mutually recursive laws: 
%\vspace{-1em}
\begin{align*}
  &g~~\mathbf{if}~~\lnot f\\
  &f~~\mathbf{if}~~\lnot g
\end{align*}
For this system description, the program $\Pi_c(\mathcal{D}_s)$
consists of the following statements:
\begin{align*}
  &g~~\leftarrow~~\lnot f\\
  &f~~\leftarrow~~\lnot g\\
  &\lnot g~~\leftarrow~~not~g\\
  &\lnot f~~\leftarrow~~not~f
\end{align*}
and $\sigma^{nd} = \emptyset$ because all the fluents of
$\mathcal{D}_s$ are defined. $\Pi_c(\mathcal{D}_s) \cup \sigma^{nd}$
has two answer sets $\{f, \lnot g\}$ and $\{g, \lnot f\}$; based on
Definition~\ref{def:ald-state}, the transition diagram
$\tau(\mathcal{D}_s)$ has no states. This outcome is expected because
the mutually recursive laws are not strong enough to uniquely define
$f$ and $g$.

Next, we define a sufficient condition for guaranteeing that the
defined fluents of a system description are uniquely defined by the
system's statics and basic fluents. To do so, we introduce some
terminology from~\cite{Gelfond:aibook14}. A system description
$\mathcal{D}$ is said to be \emph{well-founded} if for any complete
and consistent set of fluent literals and statics $\sigma$ satisfying
the state constraints of $\mathcal{D}$, program $\Pi_c(\mathcal{D})
\cup \sigma^{nd}$ has an unique answer set. Next, the \emph{fluent
  dependency graph} of $\mathcal{D}$ is the directed graph such that:
\begin{itemize}
\item its vertices are arbitrary domain literals.
\item it has an edge:
  \begin{itemize}
  \item from $l$ to $l'$ if $l$ is formed by a static or a basic
    fluent, and $\mathcal{D}$ contains a state constraint with the
    head $l$ and the body containing $l'$;
  \item from $f$ to $l'$ if $f$ is a defined fluent, and $\mathcal{D}$
    contains a state constraint with the head $f$ and body containing
    $l'$ and not $f$; and
  \item from $\lnot f$ to $f$ for every defined fluent $f$.
  \end{itemize}
\end{itemize}
Also, a fluent dependency graph is said to be \emph{weakly acyclic} if
it does not contain paths from defined fluents to their negations.  A
system description with a weakly acyclic fluent dependency graph is
also said to be weakly acyclic. Although well-foundedness is not easy
to check, it is easy to check weak acyclicity, and Proposition $8.4.1$
in~\cite{Gelfond:aibook14} establishes weak acyclicity as a sufficient
condition for well-foundedness~\cite{Gelfond:ANCL13}. It is easy to
show that all system descriptions discussed in this paper are
well-founded, a fact that we will use later in this paper.

%\commentm{Insert example 8.4.3; also see pages 169-171}

\medskip
\noindent
Next, to define the transition relation of ${\tau(\mathcal{D})}$, we
first describe the logic programming encoding $\Pi(\mathcal{D})$ of
$\mathcal{D}$. $\Pi(\mathcal{D})$ consists of the encoding of the
signature of $\mathcal{D}$ and rules obtained from statements of
$\mathcal{D}$, as described below.

\begin{defn2}\label{def:ald-program}[Logic programming encoding
  of $\mathcal{D}$]
  {\rm
  \begin{itemize}
  \item \textbf{Encoding of the signature:} we start with the encoding
    $sig(\mathcal{D})$ of signature of $\mathcal{D}$.
    \begin{itemize}
    \item For each sort $c$, $sig(\mathcal{D})$ contains:
      $sort\_name(c)$.

    \item For each subsort link $\langle c_1,c_2\rangle$ of the
      hierarchy of sorts, $sig(\mathcal{D})$ contains:
      $s\_link(c_1,c_2)$.

    \item For each constant $x : c$ from the signature of
      $\mathcal{D}$, $sig(\mathcal{D})$ contains: $m\_link(x, c)$.
    
    \item For every function symbol $f : c_1\times \dots c_n
      \rightarrow c$, the signature $sig(\mathcal{D})$ contains the
      domain: $dom(f,c_1,\dots,c_n)$, and range: $range(f,c)$.

    \item For every static $g(\bar{x})$ of $\mathcal{D}$,
      $sig(\mathcal{D})$ contains: $static(g(\bar{x}))$.

    \item For every basic fluent $f(\bar{x})$, $sig({\mathcal{D}})$
      contains: $fluent(basic,f(\bar{x}))$.
	
    \item For every defined fluent $f(\bar{x})$, $sig(\mathcal{D})$
      contains: $fluent(defined,f(\bar{x}))$.

    \item For every observable fluent $f(\bar{x})$, $sig(\mathcal{D})$
      contains: $obs\_fluent(f(\bar{x}))$.

    \item For every directly observable fluent $f(\bar{x})$,
      $sig(\mathcal{D})$ contains: $dir\_obs\_fluent(f(\bar{x}))$.

    \item For every indirectly observable fluent $f(\bar{x})$,
      $sig(\mathcal{D})$ contains: $indir\_obs\_fluent(f(\bar{x}))$.

    \item For every action $a$ of $\mathcal{D}$, $sig(\mathcal{D})$
      contains: $action(a)$.
    \end{itemize}
    We also need axioms describing the hierarchy of basic sorts:
    \begin{align}
      subsort(C_1,C_2) & \leftarrow s\_link(C_1,C_2) \\\nonumber
      subsort(C_1,C_2) & \leftarrow s\_link(C_1,C),~subsort(C,C_2) \\\nonumber
      member(X,C) & \leftarrow m\_link(X,C) \\\nonumber
      member(X,C_1) & \leftarrow m\_link(X,C_0),~subsort(C_0,C_1)
    \end{align}
    
  \item \textbf{Encoding of statements of $\mathcal{D}$:} To define
    transitions of our diagram we need two time-steps that stand for
    the beginning and the end of a transition. We would like, however,
    to later use the rules of our program to describe longer chains of
    events. To make this possible we introduce a symbolic constant $n$
    and allow time-steps of the program to range over $[0,
    max\_step]$. This is expressed by statement:
    \begin{align*}
      &step(0..max\_step) 
    \end{align*}
    For defining transitions we set $max\_step$ to $1$:
    \begin{align*}
      &\#const\ max\_step = 1
    \end{align*}
    We also need a relation $val(f(x_1,\dots,x_n),y,i)$, which states
    that the value of $f(x_1,\dots,x_n)$ at step $i$ is $y$; and
    relation $occurs(a,i)$, which states that action $a$ occurred at
    step $i$. We then encode statements of $\mathcal{D}$ as follows:
    \begin{itemize}
    \item For every causal law
      (Statements~\ref{eqn:causal-law-1}-\ref{eqn:causal-law-2}),
      where the range of $f$ is $\{y_1,\ldots,y_k\}$,
      $\Pi(\mathcal{D})$ contains a rule:
      \begin{align}
        \label{eqn:pid-causal}
        val(f(\bar{x}),y_1,I+1)\ \mathbf{or}\ \ldots \mathbf{or}\
        val(f(\bar{x}),y_k,I+1) \leftarrow & val(body,I),~ occurs(a,I),~
        I < n
      \end{align}
      where $val(body,I)$ is obtained by replacing every literal
      $f_m(\bar{x}_m) = z$ from $body$ by $val(f_m(\bar{x}_m),z,I)$.
      To encode that due to this action, $f(\bar{x})$ only takes a
      value that satisfies property $p$, $\Pi(\mathcal{D})$ contains a
      constraint:
      \begin{align}
        \label{eqn:pid-causal-constraint1}
        \leftarrow val(f(\bar{x}), Y, I+1), \textrm{ not } val(p(Y),
        true, I)
      \end{align}
      and rules:
      \begin{align}
        \label{eqn:pid-causal-constraint2}
        satisfied(p,I)  &  \leftarrow val(p(Y),true, I)\\\nonumber
        \neg occurs(a,I) & \leftarrow \textrm{ not } satisfied(p,I)
      \end{align}
        
    \item For every state constraint and definition
      (Statements~\ref{eqn:constraint},~\ref{eqn:definition}),
      $\Pi(\mathcal{D})$ contains:
      \begin{align}
        \label{eqn:pid-constraint}
        val(f(\bar{x}),y,I) \leftarrow val(body,I)
      \end{align}
      
    \item $\Pi(\mathcal{D})$ contains the CWA for defined fluents:
      \begin{align}
        \label{eqn:pid-defined}
        val(F,false,I) \leftarrow fluent(defined,F),~\textrm{ not}\
        val(F,true, I)
      \end{align}
      
    \item For every executability condition
      (Statement~\ref{eqn:exec}), $\Pi(\mathcal{D})$ contains:
      \begin{align}
        \label{eqn:pid-execute}
        \neg occurs(a_0,I) \ \mathbf{or}\ \dots \ \mathbf{or}\ \ \neg
        occurs(a_k,I) \leftarrow & val(body,I),~ I < n
      \end{align}
      
    \item For every static $g(\bar{x})$, $\Pi(\mathcal{D})$ contains:
      %\vspace{-1em}
      \begin{align}
        \label{eqn:pid-static}
        g(\bar{x}) = y
      \end{align}

    \item $\Pi(\mathcal{D})$ contains the Inertia Axiom:
      \begin{align}
        \label{eqn:pid-inertia}
        val(F,Y,I+1) \leftarrow & fluent(basic,F), \\\nonumber &
        val(F,Y,I),~ not\ \neg val(F,Y,I+1),~ I < n
      \end{align}
      
    \item $\Pi(\mathcal{D})$ contains CWA for actions:
      \begin{align} 
        \label{eqn:pid-action-cwa}
        \neg occurs(A,I) \leftarrow & \no occurs(A,I),~ I < n
      \end{align}
      
    \item Finally, we need the rule:
      \begin{equation} 
        \label{eqn:pid-rule}
        \neg val(F,Y_1,I) \leftarrow val(F,Y_2,I),~Y_1 \not= Y_2
      \end{equation}
      which says that a fluent can only have one value at each time
      step.
    \end{itemize}
  \end{itemize}
  This completes the construction of encoding $\Pi(\mathcal{D})$ of
  system description $\mathcal{D}$. Later we will consider a version
  of $\mathcal{D}$ in which time step $max\_step$ is set to some
  positive number k. We denote such a program by $\Pi^k(\mathcal{D})$.
}
% \end{defn2}
\end{defn2}
\medskip
\noindent
Recall that the axioms described above are shorthand for the set of
ground instances obtained by replacing variables by ground terms from
the corresponding sorts. We now define a transition of
$\tau(\mathcal{D})$.
%\vspace{1em}
%\hrule
\begin{defn2}\label{def:ald-trans}[Transition of $\tau(\mathcal{D})$]\\
  {\rm Let $a$ be a non-empty collection of actions, and $\sigma_0$
    and $\sigma_1$ be states of the transition diagram
    ${\tau(\mathcal{D})}$ defined by a system description
    $\mathcal{D}$. To describe a transition $\langle
    \sigma_0,a,\sigma_1 \rangle$, we construct a program
    $\Pi(\mathcal{D},\sigma_0,a)$ consisting of:
    \begin{itemize}
    \item Logic programming encoding $\Pi(\mathcal{D})$ of system
      description $\mathcal{D}$, as described above.
   
    \item The encoding $val(\sigma_0,0)$ of initial state $\sigma_0$:
      \begin{align*}
        val(\sigma_0,0) =_{def} & \{val(f(\bar{x}),y,0) :
        (f(\bar{x})=y) \in \sigma_0\},\,\, \textrm{where } f
        \textrm{ is a fluent}\,\, \cup \\
        & \{f(\bar{x})=y : (f(\bar{x})=y) \in \sigma_0\} \,\,
        \textrm{where } f \textrm{ is a static}
      \end{align*}

    \item Encoding $occurs(a,0)$ of set of actions $a$:
      \begin{align*}
        occurs(a,0) &=_{def} \{occurs(a_i,0) : a_i \in a\}
      \end{align*}
    \end{itemize}
    In other words, the program $\Pi(\mathcal{D},\sigma_0,a)$ includes
    our description of the system's laws, the initial state, and the
    actions that occur in it:
    %\vspace{-1em}
    \begin{align*}
      \Pi(\mathcal{D},\sigma_0,a)=_{def}\Pi(\mathcal{D}) \cup
      val(\sigma_0,0) \cup occurs(a,0)
    \end{align*}
    A state-action-state triple $\langle \sigma_0,a,\sigma_1\rangle$
    is a {\bf transition} of ${\tau(\mathcal{D})}$ iff
    $\Pi(\mathcal{D},\sigma_0,a)$ has an answer set $AS$ such that
    $\sigma_1 = \{f(\bar{x})=y : val(f(\bar{x}),y,1) \in AS\}$.  The
    answer sets of $\Pi(\mathcal{D},\sigma_0,a)$ thus determine the
    states the system can move into after executing of $a$ in
    $\sigma_0$.

  }
\end{defn2}
%\vspace{1pt}
%\hrule

%%%%%-----------------------------------------------------------------------------
\subsection{Histories with defaults}
\label{sec:arch-hl-hist}
In action languages, domain knowledge is typically represented by a
system description containing general knowledge about the domain and
the agent's abilities, and the domain's \emph{recorded history}
containing information pertinent to the agent's activities in the
domain. This history $\mathcal{H}$ typically contains the agent's
observations of the value of domain attributes, and the occurrences of
actions, as recorded by statements of the form:
\begin{align}
  \label{eqn:obs}
  &obs(rob_1, f(\bar{x}), y, true, i)\\
  &obs(rob_1, f(\bar{x}), y, false, i)
\end{align}
and
\vspace{-1em}
\begin{align}
  \label{eqn:act}
  hpd(a,i)
\end{align}
where $f$ is an observable fluent, $y$ is a possible value of this
fluent, $a$ is an action, and $i$ is a time-step.
Statement~\ref{eqn:obs} says that a particular fluent was observed to
have (or not have) a particular value at time-step $i$ by robot
$rob_1$, and Statement~\ref{eqn:act} says that action $a$ happened at
time-step $i$. For instance, $obs(rob_1, loc(tb_1), of\!\!fice, true,
0)$ denotes the observation of textbook $tb_1$ in the $of\!\!fice$ by
robot $rob_1$, and $hpd(move(rob_1, kitchen), 1)$ is the record of
successful execution of $rob_1$'s move to the kitchen at time step
$1$. Note that the standard representation of $obs$ does not include
the $robot$ as the first argument---we include it to emphasize that
observations are obtained by specific robots. Also, for convenience,
we write $obs(rob_1, f(\bar{x}), y, true, i)$ as $obs(rob_1,
f(\bar{x})=y, i)$, and write $obs(rob_1, f(\bar{x}), y, false, i)$ as
$obs(rob_1, f(\bar{x}) \not= y, i)$. In addition, the notion of
observations at the coarse resolution is different from that of
observations obtained from sensor inputs, which are modeled at the
fine resolution; the former is based on the latter, as described in
Section~\ref{sec:arch-refine}. Furthermore, there is a subtle
difference between relation $occurs$ used in logic programming
encoding of system descriptions and relation $hpd$.  Statement
$occurs(a,i)$ may denote an actual occurrence of action $a$ at $i$ as
well as a hypothetical action (e.g., in a plan computed for a specific
goal), whereas $hpd(a,i)$ indicates that $a$ was actually executed at
$i$. For a discussion on the need for such a distinction between $hpd$
and $occurs$, please see Section $10.5$ in~\cite{Gelfond:aibook14}.

\medskip 
\noindent
We say that $n$ is the \emph{current step} of history $\mathcal{H}$ if
$n-1$ is the maximum time step occurring in statements of the form
$hpd(a,i)$ in $\mathcal{H}$. If no such statement exists, the current
step of $\mathcal{H}$ is $0$. The recorded history thus defines a
collection of paths in the transition diagram that, from the
standpoint of the agent, can be interpreted as the system's possible
pasts. The precise formalization of this is given by the notion of a
\emph{model} of the recorded history. The definition of such a model
for histories consisting of Statements~\ref{eqn:obs} and~\ref{eqn:act}
can be found in Section $10.1$ in~\cite{Gelfond:aibook14}.

\medskip
\noindent
In our work, we extend the syntax and semantics of recorded histories
to support a more convenient description of the domain's initial
state. In addition to the statements above, we introduce an additional
type of historical record: 
%\vspace{-0.5em}
\begin{align}
  \label{eqn:default}
  {\bf initial~~default }~~ d(\bar{x}) : f(\bar{x}) = y~~ {\bf if }\
  ~body(d)
\end{align}
and:
%\vspace{-1em}
\begin{align}
  \label{eqn:pref}
  {\bf prefer}(d_1,d_2)
\end{align}
where $f$ is a basic fluent and the $d$s are the names of defaults.
Statements~\ref{eqn:default} and~\ref{eqn:pref} refer to the initial
state of the system. Statement~\ref{eqn:default} is a default named
$d$ stating that in any initial state satisfying $body(d)$, the
default value of $f(\bar{x})$ is $y$.  Statement~\ref{eqn:pref}
defines an anti-symmetric and transitive preference relation between
defaults, stating that if the simultaneous application of defaults
$d_1$ and $d_2$ leads to a contradiction, then $d_1$ is preferred to
$d_2$.

The addition of defaults makes the task of defining models
substantially more challenging. Before providing a formal semantics of
a recorded history with defaults (i.e., before defining models of such
histories), we illustrate the intended meaning of these statement with
an example.

\begin{figure}[tbh]
  \begin{center}%\hspace*{-1.5em}
    \includegraphics[width=0.9\columnwidth]{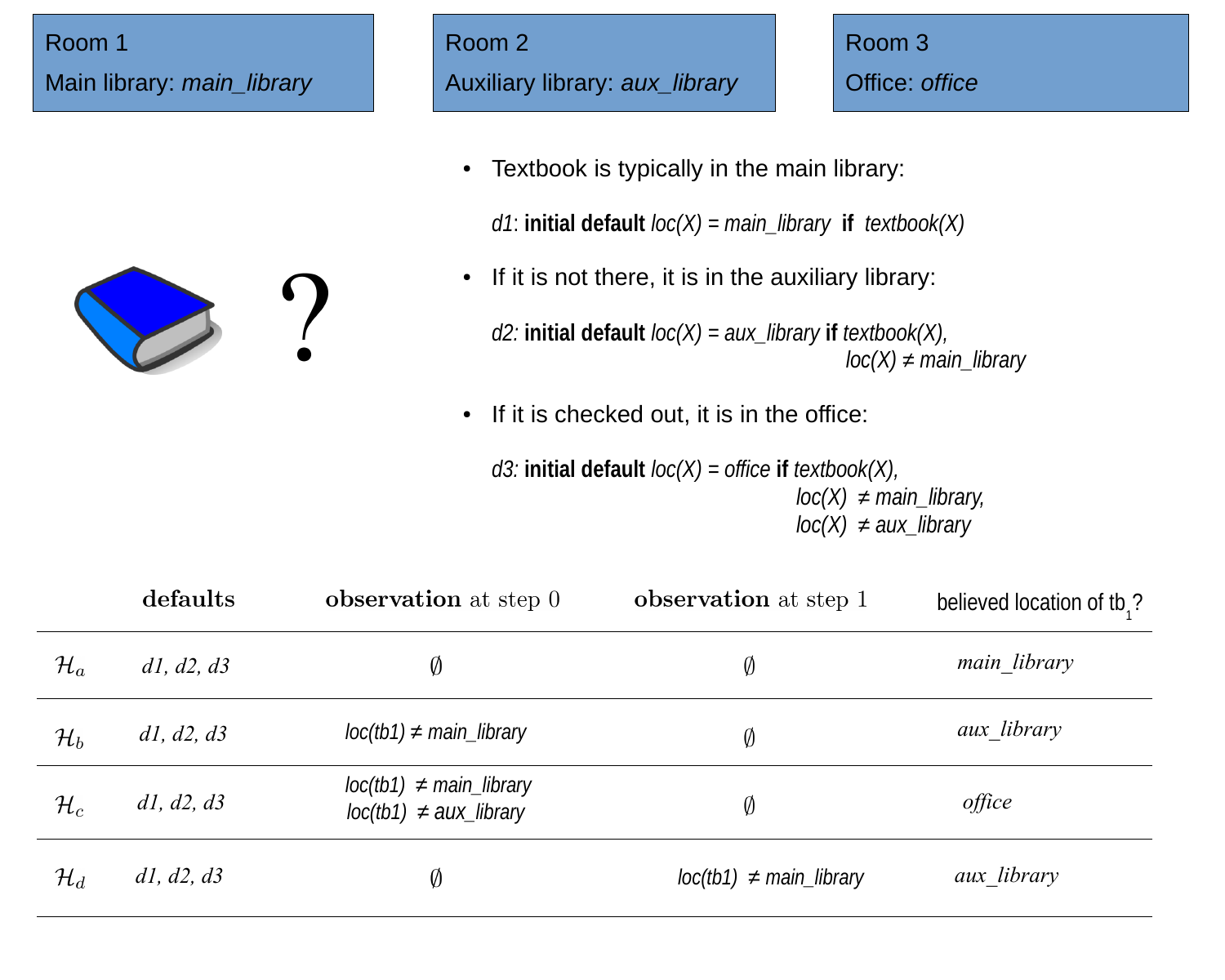}
  \end{center}
  \vspace{-2em}
  \caption{Illustration of the beliefs of a robot corresponding to the
    histories with the same initial state defaults, as described in
    Example~\ref{ex:defaults} and Example~\ref{ex:defaults-again}.}
  \label{fig:hist-default}
\end{figure}

\begin{example2}\label{ex:defaults}[Example of initial state defaults]\\
  {\rm Consider the following statements about the locations of
    textbooks in the initial state in our illustrative example.
    \emph{Textbooks are typically in the main library. If a textbook
      is not there, it is typically in the auxiliary library.  If a
      textbook is checked out, it can usually be found in the office.}
    These defaults can be represented as:
    \begin{align}
      \label{def:d1}
      {\bf initial~~default}~~d_1(X)~~:~~loc(X) = main\_library~~{\bf
        if}~~ textbook(X)
    \end{align}
    \vspace{-1.4em}
    \begin{align}
      \label{def:d2}
      {\bf initial~~default}~~d_2(X)~~:~~loc(X) = aux\_library~~{\bf
        if}~~ & textbook(X)
    \end{align}
    \vspace{-1.4em}
    \begin{align}
      \label{def:d3}
      {\bf initial~~default}~~d_3(X)~~:~~loc(X) = of\!\!fice~~{\bf
        if}~~ & textbook(X)
    \end{align}
    \vspace{-1.4em}
    \begin{align}
      \label{def:pr}
      {\bf prefer}(d_1(X),d_2(X)) \\ \nonumber
%     \end{align}
%     \vspace{-1.4em}
%     \begin{align}
%       \label{def:pr2}
      {\bf prefer}(d_2(X),d_3(X))
    \end{align}
    where the fluent $\{loc : thing \rightarrow place\}$, as before,
    represents the place where a particular thing is located.
    Intuitively, a history $\mathcal{H}_a$ with the above statements
    will entail $val(loc(tb_1)=main\_library, true, 0)$ for textbook
    $tb_1$ using default $d_1(tb_1)$. The other two defaults
    (Statements~\ref{def:d2},~\ref{def:d3}) are disabled (i.e., not
    used) due to Statement~\ref{def:pr} and the transitivity of the
    $prefer$ relation. A history $\mathcal{H}_b$ that adds
    $obs(rob_1, loc(tb_1)\not= main\_library, 0)$ as an observation to
    $\mathcal{H}_a$ renders default $d_1(tb_1)$ (see
    Statement~\ref{def:d1}) inapplicable. Now the second default
    (i.e., Statement~\ref{def:d2}) is enabled and entails
    $val(loc(tb_1)=aux\_library, true, 0)$.  A history $\mathcal{H}_c$
    that adds observation $obs(rob_1, loc(tb_1)\not= aux\_library, 0)$ to
    $\mathcal{H}_b$ should entail $val(loc(tb_1)= of\!\!fice, true,
    0)$. In all these histories, the defaults were defeated by initial
    observations and by higher priority defaults.

    Now, consider the addition of observation $obs(rob_1, loc(tb_1)\not=
    main\_library, 1)$ to $\mathcal{H}_a$ to obtain history
    $\mathcal{H}_d$. This observation is different because it defeats
    default $d_1(tb_1)$, but forces the agent to reason back in time.
    If the default's conclusion, $loc(tb_1)=main\_library$, were true
    in the initial state, it would also be true at step $1$ (by
    inertia), which would contradict the observation. Default
    $d_2(tb_1)$ will be used to conclude that textbook $tb_1$ is
    initially in the $aux\_library$; the inertia axiom will propagate
    this information to entail $val(loc(tb_1)= aux\_library, true,
    1)$. For more information on
    indirect exceptions to defaults and their formalization see
    Section $5.5$ in~\cite{Gelfond:aibook14}.

    Figure~\ref{fig:hist-default} illustrates the beliefs
    corresponding to these four histories---the column labeled ``CR
    rule outcome'' and the row labeled ``$\mathcal{H}_e$'' are
    explained later in this section. Please see \texttt{example2.sp}
    at \url{https://github.com/mhnsrdhrn/refine-arch} for the complete
    program formalizing this reasoning in SPARC.  }
\end{example2}

\noindent
To better understand the definition of histories with defaults, let us
first recall the definition of a model for histories not containing
defaults. In this case, a model of $\mathcal{H}^n$ is a path $M =
\langle \sigma_0,a_0,\sigma_1,\dots,\sigma_{n},a_n\rangle$ of
${\tau(\mathcal{D})}$ such that:
\begin{itemize}
\item $M$ \emph{satisfies} every $obs(rob_1, f(\bar{x})=y, i) \in
  \mathcal{H}^n$, i.e., for every such observation $( f(\bar{x})=y )
  \in \sigma_i$.
\item $a_i = \{ e : hpd(e,i) \in \mathcal{H}^n\}$. 
\end{itemize}
In the presence of defaults, however, these conditions, though
necessary, are not sufficient. Consider, for instance, history
$\mathcal{H}_a$ from Example~\ref{ex:defaults}. Since it contains no
actions or observations, these conditions are satisfied by any path $M
= \langle \sigma_0 \rangle$. However, $M$ is a model of
$\mathcal{H}_a$ only if $\sigma_0$ contains $loc(tb1)=main\_library$.
In general, to define the initial states of models of $\mathcal{H}^n$,
we need to understand reasoning in the presence of defaults, along
with their direct and indirect exceptions. The situation is similar
to, but potentially more complex than, the definition of transitions
of ${\tau(\mathcal{D})}$. To define the models of $\mathcal{H}^n$, we
thus pursue an approach similar to that used to define the transitions
of ${\tau(\mathcal{D})}$.  Specifically, we define models of
$\mathcal{H}^n$ in terms of answer sets of the logic program
$\Pi(\mathcal{D},\mathcal{H})$ that axiomatizes the agent's knowledge.
However, due to the presence of indirect exceptions, our language of
choice will be CR-Prolog, an extension of ASP well-suited for
representing and reasoning with such knowledge. We begin by defining
the program encoding both $\mathcal{D}$ and $\mathcal{H}$.

% Now we are ready to define models of a recorded history
% $\mathcal{H}$ of system description $\mathcal{D}$, i.e. \emph{paths
%   of ${\tau(\mathcal{D})}$ considered possible by the agent
%   recording this history}.  Similar to the definition of transitions
% of ${\tau(\mathcal{D})}$, such paths will be defined in terms of
% answer sets of logic programs axiomatizing the agent's knowledge.
% However, since this knowledge may contains defaults and their direct
% and indirect exceptions, our language of choice is not ASP, but
% CR-Prolog, an extension of ASP especially well suited for
% representing and reasoning with such knowledge.

\begin{defn2}\label{def:pdh}[Program $\Pi(\mathcal{D},\mathcal{H})$]\\
  {\rm Program $\Pi(\mathcal{D},\mathcal{H})$, encoding the system
    description $\mathcal{D}$ and history $\mathcal{H}$ of the domain,
    is obtained by changing the value of constant $n$ in
    $\Pi(\mathcal{D})$ from $1$ to the current step of $\mathcal{H}$
    and adding to the resulting program:
    \begin{itemize}
    \item Observations and actions, i.e., Statements~\ref{eqn:obs}
      and~\ref{eqn:act}, from $\mathcal{H}$.

    \item Encoding of each default, i.e., for every default such as
      Statement~\ref{eqn:default} from $\mathcal{H}$, we add:
      \begin{align} 
        \label{eqn:enc-default} 
        val(f(\bar{x}),y,0) \leftarrow & val(body(d(\bar{x})),0),\\\nonumber &
        \mbox{not } ab(d(\bar{x}))
      \end{align}
      \begin{align}
        \label{eqn:enc-crrule}
        &ab(d(\bar{x})) \rif  val(body(d(\bar{x})),0)%\qquad\qquad \textrm{\% CR rule}
      \end{align} 
      where Statement~\ref{eqn:enc-default} is a simplified version of
      the standard CR-Prolog (or ASP) encoding of a default, and the
      relation $ab(d)$, read as \emph{default $d$ is abnormal}, holds
      when default $d$ is not applicable.
      Statement~\ref{eqn:enc-crrule} is a \emph{consistency restoring}
      (CR) rule, which says that to restore consistency of the program
      one may refrain from applying default $d$. It is an axiom in
      CR-Prolog used to allow indirect exceptions to defaults---it is
      not used unless assuming $f(\bar{x})=y$ leads to a
      contradiction. For more details about CR rules, please
      see~\cite{Gelfond:aibook14}.

    \item Encoding of preference relations. If there are two or more
      defaults with preference relations, e.g.,
      Statements~\ref{def:d1}-\ref{def:pr}, we first add the
      following:
      \begin{align}
        \label{eqn:pref-rule1}
        ab(D_2) \leftarrow & prefer(D_1, D_2), \\ \nonumber &
        val(body(D_1),0), \\\nonumber & \textrm{not }~~ ab(D_1)
      \end{align}
      where $D_1$ and $D_2$ are defaults. Then, we add the following:
      \begin{align}
        \label{eqn:pref-rule2}
        prefer(D_1, D_3) \leftarrow & prefer(D_1, D_2), \\ \nonumber &
        prefer(D_2, D_3)
      \end{align}
      \begin{align}
        \label{eqn:pref-rule3}
        \neg prefer(D, D) 
      \end{align}
      Statement~\ref{eqn:pref-rule1} prevents the applicability of a
      default if another (preferred) default is applicable. The other
      two axioms
      (Statements~\ref{eqn:pref-rule2},~\ref{eqn:pref-rule3}) express
      transitivity and anti-symmetry of the preference relation.

    \item Rules for initial observations, i.e., for every basic fluent
      $f$ and its possible value $y$:
      \begin{align}
        \label{eqn:obs_at_0}
        val(f(\bar{x}),y,0) &\leftarrow obs(rob_1, f(\bar{x})=y,0)
      \end{align}
      \begin{align}\label{eqn:obs_at_0a}
        \neg val(f(\bar{x}),y,0) &\leftarrow obs(rob_1, f(\bar{x})\not=y,0)
      \end{align}
      These axioms say that the initial observations are correct. Among
      other things they may be used to defeat the defaults of
      $\mathcal{H}$.

    \item Assignment of initial values to basic fluents that have not
      been defined by other means. Specifically, the initial value of
      a basic fluent not defined by a default is selected from the
      fluent's range.  To do so, for every initial state default (of
      the form of Statement~\ref{eqn:default}) from $\mathcal{H}$:
      \begin{align}
        \label{eqn:defined-by-default}
        defined\_by\_default(f(\bar{x}))\leftarrow &
        val(body(d(\bar{x})), 0), \\\nonumber & \mbox{not }
        ab(d(\bar{x}))
      \end{align}
      Then, for every basic fluent $f$:
      \begin{align} 
        \label{eqn:disj1} 
        val(f(\bar{x}),y_1,0) \mbox{ or } \dots \mbox{ or }
        val(f(\bar{x}),y_n,0) \leftarrow \mbox{ not }
        defined\_by\_default(f(\bar{x}))
      \end{align}
      where $\{y_1,\dots,y_n\}$ are elements in the range of
      $f(\bar{x})$ not occurring in the head of any initial default of
      $\mathcal{H}$.

    \item A reality check~\cite{Balduccini:TPLP03}:
      \vspace{-1.5em}
      \begin{align} 
        \label{eqn:reality-check} 
        \leftarrow & val(F,Y_1,I),~ obs(rob_1, F=Y_2,I), ~Y_1 \not= Y_2
        %\\\nonumber 
        %\leftarrow & val(F,Y_1,I),~ obs(rob_1, F\neq Y_1, I)
      \end{align}
      which says that the value of a fluent predicted by our program
      shall not differ from its observed value.

%     \item A constraint:
%       \begin{align}
%         \leftarrow obs(R, F=Y, I),~~not~val(can\_be\_observed(R, F, Y), true, I)
%       \end{align}
%       which says that any observation has to match the capabilities
%       of the robot.

    \item And a rule:
      \vspace{-1em}
      \begin{align} 
        \label{eqn:hpd-occur}
        occurs(A,I) \leftarrow hpd(A,I)
      \end{align}
      which establishes the relation between relation $hpd$ of the
      language of recorded histories and relation $occurs$ used in the
      program. Recall that $occurs$ denotes both actual and
      hypothetical occurrences of actions, whereas $hpd$ indicates an
      actual occurrence.
    \end{itemize}
    This completes construction of the program.
}
\end{defn2}

\noindent
We will also need the following terminology in the discussion below.
Let $\mathcal{H}$ be a history of $\mathcal{D}$ and $AS$ be an answer
set of $\Pi(\mathcal{D}, \mathcal{H})$.  We say that a sequence $M =
\langle \sigma_0,a_0,\sigma_1,\dots,\sigma_{n},a_n\rangle$ such that
$\forall i\in [0, n]$:
\begin{s_itemize}
\item $\sigma_i = \{f=y: val(f,y,i) \in AS\}$, 
\item $a_i = \{e : hpd(e,i) \in AS \}$. 
\end{s_itemize}
is \emph{induced} by $AS$. Now we are ready to define semantics of
$\mathcal{H}$.

\begin{defn2}\label{def:model}[Model]\\
  \rm{ A sequence $\langle
    \sigma_0,a_0,\sigma_1,\dots,\sigma_{n},a_n\rangle$ induced by an
    answer set $AS$ of $\Pi(\mathcal{D}, \mathcal{H})$ is called a
    \emph{model} of $\mathcal{H}$ if it is a path of transition
    diagram ${\tau(\mathcal{D})}$.  }
\end{defn2}

\begin{defn2}\label{def:entail}[Entailment]\\
  \rm{ A literal $l$ is true at step $i$ of a path $M = \langle
    \sigma_0,a_0,\sigma_1,\dots,\sigma_{n},a_n\rangle$ of
    ${\tau(\mathcal{D})}$ if $l \in \sigma_i$. We say that $l$ is
    \emph{entailed} by a history $\mathcal{H}$ of $\mathcal{D}$ if $l$
    is true in all models of $\mathcal{H}$.  }
\end{defn2}

\medskip
\noindent
The following proposition shows that for well-founded system
descriptions this definition can be simplified.

\begin{prop2} \label{prop:model-hist}[Answer sets of
  $\Pi(\mathcal{D},\mathcal{H})$ and paths of ${\tau(\mathcal{D})}$]\\
  \rm{If $\mathcal{D}$ is a well-founded system description and
    $\mathcal{H}$ is its recorded history, then every sequence induced
    by an answer set of $\Pi(\mathcal{D}, \mathcal{H})$ is a model of
    $\mathcal{H}$.  }
\end{prop2}
\noindent
The proof of this proposition is in
Appendix~\ref{sec:appendix-prop-hist}. Next, we look at some examples
of histories with defaults.

\noindent
\begin{example2}\label{ex:defaults-again}[Example~\ref{ex:defaults} revisited]\\
  {\rm Let us revisit the histories described in
    Example~\ref{ex:defaults} and show how models of system
    descriptions from this example can be computed using our
    axiomatization $\Pi(\mathcal{D}, \mathcal{H})$ of models of a
    recorded history. We see that models of $\mathcal{H}_a$ are of the
    form $\langle \sigma_0 \rangle$ where $\sigma_0$ is a state of the
    system containing $\{loc(tb_1) = main\_library\}$. Since
    $textbook(tb_1)$ is a static relation, it is true in every state
    of the system. The axiom encoding default $d_1$
    (Statement~\ref{eqn:enc-default}) is not blocked by a CR rule
    (Statement~\ref{eqn:enc-crrule}) or a preference rule
    (Statement~\ref{eqn:pref-rule1}), and the program entails
    $val(loc(tb_1), main\_library,0)$. Thus, $\{loc(tb_1) =
    main\_library\}\in \sigma_0$.
    % (\textcolor{red}{It would be nice to be able to refer to the
    %   last rule of $\Pi(D)$.})

    Now consider $\mathcal{H}_b$ containing $obs(rob_1, loc(tb_1) \not=
    main\_library,0)$. Based on rules for initial observations
    (Statement~\ref{eqn:obs_at_0a}) we have $\neg val(loc(tb_1),
    main\_library,0)$ which contradicts the first default. The
    corresponding CR-rule (Statement~\ref{eqn:enc-crrule}) restores
    consistency by assuming $ab(d_1)$, making default $d_1$
    inapplicable. Default $d_2(tb_1)$, which used to be blocked by a
    preference rule (i.e., $prefer(d_1(tb_1), d_2(tb_1))$), becomes
    unblocked and we conclude that $val(loc(tb_1),aux\_library,0)$.
    Models of $\mathcal{H}_b$ are states of $\tau(\mathcal{D})$ that
    contain $\{loc(tb_1) = aux\_library\}$. A similar argument can be
    used to compute the models of $\mathcal{H}_c$ in
    Example~\ref{ex:defaults}.

    Recall that the last history, $\mathcal{H}_d$, is slightly
    different. It contains $obs(rob_1, loc(tb_1) \not= main\_library, 1)$.
    The current step of $\mathcal{H}_d$ is $1$ and its models are of
    the form $\langle \sigma_0,a,\sigma_1\rangle$.  Since
    $\Pi(\mathcal{D}, \mathcal{H}_d)$ has no rules with an action in
    the head, $a = \{\ \}$. Based on default $d_1$, $\{loc(tb_1) =
    main\_library\}$ should belong to state $\sigma_0$. However, if
    this were true, $\{loc(tb_1) = main\_library\}$ would belong to
    $\sigma_1$ by inertia, which contradicts the observation and the
    reality check axiom creates an inconsistency. This inconsistency
    is resolved by the corresponding CR-rule
    (Statement~\ref{eqn:enc-crrule}) by assuming $ab(d_1)$ in the
    initial state (i.e., at time $0$). Now the default $d_2$ is
    activated and the reasoner concludes $\{loc(tb_1) =
    aux\_library\}$ at time step $0$ and (by inertia) at time step
    $1$.

    To illustrate the use of axioms governing the initial value of a
    basic fluent not defined by a default
    (Statements~\ref{eqn:defined-by-default} and~\ref{eqn:disj1}),
    consider history $\mathcal{H}_e$ in which observations at step $1$
    establish that textbook $tb_1$ is not in any of the default
    locations. An argument similar to that used for $\mathcal{H}_d$
    would allow the reasoner to conclude $ab(d_1(tb_1))$,
    $ab(d_2(tb_1))$, and $ab(d_3(tb_1))$, and
    $defined\_by\_default(loc(tb_1))$ can not be derived.
    Statement~\ref{eqn:disj1} is now used to allow a choice between
    the four locations that form the range of the $loc()$ function.
    The first three are eliminated by observations at step $1$ and we
    thus conclude $val(loc(tb_1),kitchen,0)$, i.e., $\{loc(tb_1) =
    kitchen\} \in \sigma_1$.  Note that if the domain included other
    available locations, we would have additional models of history
    $\mathcal{H}_e$.  }
\end{example2}

% \medskip
% \noindent
% We provide some more examples of models of history.
\begin{example2}\label{ex:models}[Examples of models of history]\\
  {\rm As further examples of models of history, consider a system
    description $\mathcal{D}_a$ with basic boolean fluents $f$ and $g$
    (and no actions), and a history $\mathcal{H}_a$ consisting of:
    \vspace{-1em}
    \begin{align*}
      {\bf initial\ default} \  \neg g\  {\bf if}\  f
    \end{align*}
    The paths of this history consist of states without any
    transitions. Using axiom in Statement~\ref{eqn:disj1}, we see that
    $\{f, \neg g\}$, $\{\neg f, g\}$, and $\{\neg f, \neg g\}$ are
    models of $\langle\mathcal{D}_a, \mathcal{H}_a\rangle$ and
    $\sigma=\{f,g\}$ is not. The latter is not surprising since even
    though $\sigma$ may be physically possible, the agent, relying on
    the default, will not consider $\sigma$ to be compatible with the
    default since the history gives no evidence that the default
    should be violated. If, however, the agent were to record an
    observation $obs(rob_1, g, 0)$, the only states compatible with the
    resulting history ${\mathcal H}_b$ would be $\{f,g\}$ and $\{\neg
    f,g\}$.)
  
    \medskip 
    \noindent 
    Next, we expand our system description $\mathcal{D}_a$ by a basic
    fluent $h$ and a state constraint:
    \begin{align*}
      h\ {\bf if}\ \neg g
    \end{align*}
    In this case, to compute models of a history ${\mathcal H}_c$ of a
    system $\mathcal{D}_b$, where ${\mathcal H}_c$ consists of the
    default in $\mathcal{H}_a$ and an observation $obs(rob_1, \neg h, 0)$,
    we need CR rules.  The models are $\{f,\neg h, g\}$ and $\{\neg f,
    \neg h, g\}$.
    
    \medskip
    \noindent
    Next, consider a system description $\mathcal{D}_c$ with basic
    fluents $f$, $g$, and $h$, the initial-state default, and an
    action $a$ with the following causal law:
    \vspace{-0.5em}
    \begin{equation*}
      a\ {\bf causes} \ h \ {\bf if}\ \neg g
    \end{equation*}
    and a history ${\mathcal H}_d$ consisting of $obs(rob_1, f, 0)$,
    $hpd(a,0)$; $\langle \{f,\neg g, h\}, a, \{f,\neg g, h\} \rangle$
    and $\langle \{f,\neg g, \neg h\}, a, \{f,\neg g, h\} \rangle$ are
    the two models of ${\mathcal H}_d$. Finally, history ${\mathcal
      H}_e$ obtained by adding $obs(rob_1, \neg h, 1)$ to ${\mathcal H}_d$ has
    a single model $\langle \{f,g, \neg h\}, a, \{f,g, h\} \rangle$. The
    new observation is an indirect exception to the initial default,
    which is resolved by the corresponding CR rule. 
  }
\end{example2}

%%%%%-----------------------------------------------------------------------------
\subsection{Reasoning}
\label{sec:arch-hl-reason}
The main reasoning task of an agent with a high level deterministic
system description $\mathcal{D}$ and history $\mathcal{H}$ is to find
a plan (i.e., a sequence of actions\footnote{For simplicity we only
  consider sequential plans in which only one action occurs at a time.
  The approach can be easily modified to allow actions to be performed
  in parallel.}) that would allow it to achieve goal $G$. We assume
that the length of this sequence is limited by some number $h$
referred to as the planning horizon. This is a generalization of a
classical planning problem in which the history consists of a
collection of atoms which serves as a complete description of the
initial state. If history $\mathcal{H}$ has exactly one model, the
situation is not very different from classical planning. The agent
believes that the system is currently in some unique state
$\sigma_n$---this state can be found using
Proposition~\ref{prop:model-hist} that reduces the task of computing
the model of $\mathcal{H}$ to computing the answer set of
$\Pi(\mathcal{D}, \mathcal{H})$. Finding a plan is thus equivalent to
solving a classical planning problem $\mathcal{P}_c$, i.e., finding a
sequence of actions of length not exceeding $h$, which leads the agent
from an initial state $\sigma$ to a state satisfying $G$.  The
ASP-based solution of this planning problem can be traced back to work
described in~\cite{dimopoulos:ecp97,subrahmanian:iclp95}. Also see
program $plan(\mathcal{P}_c,h)$ and Proposition 9.1.1 in Section 9.1
of~\cite{Gelfond:aibook14}, which establish the relationship between
answer sets of this program and solutions of $\mathcal{P}_c$, and can
be used to find a plan to achieve the desired goal. A more subtle
situation arises when $\mathcal{H}$ has multiple models. Since there
are now multiple possible current states, we can either search for a
\emph{possible plan}, i.e. a plan leading to $G$ from at least one of
the possible current states, or for a \emph{conformant} plan, i.e., a
plan that can achieve $G$ independent of the current state.  In this
paper, we only focus on the first option\footnote{an ASP-based
  approach to finding conformant plans can be found
  in~\cite{tu:AIJ11}.}.

\begin{defn2}\label{def:planning}[Planning Problem]\\
  {\rm We define a \emph{planning problem} $\mathcal{P}$ as a tuple
    $(\mathcal{D}, \mathcal{H}, h, G)$ consisting of system
    description $\mathcal{D}$, history $\mathcal{H}$, planning horizon
    $h$ and a goal $G$. A sequence $\langle a_0,\dots,a_{k-1} \rangle$
    is called a \emph{solution} of $\mathcal{P}$ if there is a state
    $\sigma$ such that:
    \begin{itemize}
    \item $\sigma$ is the current state of some model $M$ of
      $\mathcal{H}$; and

    \item $\langle a_0,\dots,a_{k-1} \rangle$ is a solution of
      classical planning problem $\mathcal{P}_c =
      (\mathcal{D},\sigma,G)$ with horizon $h$.
    \end{itemize}
  }
\end{defn2}
% \textcolor{red}{Change n in definition of $\Pi(D)$ to $max\_step$}
%\medskip
\noindent 
To find a solution of $\mathcal{P}$ we consider:
\begin{itemize}
\item CR-Prolog program $Diag =_{def}\Pi^{n}(\mathcal{D},\mathcal{H})$
  with maximum time step $n$ where $n$ is the current step of
  $\mathcal{H}$.

\item ASP program $Classical\_plan$ consisting of:
  \begin{enumerate}
  \item $\Pi^{[n..n+h]}(\mathcal{D})$ obtained from $\Pi(\mathcal{D})$
    by setting $max\_step$ to $n+h$ and sort $step$ to $(n,
    max\_step)$.
  \item Encoding of the goal $f(\bar{x}) = y$ by the rule:
    \begin{align*}
      goal(I) \leftarrow val(f(\bar{x}),y,I)
    \end{align*}

  \item Simple planning module, $PM$, obtained from that in Section
    $9.1$ of~\cite{Gelfond:aibook14} (see statements on page 194) by
    letting time step variable $I$ range between $n$ and $n+h$.
  \end{enumerate}
  
\item Diagnoses Preserving Constraint (DPC):
  \begin{align*}
    \leftarrow Y = count\{X : ab(X)\}, Y > m
  \end{align*}
  where $m$ is the size of abductive support of $Diag$. For any
  program $\Pi$, if $\Pi^{reg}$ is the set of all regular rules of
  $\Pi$ and $\alpha(R)$ is the set of regular rules obtained by
  replacing $\rif$ by $\leftarrow$ in each CR rule in $R$, a
  cardinality-minimal set of CR rules such that $\Pi(R) =_{def}
  \Pi^{reg}\cup \alpha(R)$ is consistent, is called an abductive
  support of $\Pi$\footnote{Although a program may have multiple
    abductive support, they all have the same size due to the
    minimality requirement.}.
\end{itemize}  
Proposition~\ref{prop:plan-reduce} can be used to reduce finding the
solutions of the planning problem $\mathcal{P}$ to:
\begin{enumerate}
\item Computing the size, $m$, of an abductive support of $Diag$.
\item Computing answer sets of CR-Prolog program:
  \begin{align*}
    Plan = Diag \cup Classical\_plan \cup \{DPC\}
  \end{align*}
\end{enumerate}
Based on Proposition~\ref{prop:model-hist}, the first task of finding
the abductive support of $Diag$ can be accomplished by computing a
model of:
\begin{align*}
  Diag \cup \{size(Y) \leftarrow count\{X : ab(X)\} = Y\}
\end{align*}
and displaying atom $size(m)$ from this model. The second task of
reducing planning to compute answer sets is based on the following
proposition that is analogous to Proposition $9.1.1$ in Section $9.1$
of~\cite{Gelfond:aibook14}.

\begin{prop2} \label{prop:plan-reduce}[Reducing planning to computing
  answer sets]\\
  \rm{ Let $\mathcal{P} = (\mathcal{D}, \mathcal{H}, h, G)$ be a
    planning problem with a well-founded, deterministic system
    description $\mathcal{D}$. A sequence $\langle a_0,\dots,a_{k-1}
    \rangle$ where $k < h$ is a solution of $\mathcal{P}$ iff there is
    an answer set $A$ of $Plan$ such that:
    \begin{enumerate}
    \item For any $n < i \leq n+k$, $occurs(a_i,i-1) \in A$,
    \item $A$ contains no other atoms of the form $occur(*,
      i)$\footnote{The ``*'' denotes a wild-card character.} with $i
      \geq n$.
    \end{enumerate} 
  }
\end{prop2}
\noindent
The proof of this proposition is provided in
Appendix~\ref{sec:appendix-prop-plan}. Similar to classical planning,
it is possible to find plans for our planning problem that contain
irrelevant, unnecessary actions. We can avoid this problem by asking
the planner to search for plans of increasing length, starting with
plans of length $1$, until a plan is found. There are other ways to
find minimum-length plans, but we do not discuss them here.

\section{Logician's Domain Representation}
\label{sec:arch-logician}
We are now ready for the first step of our design methodology (see
Section~\ref{sec:methodology}), which is to provide a
coarse-resolution description of the robot's domain in $\mathcal{AL}_d$ along
with a description of the initial state---we re-state this step as
specifying the transition diagram of the logician.

\medskip
\noindent
\begin{center}
  \fbox{%
    \parbox{0.95\textwidth}{%
      1. Specify the transition diagram, $\tau_H$, which will be used
      by the logician for coarse-resolution reasoning, including
      planning and diagnostics.  } }
\end{center}
\medskip

\noindent
This step is accomplished by providing the signature and $\mathcal{AL}_d$ axioms
of system description $\mathcal{D}_H$ defining this diagram. We will
use standard techniques for representing knowledge in action
languages, e.g.,~\cite{Gelfond:aibook14}.  We illustrate this process
by describing the domain representation for the office domain
introduced in Example~\ref{ex:illus-example}.

%\medskip

\begin{figure*}[tb]
  \begin{center}
    \subfigure[0.45\textwidth][Some state transitions] {
      \includegraphics[width=0.45\textwidth]{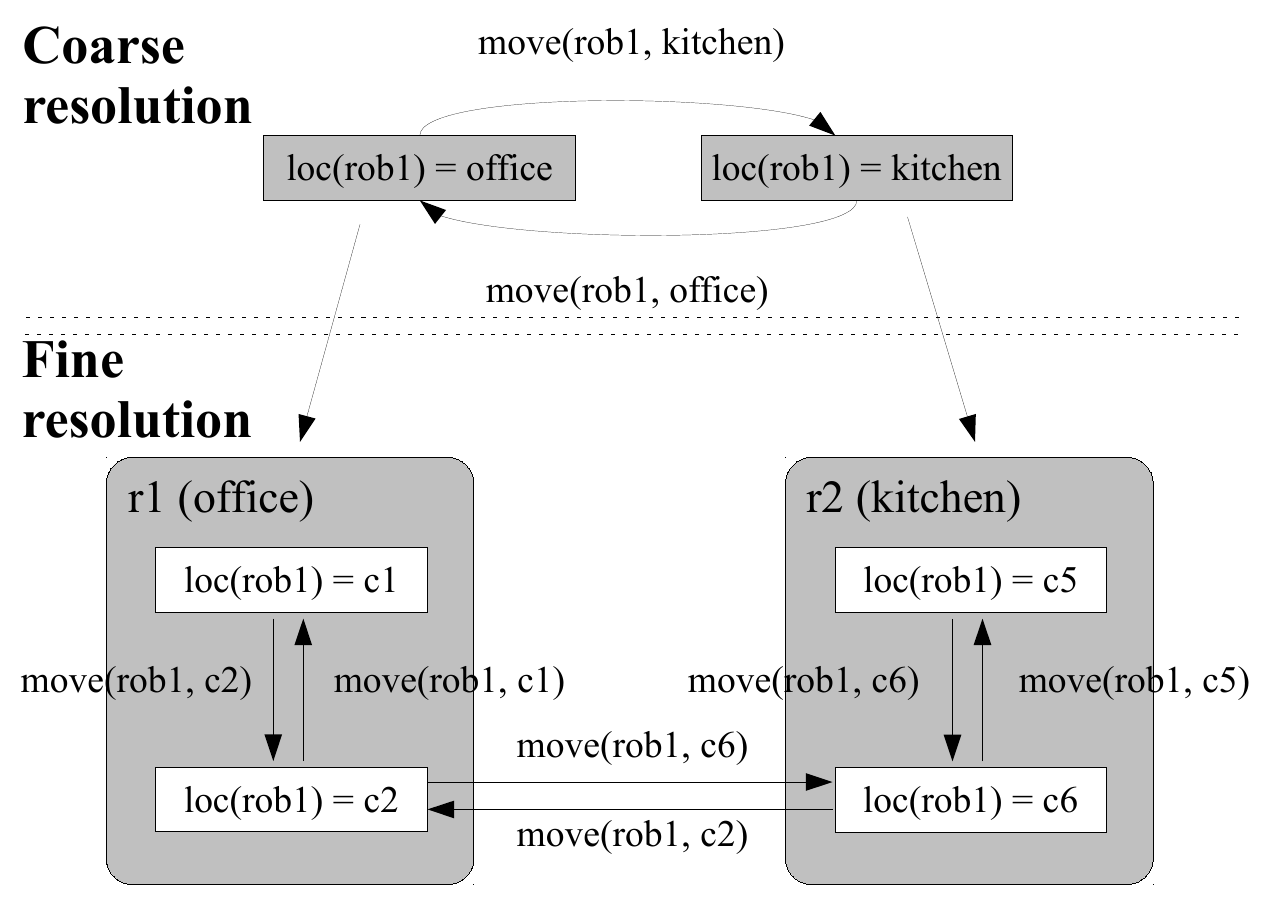}
      \label{fig:refine-state-trans}
    }%\hspace{0.1in}
    \subfigure[0.43\textwidth][Different resolutions] {
      \includegraphics[width=0.43\textwidth]{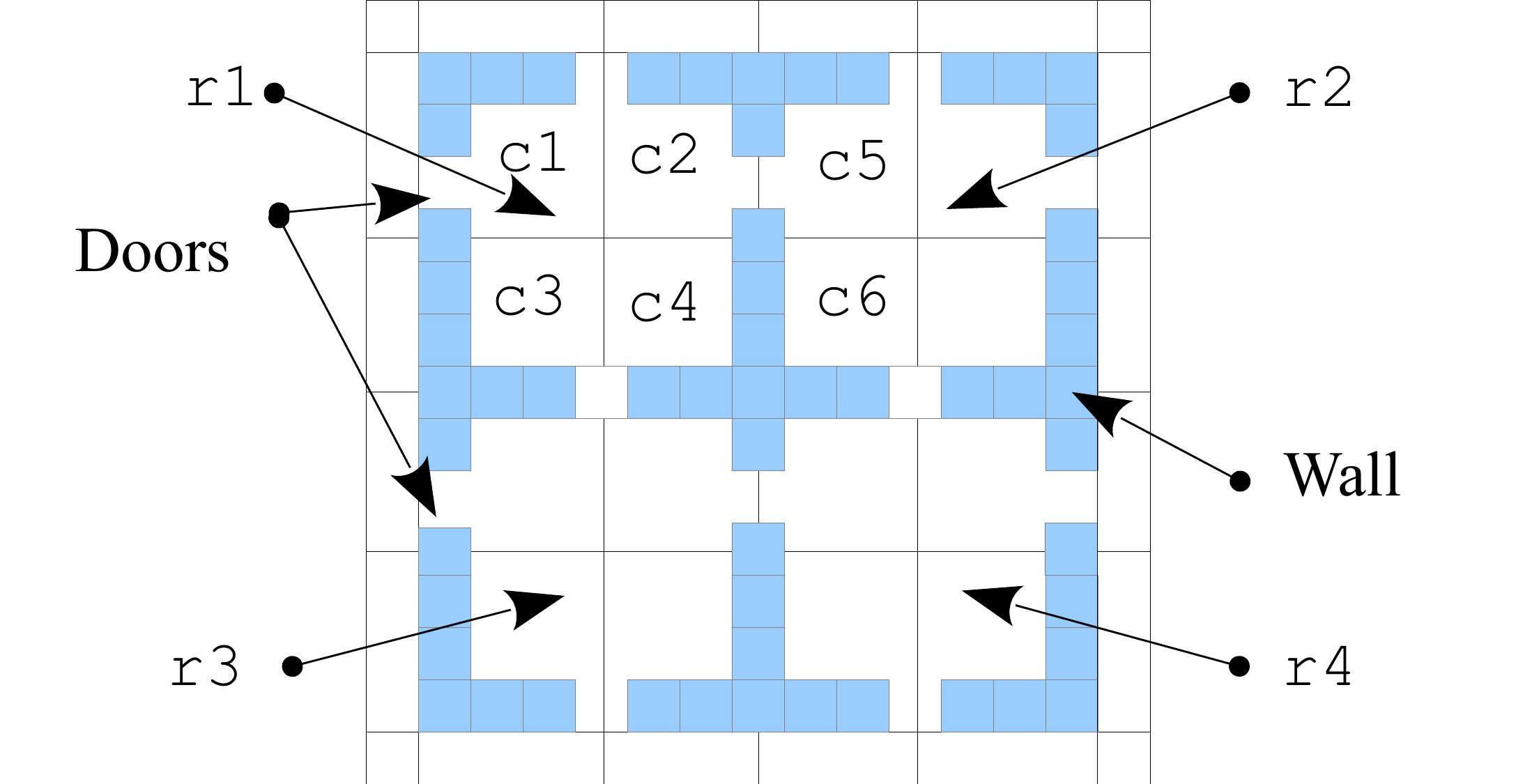}
      \label{fig:refine-gridmap}
    }
  \end{center}
  \vspace{-2em}
  \caption{(a) Illustration of state transitions for specific $move$
    actions in our illustrative (office) domain, viewed at coarse
    resolution and at fine resolution; and (b) A closer look at
    specific places brings into focus the corresponding rooms and grid
    cells in those rooms.}
  \label{fig:refine}
\end{figure*}

%%%%%-----------------------------------------------------------------------------
\begin{example2}\label{ex:logician-example}[Logician's domain representation]\\
  {\rm The system description $\mathcal{D}_H$ of the domain in
    Example~\ref{ex:illus-example} consists of a sorted signature
    ($\Sigma_H$) and axioms describing the transition diagram
    $\tau_H$.  $\Sigma_H$ defines the names of objects and functions
    available for use by the logician. Building on the description in
    Example~\ref{ex:illus-example}, $\Sigma_H$ has an ontology of
    sorts, i.e., sorts such as $place$, $thing$, $robot$, and
    $object$, which are arranged hierarchically, e.g., $object$ and
    $robot$ are subsorts of $thing$, and $textbook$ and $cup$ are
    subsorts of $object$. The statics include a relation $next\_to :
    place \times place \rightarrow boolean$, which holds iff two
    places are next to each other. This domain has two basic fluents
    that are subject to the laws of inertia: $loc : thing \rightarrow
    place$, $in\_hand : robot \times object \rightarrow boolean$.  For
    instance, the $loc(Th) = Pl$ if thing $Th$ is located at place
    $Pl$, and the value of $in\_hand(R,Ob)$ is $true$ if robot $R$ is
    holding object $Ob$. In this domain, the basic fluents are
    observable. % Also, for every robot $rob$, the domain has two
%     defined fluents: $can\_be\_observed_{loc}: robot\times thing\times
%     place \rightarrow boolean$ and
%     $can\_be\_observed_{\{rob,in\_hand\}}: robot\times object
%     \rightarrow boolean$.

    The domain has three actions: $move(robot, place)$, $grasp(robot,
    object)$, and $putdown(robot, object)$. The domain dynamics are
    defined using axioms that consist of causal laws such as:
    \begin{subequations}
      \label{eqn:logician-causal}
      \begin{align}
        &move(R, Pl)~~\mathbf{causes}~~loc(R) = Pl \\ 
        &grasp(R, Ob)~~\mathbf{causes}~~in\_hand(R, Ob) \\ 
        &putdown(R, Ob)~~\mathbf{causes}~~\neg in\_hand(R, Ob) 
      \end{align}
    \end{subequations}
    state constraints such as:
    \vspace{-1em}
    \begin{subequations}
      \label{eqn:logician-constraint}
      \begin{align}
        &loc(Ob) = Pl~~\mathbf{if}~~loc(R) = Pl,~~in\_hand(R, Ob)\\
        &next\_to(P1, P2)~~\mathbf{if}~~next\_to(P2, P1) % \\
%         &can\_be\_observed(R, loc(Th), Pl)~~\mathbf{if}~~loc(R) = Pl
      \end{align}
    \end{subequations}
    and executability conditions such as:
    % \vspace{-0.5em}
    \begin{subequations}
      \label{eqn:logician-executability}
      \begin{align}
        &\mathbf{impossible}~~move(R, Pl)~~\mathbf{if}~~loc(R)=Pl \\
        &\mathbf{impossible}~~move(R, Pl_2) ~~\mathbf{if}~~
        loc(R)=Pl_1,~\neg next\_to(Pl_1,Pl_2) \\
        &\mathbf{impossible}~~A_1,~A_2~~\mathbf{if}~~A_1\neq A_2\\
        &\mathbf{impossible}~~grasp(R, Ob) ~~\mathbf{if}~~
        loc(R)=Pl_1,~loc(Ob)=Pl_2,~Pl_1\neq Pl_2 \\
        &\mathbf{impossible}~~grasp(R, Ob)~~\mathbf{if}~~ in\_hand(R,
        Ob) \\
        &\mathbf{impossible}~~putdown(R, Ob)~~\mathbf{if}~~\neg
        in\_hand(R, Ob)
      \end{align}
    \end{subequations}
    The part of $\Sigma_H$ described so far, the sort hierarchy and
    the signatures of functions, is unlikely to undergo substantial
    changes for any given domain. However, the last step in the
    constructions of $\Sigma_H$ is likely to undergo more frequent
    revisions---it populates the sorts of the hierarchy with specific
    objects; e.g $robot = \{rob_1\}$, $place = \{r_1,\dots, r_n\}$
    where $r$s are rooms, $textbook = \{tb_1,\dots\, tb_m\}$,
    $kitchenware = \{cup_1, cup_2, plate_1, plate_2\}$ etc. Ground
    instances of the axioms are obtained by replacing variables by
    ground terms from the corresponding sorts.

    The transition diagram $\tau_H$ described by $\mathcal{D}_H$ is
    too large to depict in a picture. The top part of
    Figure~\ref{fig:refine-state-trans} shows the transitions of
    $\tau_H$ corresponding to a $move$ between two places. The only
    fluent shown there is the location of the robot $rob_1$---the
    values of other fluents remain unchanged and are not shown here.
    The actions of this coarse-resolution transition diagram $\tau_H$
    of the logician, as described above, are assumed to be
    deterministic. Also, the values of coarse-resolution fluents are
    assumed to be known at each time step. These assumptions allow the
    robot to do fast, tentative planning and diagnostics necessary for
    achieving its assigned goals.

    The domain representation described above should ideally be tested
    extensively. This can be done by including various recorded
    histories of the domain, which may include histories with
    prioritized defaults (Example~\ref{ex:defaults}), and using the
    resulting programs to solve various reasoning tasks.  }
\end{example2}

\medskip
\noindent
The logician's model of the world thus consists of the system
description $\mathcal{D}_H$ (Example~\ref{ex:logician-example}), and
recorded history $\mathcal{H}$ of initial state defaults
(Example~\ref{ex:defaults}), actions, and observations. The logician
achieves any given goal by first translating the model (of the world)
to an ASP program $\Pi(\mathcal{D}_H, \mathcal{H})$, as described in
Sections~\ref{sec:arch-ald},~\ref{sec:arch-hl-hist}, and expanding it
to include the definition of goal and suitable axioms, as described at
the end of Section~\ref{sec:arch-hl-reason}. For planning and
diagnostics, this program is passed to an ASP solver---we use SPARC,
which expands CR-Prolog and provides explicit constructs to specify
objects, relations, and their sorts~~\cite{balai:lpnmr13}. Please see
\texttt{example4.sp} at \url{https://github.com/mhnsrdhrn/refine-arch}
for the SPARC version of the complete program.  The solver returns the
answer set of this program. Atoms of the form: \vspace{-0.5em}
\begin{align*}
  occurs(action, step)
\end{align*}
belonging to this answer set, e.g., $occurs(a_1, 1), \ldots,
occurs(a_n, n)$, represent the shortest plan, i.e., the shortest
sequence of abstract actions for achieving the logician's goal. Prior
research results in the theory of action languages and ASP ensure that
the plan is provably correct~\cite{Gelfond:aibook14}. In a similar
manner, suitable atoms in the answer set can be used for diagnostics,
e.g., to explain unexpected observations by triggering suitable CR
rules.

% \textcolor{red}{?? no exogenous actions. Need to connect/combine with
%   the story at the end of section 5}.

%%%%%-----------------------------------------------------------------------------
%%%%%-----------------------------------------------------------------------------
\section{Refinement, Zoom and Randomization}
\label{sec:arch-refzoomrand}
For any given goal, each abstract action in the plan created by
reasoning with the coarse-resolution domain representation is
implemented as a sequence of concrete actions by the statistician. To
do so, the robot probabilistically reasons about the part of the
fine-resolution transition diagram relevant to the abstract action to
be executed.  This section defines refinement, randomization, and the
zoom operation, which are necessary to build the fine-resolution
models for such probabilistic reasoning, along with the corresponding
steps of the design methodology.

%%%%%-----------------------------------------------------------------------------
\subsection{Refinement}
\label{sec:arch-refine}
Although the representation of a domain used by a logician specifies
fluents with observable values and assumes that all of its actions are
executable, the robot may not be able to directly make some of these
observations or directly execute some of these actions. For instance,
a robot may not have the physical capability to directly observe if it
is located in a given room, or to move in a single step from one room
to another. We refer to such actions that cannot be executed directly
and fluents that cannot be observed directly as \emph{abstract};
actions that can be executed and fluents that can be observed directly
are, on the other hand, referred to as \emph{concrete}. The second
step of the design methodology (see Section~\ref{sec:methodology})
requires the designer to refine the coarse-resolution transition
diagram $\tau_H$ of the domain by including information needed to
execute the abstract actions suggested by a logician, and to observe
values of relevant abstract statics and fluents. This new transition
diagram $\tau_L$ defined by system description $\mathcal{D}_L$, is
called the \emph{refinement} of $\tau_H$.  Its construction may be
imagined as the designer taking a closer look at the domain through a
magnifying lens.  Looking at objects of a sort $s$ of $\Sigma_H$ at
such finer resolution may lead to the discovery of parts of these
objects and their attributes previously abstracted out by the
designer. Instead of being a single entity, a room may be revealed to
be a collection of cells with some of them located next to each other,
a cup may be revealed to have parts such as handle and base, etc. If
such a discovery happens, the sort $s$ and its objects will be said to
have been \emph{magnified}, and the newly discovered parts are called
the \emph{components} of the corresponding objects. In a similar
manner, a function $f: s_1,\dots,s_n \rightarrow s_0$ from $\Sigma_H$
is affected by the increased resolution or magnified if:
\begin{itemize}
\item It is an abstract action or fluent and hence can not be executed
  or observed directly by robots; and
\item At least one of $s_0,\dots,s_n$ is magnified.
\end{itemize}
In the signature $\Sigma_L$ of the fine-resolution model $\tau_L$, the
newly discovered components of objects from a sort $s$ of $\Sigma_H$
form a new sort $s^*$, which is called the \emph{fine-resolution
  counterpart} of $s$. For instance, in our example domain, $place^*$,
which is a collection of grid cells $\{c_1,\ldots, c_n\}$, is the
fine-resolution counterpart of $place$, which is a collection of
rooms, and $object^*$ may be the collection of parts of cups. A vector
$\bar{s}^*$ is a fine-resolution counterpart of $\bar{s}$ with respect
to magnified sorts $s_{i_1},\dots,s_{i_k}$ (with $k > 0$) if it is
obtained by replacing $s_{i_1},\dots,s_{i_k}$ by
$s^{*}_{i_1},\dots,s^{*}_{i_k}$.  Every element $\bar{x}$ of
$\bar{s}^*$ is obtained from the unique\footnote{For simplicity we
  assume that no object can be a component of two different objects.}
element $\bar{u}$ of $\bar{s}$ by replacing $u_{i_1},\dots,u_{i_k}$
from $s_{i_1},\dots,s_{i_k}$ by their components. We say that
$\bar{u}$ is the \emph{generator} of $\bar{x}$ in $\bar{s}^*$ and
$\bar{x}$ is a fine-resolution counterpart of $\bar{u}$. A function
$f^*$ with signature $\bar{s}^*$ is called the fine-resolution
counterpart of a magnified function $f$ with respect to $\langle
s_{i_1},\dots, s_{i_k}\rangle$ if for every $\langle u_1,\dots,u_n,v
\rangle \in \bar{s}$, $f(u_1,\dots,u_n) = v$ iff there is a
fine-resolution counterpart $\langle x_1,\dots,x_n,y \rangle \in
\bar{s}^*$ of $\langle u_1,\dots,u_n,v \rangle$ such that
$f^*(x_1,\dots,x_n) = y$.  For instance, fluents $loc^* : thing
\rightarrow place^*$, $loc^* : object^* \rightarrow place^*$, and
$loc^* : object^* \rightarrow place$ are fine-resolution counterparts
of $loc$ with respect to $\langle place \rangle $, $\langle object,
place \rangle$ and $\langle object \rangle$ respectively; and action
$move^* : robot \times place^* \rightarrow boolean$ is the
fine-resolution counterpart of $move: robot \times place \rightarrow
boolean$ with respect to $place$. In many interesting domains, some
fine-resolution counterparts can be used to execute or observe
magnified functions of $\Sigma_H$, e.g., an abstract action of moving
to a neighbouring room can be executed by a series of moves to
neighbouring cells. We describe other such examples later in this
section.

% The signature $\Sigma_L$ of $\tau_L$ consists of:
% \begin{itemize}
% \item All elements of signature $\Sigma_H$. 
% \item Names of components of objects from $\Sigma_H$.

% \item A new sort $s^*$ for every sort $s$ of $\Sigma_H$; $s^*$
%   consists of components of elements of $s$.

% \item One or more symbols $f^* : \bar{U}^* \rightarrow V^*$ for every
%   function $f : \bar{U} \rightarrow V$ in $\Sigma_H$. The declaration
%   of $f^*$ in $\Sigma_L$ is the same as declaration of $f$ in
%   $\Sigma_H$.
% \end{itemize}
% Here, $s^*$ and $f^*$ are referred to as \emph{fine-resolution
%   counterparts} of $s$ and $f$ respectively. 
% In a similar manner, $loc:thing\rightarrow place$ is replaced by
% $loc^*: thing\rightarrow place^*$ in $\Sigma_L$. 
% Note that we have assumed here (for ease of explanation) that all
% sorts and functions of $\Sigma_H$ are magnified to create $\Sigma_L$,
% which is not always true, as described later in this section.

We now define the refinement of a transition diagram. We do so in two
steps. We first define a notion of \emph{weak refinement} that does
not consider the robot's ability to observe the values of domain
fluents. We then introduce our theory of observations, and define a
notion of \emph{strong refinement} (or simply refinement) that
includes the robot's ability to observe the values of domain fluents.

%%%%%-----------------------------------------------------------------------------
\subsubsection{Weak Refinement}
\label{sec:arch-refine-weak}
We introduce some terminology used in the definition below.  Let
signature $\Sigma_1$ be a subsignature of signature $\Sigma_2$ and let
$\sigma_1$ and $\sigma_2$ be interpretations over these signatures. We
say that $\sigma_2$ is an \emph{extension} of $\sigma_1$ if $\sigma_2
|_{\Sigma_1} = \sigma_1$\footnote{As usual $f |_B$ where $f$ is a
  function with domain $A$ and $ B \subset A$ denotes the restriction
  of $f$ on $B$.}.

\begin{defn2}\label{def:weak-refinement}[Weak refinement of $\tau_H$]\\
  A transition diagram $\tau_L$ over $\Sigma_L$ is called a \emph{weak
    refinement} of $\tau_H$ if:
  \begin{enumerate}
  \item For every state $\sigma^\diamond$ of $\tau_L$, the collection
    $\sigma^\diamond |_{\Sigma_H}$ of atoms of $\sigma^\diamond$
    formed by symbols from $\Sigma_H$ is a state of $\tau_H$.
  
  \item For every state $\sigma$ of $\tau_H$, there is a state
    $\sigma^\diamond$ of $\tau_L$ such that $\sigma^\diamond$ is an
    extension of $\sigma$.

  \item For every transition $T = \langle \sigma_1,a^H,\sigma_2
    \rangle$ of $\tau_H$, if $\sigma^{\diamond}_1$ and
    $\sigma^{\diamond}_2$ are extensions of $\sigma_1$ and $\sigma_2$
    respectively, then there is a path $P$ in $\tau_L$ from
    $\sigma^{\diamond}_1$ to $\sigma^{\diamond}_2$ such that:
    \begin{itemize}
    \item actions of $P$ are concrete, i.e., directly executable by
      robots; and
    \item $P$ is \emph{pertinent} to $T$, i.e., all states of $P$ are
      extensions of $\sigma_1$ or $\sigma_2$.
    \end{itemize}
  \end{enumerate}
\end{defn2}
%\medskip 
\noindent
We are now ready to construct the fine-resolution system description
$\mathcal{D}_{L, nobs}$ corresponding to the coarse-resolution system
description $\mathcal{D}_H$ for our running example
(Example~\ref{ex:logician-example}). This construction does not
consider the robot's ability to observe the values of domain fluents
(hence the subscript ``nobs''). We start with the case in which the
only magnified sort in $\Sigma_H$ is $place$. The signature $\Sigma_L$
will thus contain three fine-resolution counterparts of functions from
$\Sigma_H$: (i) basic fluent $loc^* : thing \rightarrow place^*$; (ii)
action $move^* : robot \rightarrow place^*$; and (iii) defined static
$next\_to^* : place^* \times place^* \rightarrow boolean$. We assume
that $loc^*$ and $next\_to^*$ are directly observable and $move^*$ is
executable by the robot. These functions ensure indirect observability
of $loc$ and $next\_to$ and indirect executability of $move$. Although
this construction is domain dependent, the approach is applicable to
other domains.

\medskip
\noindent
\begin{center}
  \fbox{%
    \parbox{0.95\textwidth}{%
      2. Constructing the fine-resolution system description
      $\mathcal{D}_L$ corresponding to the coarse-resolution
      system description $\mathcal{D}_H$.\\
      \noindent 
      (a) Constructing $\mathcal{D}_{L,nobs}$. } }
\end{center}
\medskip
\noindent
To construct signature $\Sigma_{L,nobs}$:
\begin{enumerate}
\item Preserve all elements of signature $\Sigma_H$.

  \smallskip
  \noindent
  In our running example, this includes sorts $thing$, $place$,
  $robot$, $cup$ etc, object constants $rob_1$, $kitchen$,
  $of\!\!fice$, $cup_1$, $tb_1$ etc, static $next\_to(place, place)$,
  fluents $loc: thing \to place$ and $in\_hand: robot\times place
  \rightarrow boolean$, and actions $move(robot, place)$,
  $grasp(robot, object)$ and $putdown(robot, object)$.

% \item Names of components of objects from $\Sigma_H$.
  
%   \smallskip
%   \noindent
%   For instance, in our running example, we introduce grid cells
%   $\{c_1,\ldots, c_n\}$, which is a component of $place$.

\item Introduce a new sort $s^*$ for every sort $s$ of $\Sigma_H$ that
  is magnified by the increase in resolution, with $s^*$ consisting of
  components of elements of $s$. Add $s^*$ to the sort hierarchy as a
  sibling of $s$.  Also, for each abstract function $f$ magnified by
  the increase in resolution, introduce appropriate fine-resolution
  counterparts that support the execution or observation of $f$ at the
  fine-resolution. % Also change the original abstract fluents and
%   statics to be defined fluents.

  \medskip
  \noindent
  In our example, we introduce the sort $place^*$ as the
  fine-resolution counterpart of $place$, and object constants
  $c_1,\ldots, c_n$ of sort $place^*$ that are grid cells; no new sort
  is introduced for the sort $object$.  Also, $\Sigma_{L, nobs}$
  includes new static relation $next\_to^*(place^*, place^*)$, new fluent
  $loc^*: thing \to place^*$, and new action $move^*(robot, place^*)$,
  but no new symbols corresponding to $in\_hand$ or $grasp$.

\item Introduce static relations $component(O^*, O)$, which hold iff
  object $O^*$ of sort $s^*$ is a component of magnified object $O$ of
  sort $s$ of $\Sigma_H$. \emph{These relations are domain dependent
    and need to be provided by the designer}.

  \medskip
  \noindent 
  Continuing with our running example, we introduce the static
  relation:
  \begin{align*}
    &component : place^* \times place \rightarrow boolean
    %&component : object^* \times object \rightarrow boolean
  \end{align*}
  where $component(c,r)$ is true iff cell $c$ is part of room $r$.  

\end{enumerate}

\medskip
\noindent
Next, to construct the axioms of $\mathcal{D}_{L, nobs}$:
\begin{enumerate}
\item For each axiom of $\mathcal{D}_H$, if it contains any abstract
  functions, replace them by their fine-resolution counterparts and
  make these functions' variables range over appropriate sorts
  required by these counterparts.

  \medskip
  \noindent 
  In our running example, all occurrences of the functions $loc: thing
  \rightarrow place$, $next\_to(place, place)$, and $move(robot,
  place)$ in the axioms of $\mathcal{D}_H$ are replaced by $loc: thing
  \rightarrow place^*$, $next\_to^*(place^*, place^*)$ and
  $move^*(robot, place^*)$ respectively. At the same time,
  $in\_hand(robot, object)$, $grasp(robot, object)$, and
  $putdown(robot, object)$ are functions that remain unchanged. This
  results in $\mathcal{D}_{L, nobs}$ having causal laws:
  \begin{subequations}
    \label{eqn:refine-causal-specific}
    \begin{align}
      &move^*(R, C) ~~\mathbf{causes}~~ loc^*(R) = C\\
      &grasp(R, O) ~~\mathbf{causes} ~~in\_hand(R, O)\\
      &putdown(R, O) ~~\mathbf{causes}~~ \neg in\_hand(R, O)
    \end{align}
  \end{subequations} 
  state constraints: 
  \begin{subequations}
    \label{eqn:refine-constraint-specific}
    \begin{align}
      &loc^*(O)=C ~~\mathbf{if}~~ loc^*(R)= C, ~in\_hand(R, O)\\
      &next\_to^*(C_2, C_1) ~~\mathbf{if}~~ next\_to^*(C_1, C_2)
    \end{align}
  \end{subequations}
  and executability conditions such as:
  \begin{subequations}
    \label{eqn:refine-executability-specific}
    \begin{align}
      &\mathbf{impossible}~~ move^*(R, C) ~~\mathbf{if}~~ loc^*(R)=C \\
      &\mathbf{impossible}~~ move^*(R, C_2) ~~\mathbf{if}~~ loc^*(R)=
      C_1, ~\neg next\_to^*(C_1,C_2) \\
      &\mathbf{impossible}~~ grasp(R, O) ~~\mathbf{if}~~ loc^*(R)=
      C_1, ~loc^*(O)=C_2, ~C_1\neq C_2 \\
      &\mathbf{impossible}~~putdown(R, O) ~~\mathbf{if}~~ \neg
      in\_hand(R, O)
    \end{align}
  \end{subequations}
  where $C$, $C_1$, and $C_2$ are grid cells.

\item Introduce \emph{bridge axioms}, i.e., axioms relating the
  coarse-resolution functions and their fine-resolution counterparts.
  These axioms have the form:
  \begin{align}
    \label{eqn:refine-bridge-general}
    f(X_1,\dots,X_m)=Y~~\mathbf{if}~~&component(C_1,X_1),~\ldots~,
    ~component(C_m,X_m),~component(C,Y),\\\nonumber
    &f^*(C_1,\dots,C_m)=C
  \end{align}
  In our running example, we have:
  \begin{subequations}
    \label{eqn:refine-bridge-specific}
    \begin{align}
      &loc(Th) = P~~ {\bf if}~~component(C, P),~loc^*(Th)=C\\
      &next\_to(P_1, P_2) ~~ \mathbf{if}~~ component(C_1, P_1),
      ~component(C_2, P_2),~ next\_to^*(C_1, C_2)
    \end{align}
  \end{subequations}
  \emph{These axioms are domain dependent and need to be provided by
    the designer}.
\end{enumerate}
This completes the construction of $\mathcal{D}_{L, nobs}$ for our
running example.

\medskip
\noindent
To illustrate the robot's reasoning with $\mathcal{D}_{L, nobs}$,
consider a fine-resolution state $\delta_1$ in which the robot is in
cell $c_2$ of the $of\!\!fice$, i.e., $(loc^*(rob_1)=c_2) \in
\delta_1$.  If $\delta_1$ is a fine-resolution counterpart of a
coarse-resolution state $\sigma_1$, then $(loc(rob_1) = of\!\!fice)
\in\sigma_1$ because the bridge axiom in
Statement~\ref{eqn:refine-bridge-specific}(a) infers $loc(rob_1) =
of\!\!fice$ from $loc^*(rob_1) = c_2$. Next, consider the robot's move
from $\delta_1$, with book $tb_1$ in its hand, to a cell $c_5$ in the
$kitchen$.  If $\delta_2$ is the resultant fine-resolution state,
$(loc^*(rob_1) = c_5)\in \delta_2$ based on
Statement~\ref{eqn:refine-causal-specific}(a), and $(loc^*(tb_1) =
c_5) \in \delta_2$ based on
Statement~\ref{eqn:refine-constraint-specific}(a). Now, if $\delta_2$
is a fine-resolution counterpart of a coarse-resolution state
$\sigma_2$, then based on the bridge axiom in
Statement~\ref{eqn:refine-bridge-specific}(a), $(loc(rob_1) =
kitchen)\in \sigma_2$ and $(loc(tb_1) = kitchen)\in \sigma_2$.

\medskip
\noindent
The following proposition says that $\mathcal{D}_{L, nobs}$ as
constructed above is a weak refinement of $\mathcal{D}_H$.
\begin{prop2}\label{prop:weak-refinement}[Weak Refinement]\\
  {\rm Let $\mathcal{D}_H$ and $\mathcal{D}_{L, nobs}$ be the
    coarse-resolution and fine-resolution system descriptions from our
    running example.  Then $\tau_{L, nobs}$ is a weak refinement of
    $\tau_H$.  
  }
\end{prop2}
\noindent
The proof of this proposition is in
Appendix~\ref{sec:appendix-prop-weak-refine}. 
% and it uses the following property. If $\mathcal{D}$ over signature
% $\Sigma$ is a well-founded system description defining transition
% diagram $\tau$, then an interpretation $\delta$ of $\Sigma$ is a
% state of $\tau_H$ iff:
% \begin{itemize}
% \item $\delta$ satisfies constraints of $\mathcal{D}$; and
% \item For every defined fluent $f$ of $\Sigma$, $f(\bar{u})\in \delta$
%   iff there is an axiom defining $f(\bar{u})$ whose body is satisfied
%   by $\delta$.
% \end{itemize}
Although the statement of the proposition and its proof are provided
here for our example domain, this approach can also be used to
construct $\mathcal{D}_{L, nobs}$ and establish weak refinement in
many other robotics domains.

%%%%%-----------------------------------------------------------------------------
\subsubsection{Theory of Observations}
\label{sec:arch-refine-observe}
The definition of weak refinement does not take into account the
robot's ability to observe the values of fluents in the domain.  This
ability to observe plays an important role in updating beliefs and
monitoring the results of action execution in the fine-resolution.
Recall that abstract fluents and statics are indirectly observable and
the concrete fluents and statics are directly observable. In our
running example, the observation of a thing being in a room can be
inferred by checking if the thing can be observed in some cell of this
room. Also, the robot has to monitor its movement between a series of
neighboring cells when it attempts to execute the abstract action of
moving to a neighbouring room. In this section, we introduce a
\emph{Theory of Observations} that supports this ability. This theory
is used in conjunction with any given system description
$\mathcal{D}_{L, nobs}$ as follows.
\begin{enumerate}
\item Expand $\Sigma_{L, nobs}$:
  \begin{itemize}
  \item For every directly observable function $f$, include the
    actions:
    \begin{align*}
      test_{f}: robot\times dom(f) \times range(f) \rightarrow boolean
    \end{align*}
    For $y \in range(f)$, this action checks if the value of $f$ is
    $y$ in a given state. For readability, we will sometimes abuse
    notation and write this action as $test(R, F, Y)$.

    \medskip
    \noindent
    In our example domain, we include an action such as
    $test_{loc^*}(rob_1, O, C)$ for the robot to check if the location
    of an object $O$ is a particular cell $C$.

  \item For every (directly or indirectly) observable function $f$,
    include the basic knowledge fluent:
    \begin{align*}
      observed_{f} : robot\times dom(f) \times range(f) \rightarrow
      \{true,false,undet\}
    \end{align*}
    where the outcome \emph{undet} stands for ``undetermined''. For
    every $x \in dom(f)$ and $y \in range(f)$, the value of
    $observed_f(rob_1, x, y)$ is the result of the most recent
    execution of action $test_f(rob_1, x, y)$. Initially, the value is
    set to \emph{undet}.  After $test_f(rob_1, x, y)$ is executed at
    least once, the value becomes (and remains) boolean.  It is
    \emph{true} if the most recent test returned \emph{true} and
    \emph{false} otherwise.

%     For $y \in range(f)$, the value of this fluent is initially set to
%     \emph{undet} (i.e., ``undetermined''). The value is set of
%     \emph{true} (\emph{false}) if executing an appropriate $test$
%     action determines that $f=y$ ($f=y_1, y_1\not=y$). Also, the value
%     of this fluent is the value determined by the most recent
%     execution of the corresponding $test$ action. For readability, we
%     will abuse notation and write this fluent as $observed(R, F, Y)$.

    \medskip
    \noindent
    In our example domain, we have basic knowledge fluents such as:
    \begin{align*}
      &observed_{loc^*}(rob_1, O, C)\\
      &observed_{loc}(rob_1, O, P)
    \end{align*}

  \item For every indirectly observable function $f$, introduce
    observation-related, domain-dependent defined fluent, as described
    in Statement~\ref{eqn:coarse-observable}:
    \begin{align*}
      observable_{f}: robot\times dom(f) \times range(f) \to boolean
    \end{align*}
    Also, for every directly observable domain function $f$, introduce
    observation-related, domain-dependent defined fluents:
    \begin{align}
      \label{eqn:fine-can-be-observed}
      can\_be\_observed_{f} : robot\times dom(f) \times range(f)
      \rightarrow boolean
    \end{align}
    These fluents will be used to describe conditions for the
    observability of the corresponding functions.  \emph{These domain
      dependent fluents need to be defined by the designer}.
  \end{itemize}

\item Expand axioms of $\mathcal{D}_{L, nobs}$ by axioms that model
  the robot's ability to observe.
  \begin{itemize}
  \item Introduce causal laws to describe the effect of the
    knowledge-producing action $test_{f^*}(R, \bar{X}, Y)$ on the
    fine-resolution basic fluent $f^*$:
    \begin{align}
      \label{eqn:refine-test}
      test_{f^*}(R, \bar{X}, Y) &~~\mathbf{causes}~~ observed_{f^*}(R,
      \bar{X}, Y)=true ~~{\bf if }~~ f^*(\bar{X}) = Y \\\nonumber
      test_{f^*}(R, \bar{X}, Y) &~~\mathbf{causes}~~ observed_{f^*}(R,
      \bar{X}, Y) = false ~~{\bf if }~~ f^*(\bar{X}) =Y_1, ~Y_1\not=Y
    \end{align} 
    Also introduce the executability condition:
    \begin{align}
      \label{eqn:refine-test-impossible}
      {\bf impossible}~~test_{f^*}(R, \bar{X}, Y)~~{\bf if}~~\lnot
      can\_be\_observed_{f^*}(R, \bar{X}, Y)
    \end{align}
    where $\bar{X}$ represents the domain of $f^*$.
    
    \medskip
    \noindent
    In our running example, if robot $rob_1$ located in cell $c$
    checks the presence or absence of an object $o$,
    $observed_{loc^*}(rob_1, o, c)$ will be \emph{true} iff $o$ is in
    $c$ during testing; it will be \emph{false} iff $o$ is not in $c$.
    These values will be preserved by inertia axioms until the state
    is observed to have changed when the same cell is tested again. If
    the robot has not yet tested a cell $c$ for an object $o$, the
    value of $observed_{loc^*}(rob_1, o, c)$ remains \emph{undet}.

  \item Introduce axioms for domain-dependent defined fluents
    describing the ability of the robot to sense the values of
    directly and indirectly observable functions.

    \medskip
    \noindent
    In our running example, an object's room location is observable by
    a robot only when the robot and the object are in the same room:
    \begin{align}
      \label{eqn:fine-observable-specific}
      observable_{loc}(rob_1, O, Pl)~~\mathbf{if}~~loc(rob_1)=Pl
    \end{align}
    Also, the robot can test the presence (or absence) of an object in
    the current cell where the robot is located, and it can always
    test whether it has an object in its grasp. We encode this
    knowledge as:
    \begin{subequations}
      \label{eqn:fine-can-be-observed-specific}
      \begin{align}
        &can\_be\_observed_{loc^*}(rob_1, Th, C) ~~\mathbf{if}~~
        loc^*(rob_1)=C\\
        &can\_be\_observed_{in\_hand}(rob_1, Th, true)
      \end{align}
    \end{subequations}
    We use different fluents ($observable$ and $can\_be\_observed$) to
    serve a similar purpose because the conditions under which a
    particular value of a particular function can be observed may be
    significantly different at the coarse-resolution and
    fine-resolution.

  \item Introduce axioms for indirect observation of functions.
    First, we introduce a defined fluent for each indirectly
    observable function $f$:
    \begin{align*}
      may\_be\_true_{f}: robot\times dom(f) \times range(f)
      \rightarrow boolean
    \end{align*}
    which holds true if the value of $f(x)$, where $x\in dom(f)$ may
    be discovered to be $y\in range(f)$.
    
    The axioms for indirect observation are then given by:
    \begin{subequations}
      \label{eqn:refine-indirect-obs}
      \begin{align}
        &observed_{f}(R, \bar{X}, Y) = true~~{\bf
          if}~~observed_{f^*}(R, \bar{X}^*, C) = true,\\
        \nonumber
        &~~~~~~~~~~~~~~~~~~~~~~~~~~~~~~~~~~~~component(X^*_1,
        X_1),\ldots,~component(X^*_m, X_m),~component(C,Y)\\
        &may\_be\_true_{f}(R, \bar{X}, Y) ~~{\bf if}~~ observed_{f}(R,
        \bar{X}, Y) = true\\
        &may\_be\_true_{f}(R,\bar{X},Y)~~{\bf if}~~ observed_{f}(R,
        \bar{X}, Y) = undet, ~component(C,Y),\\\nonumber
        &~~~~~~~~~~~~~~~~~~~~~~~~~~~~~~~~~~~~~~~~~observed_{f^*}(R,
        \bar{X}^*, C)=undet \\ &observed_{f}(R, \bar{X}, Y) =
        false~~{\bf if}~~\neg may\_be\_true_{f} (R, \bar{X}, Y) \\
        &observed_{f}(R, \bar{X}, Y_1) = false ~~\mathbf{if}~~
        observed_{f}(R, \bar{X}, Y_2), ~Y_1\not=Y_2
      \end{align}
    \end{subequations}
    % where $\bar{X}^* = \{X_1^*, \ldots, X_m^*\}$ is any
    % fine-resolution counterpart of $\bar{X} = \{X_1, \ldots, X_m\}$.
    which implies that a coarse-resolution function is observed to
    have a particular value if any of its fine-resolution counterparts
    is observed to be true, and that the coarse-resolution function
    may be observed to have a particular value as long as it is
    possible that at least one of its fine-resolution counterparts may
    be observed to be true.

    \medskip
    \noindent
    In our example domain, observing an object in a cell in a room
    implies that the object is indirectly observed to be in the room:
    \begin{align*}
      &observed_{loc}(R, O, P) = true~~{\bf if}~~ observed_{loc^*}(R,
      O, C) = true, component(C, P)
    \end{align*}
    Example~\ref{ex:refined2} includes other examples of the use of
    such axioms for indirect observations.
  \end{itemize}
\end{enumerate}

%%%%%-----------------------------------------------------------------------------
\subsubsection{Strong Refinement}
\label{sec:arch-refine-strong}
We are now ready to define a notion of strong refinement that takes
into account the theory of observations. We do so by expanding
Definition~\ref{def:weak-refinement} of weak refinement as follows.

\begin{defn2}\label{def:strong-refinement}[Strong refinement of $\tau_H$]\\
  A transition diagram $\tau_L$ over $\Sigma_L$ is called a
  \emph{strong refinement} of $\tau_H$ if:
  \begin{enumerate}
  \item For every state $\sigma^\diamond$ of $\tau_L$, the collection
    $\sigma^\diamond |_{\Sigma_H}$ of atoms of $\sigma^\diamond$
    formed by symbols from $\Sigma_H$ is a state of $\tau_H$.
  
  \item For every state $\sigma$ of $\tau_H$, there is a state
    $\sigma^\diamond$ of $\tau_L$ such that $\sigma^\diamond$ is an
    extension of $\sigma$.

  \item For every transition $T = \langle \sigma_1,a^H,\sigma_2
    \rangle$ of $\tau_H$, if $\sigma^{\diamond}_1$ is an extension of
    $\sigma_1$, then for every observable fluent $f$ such that
    $observable_f(rob_1, \bar{x}, y)\in \sigma_2$, there is a path $P$
    in $\tau_L$ from $\sigma^{\diamond}_1$ to an extension
    $\sigma^{\diamond}_2$ of $\sigma_2$ such that:
    \begin{itemize}
    \item $P$ is \emph{pertinent} to $T$, i.e., all states of $P$ are
      extensions of $\sigma_1$ or $\sigma_2$;
    \item actions of $P$ are concrete, i.e., directly executable by
      robots; and
    \item $observed_f(rob_1, x, y)= true \in \sigma^{\diamond}_2$ iff
      $(f(x)=y) \in \sigma^{\diamond}_2$, and $observed_f(rob_1, x,
      y)= false \in \sigma^{\diamond}_2$ iff $(f(x)=y_1) \in
      \sigma^{\diamond}_2$ and $y_1 \not= y$.
    \end{itemize}
  \end{enumerate}
\end{defn2}

\medskip
%\noindent
We are now ready to complete the second step of the design
methodology, i.e., constructing the fine-resolution system description
that considers the robot's ability to observe the values of domain
fluents. We do so by expanding $\mathcal{D}_{L,nobs}$ to include the
theory of observations.

\medskip
\noindent
\begin{center}
  \fbox{%
    \parbox{0.95\textwidth}{%
      2. Constructing the fine-resolution system description
      $\mathcal{D}_L$ that is the refinement of the coarse-resolution
      system description $\mathcal{D}_H$.\\
      \noindent 
      (b) Constructing $\mathcal{D}_L$ with theory of observations. }
  }
\end{center}
\medskip
\noindent
Specifically, the system description $\mathcal{D}_L$ is obtained by:
\begin{enumerate}
\item Augmenting signature $\Sigma_{L,nobs}$ of $\mathcal{D}_{L,
    nobs}$ with the actions (e.g., $test$) and fluents (e.g.,
  $can\_be\_observed$) of the theory of observations.

\item Augmenting the axioms of $\mathcal{D}_{L, nobs}$ with the axioms
  needed to represent the robot's ability to observe, i.e.,
  Statements~\ref{eqn:refine-test},~\ref{eqn:refine-test-impossible},
  and~\ref{eqn:refine-indirect-obs}, and axioms for domain-dependent,
  observation-related defined fluents.
\end{enumerate}

\medskip
\noindent 
Next, consider the relationship between $\mathcal{D}_{L, nobs}$, a
well-founded system description, and $\mathcal{D}_L$, its extension by
the theory of observations, which is used in the proof of the
following proposition.  If $\tau_{L, nobs}$ and $\tau_L$ are the
transition diagrams defined by these system descriptions, then:
\begin{itemize}
\item The states of $\tau_{L, nobs}$ and $\tau_L$ differ mainly in the
  knowledge functions $observed_f$ for directly or indirectly
  observable fluents.
\item For every transition $\langle \sigma^{\diamond}_1,
  test_f(rob_1,y), \sigma^{\diamond}_2 \rangle$ of $\tau_L$:
  \begin{itemize}
  \item Physical fluents are the same in $\sigma^{\diamond}_1$ and
    $\sigma^{\diamond}_2$.
  \item $observed_f(rob_1, \bar{x}, y)= true \in \sigma^{\diamond}_2$
    iff $(f(\bar{x}) = y) \in \sigma^{\diamond}_2$.
  \item $observed_f(rob_1, \bar{x}, y)= false \in \sigma^{\diamond}_2$
    iff $(f(\bar{x}) = y_1) \in \sigma^{\diamond}_2$ and $y_1 \not=
    y$.
  \end{itemize}
\end{itemize}

\medskip
\noindent
Finally, the following proposition says that $\mathcal{D}_{L}$ as
constructed above is a strong refinement of $\mathcal{D}_H$.
\begin{prop2}\label{prop:strong-refinement}[Strong Refinement]\\
  {\rm Let $\mathcal{D}_H$ and $\mathcal{D}_{L}$ be the
    coarse-resolution and fine-resolution system descriptions from our
    running example.  Then $\tau_{L}$ is a strong refinement of
    $\tau_H$.  }
\end{prop2}
\medskip
\noindent
The proof of this proposition is in
Appendix~\ref{sec:appendix-prop-strong-refine}. Please see
\texttt{refined.sp} at \url{https://github.com/mhnsrdhrn/refine-arch}
for the ASP program (in SPARC format) describing the refined signature
and refined axioms for our illustrative example, along with additional
axioms that support planning to achieve particular goals.

\begin{example2}\label{ex:refined2}[Another example of refinement]\\
  {\rm Let $\mathcal{D}_H$ and its refinement $\mathcal{D}_L$ be as in
    example Example~\ref{ex:logician-example}. The key difference is
    that, in addition to the cells of rooms, the increase in
    resolution has also led to the discovery of component of cups such
    as $handle$ and $base$. To construct a refinement
    $\mathcal{D}_{L}^e$ of $\mathcal{D}_H$ suitable for this expanded
    domain, we expand the signature of $\mathcal{D}_L$ by a new sort
    $cup^*$ and add it to the sort hierarchy of $\mathcal{D}_H$ as a
    sibling of sort $cup$.  Now the sort $object$ has three children:
    $cup$, $cup^*$ and $textbook$. Similar to $\mathcal{D}_L$, we will
    need the sort $place^*$ and object constants of specific sorts
    such as:
    \begin{align*}
      &textbook = \{tb_1, tb_2\}\\
      &cup = \{cup_1\}\\
      &cup^* = \{cup\_base_1, cup\_handle_1\} \\
    \end{align*}
    Similar to $\mathcal{D}_L$, we will need the function $loc^*$, and
    we need new instances of the component relation:
    \begin{align*}
      &component(cup\_base_1, cup_1)\\
      &component(cup\_handle_1, cup_1) \\
      &\ldots
    \end{align*}
    To construct $\mathcal{D}_{L}^e$, we consider actions that can no
    longer be executed directly, and then consider fluents that can no
    longer be observed directly.

    \medskip
    \noindent
    In our example, actions $grasp$ and $putdown$ are no longer
    directly executable on cups, but are executable on the components
    of cups. To support indirect execution of these actions on cups,
    we introduce new executable actions $grasp^*(robot, cup^*)$ and
    $putdown^*(robot, cup^*)$ for grasping and putting down a cup's
    handle and base. The system description $\mathcal{D}_{L}^e$ will
    inherit from $\mathcal{D}_L$ the axioms for $next\_to^*$, $move^*$
    and $loc^*$, i.e., Statements~\ref{eqn:refine-causal-specific}(a),
    ~\ref{eqn:refine-constraint-specific}(a-b),
    ~\ref{eqn:refine-executability-specific}(a-b),
    and~\ref{eqn:refine-bridge-specific}(a-b).  Ground instances of
    the axiom describing the effects of $grasp$ for objects other than
    cups and their parts will remain as in $\mathcal{D}_H$; using
    variables, this can be written as:
    \begin{align}
      \label{eqn:refine-causal-grasp-old}
      grasp(R, O)~~&\mathbf{causes}~~ in\_hand(R, O) ~~\mathbf{if}~~ O
      \not\in cup, O \not\in cup^*
    \end{align}
    A new axiom is needed to describe the effects of grasping parts of
    cups:
    \begin{align}
      \label{eqn:refine-causal-grasp-new}
      grasp^*(R, O)~~&\mathbf{causes}~~ in\_hand(R, O)~~\mathbf{if}~~
      O \in cup^*
    \end{align}
    Executability conditions for $grasp$ and $grasp^*$ are handled in
    a similar manner. In addition to
    Statement~\ref{eqn:refine-executability-specific}(c) of
    $\mathcal{D}_L$:
    \begin{align*}
      \mathbf{impossible}~~ grasp(R, O) ~~\mathbf{if}~~ loc^*(R)= C_1,
      ~loc^*(O)=C_2, ~C_1\neq C_2
    \end{align*}
    we will need an additional axiom for $grasp^*$:
    \begin{align}
      \label{eqn:refine-exec-grasp-new}
      \mathbf{impossible}~~ grasp^*(R, O) ~~\mathbf{if}~~ loc^*(R)=
      C_1, ~loc^*(O)=C_2, ~C_1\neq C_2
    \end{align}
    Similar axioms are also introduced for actions $putdown$ and
    $putdown^*$. Finally, we will need axioms describing newly
    discovered relationships between objects and their parts:
    \begin{subequations}
      \label{eqn:refine-bridge-specific-new}
      \begin{align}
        &in\_hand(R,O) = in\_hand(R, OPart)~~\mathbf{if}~~ component(OPart, O)\\
        &(loc^*(O) = C) = (loc^*(OPart) = C) ~~\mathbf{if}~~
        component(OPart, O)
      \end{align}
    \end{subequations}
    where the equality is shorthand for two statements\footnote{$f(x)
      = g(x)~~\mathbf{if}~~ body$ is shorthand for $f(x) = y
      ~~\mathbf{if}~~ body,~g(x) = Y$ and $g(x) = y ~~\mathbf{if}~~
      body,~f(x) = Y$.}.  To illustrate reasoning with $D_{L}^e$
    consider initial situation in which $rob_1$ and $cup_1$ are in a
    cell $c_5$ of $kitchen$.  Suppose $rob_1$ grasps the cup's handle,
    i.e., $grasp^*(rob_1, cup\_handle_1)$ is executed, and moves to
    location $c_2$ of $of\!\!fice$, i.e., executes $move^*(rob_1,
    c_2)$.  Both actions are clearly executable based on
    Statement~\ref{eqn:refine-exec-grasp-new} and
    Figure~\ref{fig:refine-gridmap}. By
    Statement~\ref{eqn:refine-causal-grasp-new}, after the execution
    of $grasp^*$, the handle will be in the robot's hand, i.e.,
    $in\_hand(rob_1, cup\_handle_1)$. Based on
    Statement~\ref{eqn:refine-causal-specific}(a), executing action
    $move^*$ will result in $loc^*(rob_1) = c_2$. Based on
    Statement~\ref{eqn:refine-constraint-specific}(a), we conclude
    that $loc^*(cup\_handle_1) = c_2$. Then, based on
    Statement~\ref{eqn:refine-bridge-specific-new}(b), we have
    $loc^*(cup_1) = c_2$ and thus, by
    Statement~\ref{eqn:refine-bridge-specific}(a), $loc(cup_1) =
    of\!\!fice$.

    \medskip 
    \noindent
    Next, we examine the effect of the robot being able to directly
    observe neither the location of a cup nor if it is holding a cup,
    but being able to do so for a cup's parts and for textbooks.  This
    requires us to introduce fine-resolution counterparts $loc^*:
    cup^*\rightarrow place^*$ and $in\_hand^*(robot, cup^*)$ of $loc$
    and $in\_hand$ respectively for cups, and change the related
    axioms.  Statement~\ref{eqn:refine-causal-grasp-old} is not
    related to cups and remains unchanged.
    Statement~\ref{eqn:refine-causal-grasp-new} will, on the other
    hand, be replaced by the axiom:
    \begin{align}
      grasp^*(R, O)~~&\mathbf{causes}~~ in\_hand^*(R, O)
      ~~\mathbf{if}~~ O \in cup^*
    \end{align}
    The bridge axioms in
    Statement~\ref{eqn:refine-bridge-specific-new}(a-b) will be
    replaced by bridge axioms:
    \begin{subequations}
      \begin{align}
        &in\_hand(R, Cup) ~~\mathbf{if}~~ in\_hand^*(R, Part),
        ~component(Part, Cup)\\
        &(loc^*(O) = C) ~~\mathbf{if}~~ (loc^*(OPart) = C),
        ~component(OPart, O)
      \end{align} 
    \end{subequations}
    defining $in\_hand$ and $loc$ for cups in terms of its
    fine-resolution counterparts. Next, we introduce actions
    $test_{loc^*}$ and $test_{in\_hand^*}$ to check the cell location
    of a cup's part and to check whether a part of a cup is in the
    robot's hand:
    \begin{align*}
      test_{loc^*}(R, OPart, C) ~~\mathbf{causes}~~
      observed_{loc^*}(R, OPart, C) = true ~~\mathbf{if}&~~ loc^*(OPart) = C\\
      test_{loc^*}(R, OPart, C) ~~\mathbf{causes}~~
      observed_{loc^*}(R, OPart, C) = false ~~\mathbf{if}&~~
      loc^*(OPart) = C_1, ~C_1\not= C \\
      test_{in\_hand^*}(R, OPart, V) ~~\mathbf{causes}~~
      observed_{in\_hand^*}(R, OPart, V) = true ~~\mathbf{if}&~~
      in\_hand^*(R, OPart) = V \\
      test_{in\_hand^*}(R, OPart, V) ~~\mathbf{causes}~~
      observed_{in\_hand^*}(R, OPart, V) = false ~~\mathbf{if}&~~
      in\_hand^*(R, OPart) = V_1,\\ &~V_1\not= V
    \end{align*}
    We also replace
    Statements~\ref{eqn:fine-can-be-observed-specific}(a-b) about the
    observation-related, domain-dependent defined fluents with:
    \begin{subequations}
      \begin{align}
        &can\_be\_observed_{loc^*}(R, Th, C) ~~\mathbf{if}~~
        loc^*(R)=C, ~ Th\not\in cup \\
        &can\_be\_observed_{in\_hand}(R, O, V) ~~\mathbf{if}~~ O
        \not\in cup, \not\in cup^*\\
        &can\_be\_observed_{in\_hand^*}(R, O, V) ~~\mathbf{if}~~ O
        \in cup^*
      \end{align}
    \end{subequations}
    which imply that the robot can no longer directly observe the
    location of a cup or whether a cup is in its hand; it can do so
    for parts of cups.  Reasoning similar to that used in the context
    of $grasp^*$ above can be used to show that if the robot grasps a
    cup's handle and moves to a cell $c_2$ of the $of\!\!fice$, the
    robot, the cup, and the cup's handle will be in the $of\!\!fice$.
    If needed, test actions and the theory of observations can be used
    to observe that the robot is holding the cup, and to observe the
    locations of other things in the domain.

    % SHOULD $loc^*$ be still directly observable for cups?  INCLUDE
    % FILLING UP THE CAP and SAME.
    \medskip
    \noindent
    Next, consider the inclusion of an additional action $fill(robot,
    cup)$ in $\mathcal{D}_H$. Executing this action causes a cup to be
    filled, i.e., we introduce a basic fluent $filled: cup \rightarrow
    boolean$ in $\mathcal{D}_H$, and the corresponding axioms:
    \begin{subequations}
      \begin{align}
        &fill(R, C) ~~\mathbf{causes}~~ filled(C)\\
        &\mathbf{impossible}~~ fill(R, C) ~~\mathbf{if}~~ filled(C)\\
        &\mathbf{impossible}~~ fill(R, C) ~~\mathbf{if}~~ loc(C) =
        P_1, ~loc(R) = P_2, ~P_1 \neq P_2
      \end{align}
    \end{subequations}
    Here, action $fill$ is directly executable and fluent $filled$ in
    directly observable. We also include $same\_loc: object\times
    object \rightarrow boolean$ in $\mathcal{D}_H$, a defined fluent
    to reason about co-occurrence of objects, with the corresponding
    axiom:
    \begin{align}
      \label{eqn:refine-before-same-loc}
      same\_loc(O_1, O_2)~~\mathbf{if}~~loc(O_1) = Pl,~loc(O_2) = Pl
    \end{align}
    which defines when two objects are considered to be in the same
    place. In the refined system description, the action $fill$ is
    still directly executable, and the fluent $filled$ is directly
    observable, for cups. These functions are not defined for parts of
    cups, e.g., we cannot fill a cup's handle, and we thus do not
    create their fine-resolution counterparts (e.g., $fill^*$). We do,
    however, introduce a new function in the signature of the refined
    system description:
    \begin{align}
      &same\_loc^*(O_1, O_2), ~O_1, O_2 \not\in cup
    \end{align}
    representing parts of cups and/or other objects being in the same
    grid cell location. Note that elements of sort $cup$ are not
    included because we cannot directly observe the location of a cup
    in the fine resolution. We also introduce the following axiom in
    the refined system description, which corresponds to
    Statement~\ref{eqn:refine-before-same-loc} in $\mathcal{D}_H$:
    \begin{align}
      &same\_loc^*(O_1, O_2) ~~\mathbf{if}~~ loc^*(O_1) =
      C,~loc^*(O_2) = C
    \end{align}
    Finally, we need to introduce a suitable bridge axiom:
    \begin{align}
      &same\_loc(O_1, O_2) ~~\mathbf{if}~~ loc^*(OPart_1) = C_1,
      ~loc^*(OPart_2) = C_2,\\\nonumber
      &~~~~~~~~~~~~~~~~~~~~~~~~~component(C_1, P), ~component(C_2,
      P),\\ \nonumber &~~~~~~~~~~~~~~~~~~~~~~~~component(OPart_1,
      O_1), ~component(OPart_2, O_2)
    \end{align}
    Once we have the refined system description, we can reason with it
    as before. For instance, consider a fine-resolution state in which
    $loc^*(cup\_handle_1) = c_5$ and $loc^*(tb_1) = c_6$ where $c_5$
    and $c_6$ are grid cells in the $kitchen$. Based on the bridge
    axioms, we can infer that $loc(cup_1) = kitchen$, $loc(tb_1) =
    kitchen$ and $same\_loc(cup_1, tb_1)$.  }
\end{example2}

\subsection{Randomization}
\label{sec:arch-random}
%\medskip\noindent 
The system description $\mathcal{D}_L$ of transition diagram $\tau_L$,
obtained by refining transition diagram $\tau_H$, is insufficient to
implement a coarse-resolution transition $T=\langle \sigma_1, a^H,
\sigma_2\rangle \in \tau_H$. We still need to capture the
non-determinism in action execution, which brings us to the third step
of the design methodology (Section~\ref{sec:methodology}).

\medskip
\noindent
\begin{center}
  \fbox{%
    \parbox{0.95\textwidth}{%
      3. Provide domain-specific information and randomize the
      fine-resolution description of the domain to capture the
      non-determinism in action execution.  } }
\end{center}
\medskip
\noindent
This step models the non-determinism by first creating
$\mathcal{D}_{LR}$, the randomized fine-resolution system description,
by:
\begin{itemize}
\item Replacing each action's deterministic causal laws in
  $\mathcal{D}_L$ by non-deterministic ones; and
\item Modifying the signature by declaring each affected fluent as a
  \emph{random fluent}, i.e., define the set of values the fluent can
  choose from when the action is executed. A defined fluent may be
  introduced to describe this set of values in terms of other
  variables.
\end{itemize}
For instance, consider a robot moving to a specific cell in the
$of\!\!fice$.  During this move, the robot can reach the desired cell or
one of the neighboring cells. The causal law for the $move$ action in
$\mathcal{D}_L$ can therefore be (re)stated as:
\begin{align}
  \label{eqn:move-randomize}
  move^*(R, C_2)~~\mathbf{causes}~~loc^*(R)~=~\{C : range(loc^*(R),
  C)\}
\end{align}
where the relation $range$ is a defined fluent used to represent the
cell the robot currently is in, and the cells next to its current
location:
\begin{align*}
  &range(loc^*(R), C)~~\mathbf{if}~~loc^*(R) = C \\
  &range(loc^*(R), C)~~\mathbf{if}~~loc^*(R)=C_1,~next\_to^*(C, C_1)
\end{align*}
As described by Statement~\ref{eqn:refine-executability-specific}(b),
the robot can still only move to a cell that is $next\_to$ its current
location.
% \begin{align}
%   \label{eqn:move-randomize-execute}
%   \mathbf{impossible}~~move^*(R, C_2)~~\mathbf{if}~~loc^*(R)=C_1,~\neg
%   next\_to^*(C_1,C_2)
% \end{align}
In general, the fluent affected by the change in the causal law can
take one of a set of values that satisfy a given property ($range$ in
the current example), as described in
Statement~\ref{eqn:causal-law-1}. In a similar manner, the
non-deterministic version of the $test$ action used to determine the
robot's cell location in the $of\!\!fice$, is given by:
\begin{align*}
  test_{loc^*}(rob_1, rob_1, c_i)~~{\bf causes }~~
  observed_{loc^*}(rob_1, rob_1, c_i)= \{true, false\} ~~{\bf if}~~
  loc^*(rob_1) = c_i
\end{align*}
which indicates that the result of the $test$ action may not always be
as expected, and $c_i$ are cells in the $of\!\!fice$. Similar to
refinement, \emph{it is the designer's responsibility to provide
  domain-specific information needed for randomization}. Furthermore,
note that the paths in the randomized transition diagram $\tau_{LR}$
match those in $\tau_L$ except for the addition of the defined fluents
that model the domain-specific information.

% \commentm{Need proper transition from DLR to DLR-P and the need for
%   collecting statistics...}

\medskip
\noindent
Once the randomized system description $\mathcal{D}_{LR}$ has been
constructed, we can construct its probabilistic version
$\mathcal{D}_{LR}^P$ that consists of:
\begin{itemize}
\item System description $\mathcal{D}_{LR}$ that defines a
  non-deterministic transition diagram $\tau_{LR}$ of all the system's
  trajectories.
\item A function $P$ assigning probabilities to each of the
  transitions.
\end{itemize}
In addition to the actual states, $\mathcal{D}_{LR}^P$ will reason
with probability distributions over the states, and consider
transitions between one such probabilistic state to another as a
result of executing particular actions. We obtain the probabilities
needed to construct $\mathcal{D}_{LR}^P$, we experimentally collect
statistics of action outcomes and the reliability of observations as
described below.

\medskip
\noindent
\paragraph{\underline{Collecting statistics:}} 
Running experiments to collect statistics that are used to compute the
probabilities of action outcomes and the reliability of observations,
corresponds to the fourth step of the design methodology (see
Section~\ref{sec:methodology}).

\medskip
\noindent
\begin{center}
  \fbox{%
    \parbox{0.95\textwidth}{%
      4. Run experiments, collect statistics, and compute
      probabilities of action outcomes and reliability of
      observations.} }
\end{center}
\medskip

\noindent
Specifically, we need to compute the:
\begin{itemize}
\item Causal probabilities for the outcomes of physical actions; and
\item Probabilities for the outcomes of the knowledge-producing
  actions, i.e., a quantitative model for the observations being
  correct.
\end{itemize}
This collection of statistics is \emph{typically a one-time process
  performed in an initial training phase}, although it is also
possible to do this incrementally over time. Also, the statistics are
computed separately for each basic fluent in $\mathcal{D}_{LR}$. To
collect the statistics, we consider one non-deterministic causal law
in $\mathcal{D}_{LR}$ at a time. We sample some ground instances of
this causal law, e.g., corresponding to different atoms in the causal
law. The robot then executes the action corresponding to this sampled
instance multiple times, and collects statistics (e.g., counts) of the
number of times each possible outcome (i.e., value) is obtained. The
robot also collects information about the amount of time taken to
execute each such action.

As an example, consider a ground instance of the non-deterministic
causal law for $move^*$, considering grid cell locations in a
particular room:
\begin{align*}
  move^*(rob_1, c_2)~~\mathbf{causes}~~loc^*(R)~=~\{c_1, c_2, c_3\}
\end{align*}
where $rob_1$ in cell $c_1$ can end up in one of three possible cells
when it tries to move to $c_2$. In ten attempts to move to $c_2$,
assume that $rob_1$ remains in $c_1$ in one trial, reaches $c_2$ in
eight trials, and reaches $c_3$ in one trial. The maximum likelihood
estimates of the probabilities of these outcomes are then $0.1$, $0.8$
and $0.1$ respectively---the probability of $rob_1$ moving to other
cells is zero. Similar statistics are collected for other ground
instances of this causal law, and averaged to compute the statistics
for the fluent $loc$ for $rob_1$. The same approach is used to collect
statistics for other causal laws and fluents, including those related
to knowledge actions and basic knowledge fluents. For instance, assume
that the collected statistics indicate that testing for the presence
of a textbook in a cell requires twice as much computational time (and
thus effort) as testing for the presence of a cup.  This information,
and the relative accuracy of recognizing textbooks and cups, will be
used to determine the relative value of executing the corresponding
test actions. The collected statistics are thus used to define the
probabilities of two different types of transitions in
$\mathcal{D}_{LR}$.

\begin{defn2}\label{def:learned-probs}[Learned transition probabilities]\\
  {\rm The learned state transition probabilities $P(\delta_x, a,
    \delta_y)$ are of two types depending on the type of transition
    between states $\delta_x$ and $\delta_y$ of $\mathcal{D}_{LR}$:
    \begin{enumerate}
    \item \emph{Physical state transition probability}, where
      $\delta_x$ and $\delta_y$ differ in a literal formed of a
      non-knowledge fluent term, e.g., when we consider the
      probability of the robot's location changing from $loc^*(rob_1)
      = c_1$ to $loc^*(rob_1) = c_2$ after executing $move^*(rob_1,
      c_2)$.
    \item \emph{Knowledge state transition probabilities}, where
      $\delta_x$ and $\delta_y$ differ in a literal formed of a
      knowledge fluent term, e.g., when we consider the probability
      that the value of $observed_{loc^*}(rob_1, cup_1, c_2)$ changes
      from $undet$ in $\delta_x$ to $true$ in $\delta_y$.
    \end{enumerate}
  }
\end{defn2}

\noindent
There are some important caveats about collecting statistics and
computing probabilities.
% First, we are able to consider one causal law at a time because each
% such law (by design) has a causal relationship with only one basic
% (knowledge or non-knowledge) fluent, as highlighted in the context
% of Example~\ref{ex:logician-example}. If this restriction is to be
% relaxed, the designer has to collect statistics in the joint space
% of multiple fluents and describe how to use these statistics, e.g.,
% for POMDP construction in Section~\ref{sec:arch-pomdp-construct}.
\begin{itemize}
\item First, although an action can affect different fluents through
  different causal laws, and a change in the value one fluent can
  constrain the value of other fluents, we are collecting statistics
  of the change in the value of one basic fluent by considering one
  causal law at a time. For instance, while collecting the statistics
  of a robot's move from cell $c_1$ to cell $c_2$, we do not consider
  whether the robot is holding a book in hand. In our domain, the
  transitions (by design) only differ in a literal formed of one
  non-knowledge or knowledge fluent term. However, states do comprise
  multiple fluents. As discussed later, we will ensure that impossible
  scenarios are not considered and compute the probabilities of valid
  states by suitably combining the individual probabilities. For
  instance, if a robot is holding a book, having the robot and the
  book in different locations does not constitute a valid state.

\item Second, the collection of statistics depends on the availability
  of relevant ground truth information, e.g., we need the actual
  location of robot $rob_1$ after executing $move^*(rob_1, c_2)$. This
  ground truth information is often provided by an external
  high-fidelity sensor during the initial training phase, or by a
  human observer.

\item Third, although we do not do so in our experiments, it is
  possible to use heuristics to model the computational effort, and to
  update the statistics incrementally over time, e.g., the execution
  time of a knowledge-producing action can be computed as a function
  of the size of the input image. If any heuristic functions are to be
  used, the designer has to make them available to automate subsequent
  steps of our control loop.

\item Fourth, considering all ground instances of one causal law at a
  time can require a lot of training in complex domains, but this is
  often unnecessary. For instance, it is often the case that the
  statistics of moving from a cell to one of its neighbors is the same
  for cells in a room and any given robot. In a similar manner, if the
  robot and an object are (are not) in the same cell, the probability
  of the robot observing (not observing) the object is often the same
  for any cell.
  % Also, if the robot tries to move to a cell that it cannot reach
  % (e.g., because of an obstacle), we assume that the robot will
  % remain in the same cell. While these assumptions simplify the
  % collection of statistics, they impose limitations on how the
  % domain and the robot's capabilities are modeled, and on the
  % resulting behavior of the robot.  If these assumptions do not
  % hold, e.g., different rooms have different surfaces, or if the
  % designer desires to construct different models for different
  % action outcomes, it will be the designer's responsibility to
  % collect statistics appropriately.
  The designer thus only considers representative samples of the
  distinct cases to collect statistics, e.g., statistics corresponding
  to moving between cells will be collected in two different rooms
  only if these statistics are expected to be different.
\end{itemize}
There is an extensive literature on estimation of such statistical
models for robots. Types of models learned from data in cognitive
robotics are sensor models and robot motion
models~\cite{thrun:robotics05}, motion models for manipulated
objects~\cite{kopicki2017}, and success-failure models for actions
(e.g., grasping)~\cite{lu:isrr17}. In particular, robot motion models
for standard mobile platforms are available for use in the robotics
community without need for re-learning. In addition, rigid body
physics engines can be used as a source of data for
learning~\cite{haidu2015}. In other work, we have explored incremental
learning of statistics~\cite{zhang:TRO15} and domain
knowledge~\cite{mohan:icaps17}, but these topics are beyond the scope
of this paper.

\medskip
\noindent
Even after the desired probabilities of transitions in
$\mathcal{D}_{LR}$ are computed, reasoning with $\mathcal{D}_{LR}^P$
(as described earlier) will be computationally infeasible for complex
domains. Our architecture addresses this problem by automatically
zooming to the part of $\mathcal{D}_{LR}$ relevant to each coarse
resolution transition $T$ under consideration (as described below),
and then reasoning probabilistically over this zoomed system
description $\mathcal{D}_{LR}(T)$ using POMDPs (see
Section~\ref{sec:arch-pomdp-construct}).

%%%%%-----------------------------------------------------------------------------
\subsection{Zoom} 
\label{sec:arch-zoom}
Reasoning probabilistically about the entire randomized
fine-resolution system description can become computationally
intractable. For any given transition $T=\langle \sigma_1, a^H,
\sigma_2\rangle \in \tau_H$, this intractability could be offset by
limiting fine-resolution probabilistic reasoning to the part of
transition diagram $\tau_{LR}$ whose states are pertinent to $T$. For
instance, for the state transition corresponding to a robot moving
from the $of\!\!fice$ to the $kitchen$ in
Example~\ref{ex:logician-example}, i.e., $a^H = move(rob_1, kitchen)$,
we could only consider states of $\tau_{LR}$ in which the robot's
location is a cell in the $of\!\!fice$ or the $kitchen$.  However,
these states would still contain fluents and actions not relevant to
the execution of $a^H$, e.g., locations of domain objects, and the
$grasp$ action. What we need is a fine-resolution transition diagram
$\tau_{LR}(T)$ whose states contain no information unrelated to the
execution of $a^H$, while its actions are limited to those which may
be useful for such an execution. In the case of $a^H = move(rob_1,
kitchen)$, for instance, states of $\tau_{LR}(T)$ should not contain
any information about domain objects. In the proposed architecture,
the controller constructs such a zoomed fine-resolution system
description $\mathcal{D}_{LR}(T)$ in two steps.  First, a new action
description is constructed by focusing on the transition $T$, creating
a system description $\mathcal{D}_H(T)$ that consists of ground
instances of $\mathcal{D}_H$ built from object constants of $\Sigma_H$
relevant to $T$. In the second step, the refinement of
$\mathcal{D}_H(T)$ is extracted from $\mathcal{D}_{LR}$ to obtain
$\mathcal{D}_{LR}(T)$. We first consider the requirements of the zoom
operation.
%\vspace{1em}
%\hrule
\begin{defn2}\label{def:zoom-require}[Requirements of zoom]\\
  {\rm The following are the requirements the zoom operation should
    satisfy:
    \begin{enumerate}
    \item Every path in the transition diagram obtained after zooming
      should correspond to a path in the transition diagram before
      zooming.  In other words, for every path $P^z$ of $\tau_{LR}(T)$
      between states $\delta^z_1 \subseteq \delta_1$ and $\delta^z_2
      \subseteq \delta_2$, where $\delta_1$ and $\delta_2$ are
      refinements of $\sigma_1$ and $\sigma_2$ respectively, there is
      a path $P$ between states $\delta_1$ and $\delta_2$ in
      $\tau_{LR}$.
%       For instance, $P^z = \langle
%       \delta^z_1,a_1,\delta^z_{1,1},\ldots,a_{n},\delta^z_n\rangle$,
%       there is a path $P = \langle
%       \delta_0,a_0,\delta_1,\dots,a_{n-1},\delta_n\rangle$ in
%       $\tau_{LR}$ such that $\forall i, \,\,\delta_i = \delta^z_i \cup
%       (\delta_0 \setminus \delta^z_0)$.

    \item Every path in the transition diagram before zooming should
      correspond to a path in the zoomed transition diagram. In other
      words, for every path $P$ of $\tau_{LR}$, formed by actions of
      $\tau_{LR}(T)$, between states $\delta_1$ and $\delta_2$ that
      are refinements of $\sigma_1$ and $\sigma_2$ respectively, there
      is a path $P^z$ of $\tau_{LR}(T)$ between states $\delta^z_1
      \subseteq \delta_1$ and $\delta^z_2 \subseteq \delta_2$.

    \item Paths in $\tau_{LR}(T)$ should be of sufficiently high
      probability for the probabilistic solver to find them.
      % ---Section~\ref{sec:arch-pomdp} has more details about the
      % probabilistic solver.
    \end{enumerate}
  }
\end{defn2}
%\vspace{1pt}
%\hrule
\noindent
To construct such a zoomed system description $\mathcal{D}_{LR}(T)$
defining transition diagram $\tau_{LR}(T)$, we begin by defining
$relObCon_H(T)$, the collection of object constants of signature
$\Sigma_H$ of $\mathcal{D}_H$ \emph{relevant} to transition $T$.
%\vspace{1em}
%\hrule
\begin{defn2}\label{def:relevant}[Constants relevant to a transition]\\
  {\rm For any given (ground) transition $T = \langle \sigma_1, a^H,
    \sigma_2\rangle$ of $\tau_H$, by $relObCon_H(T)$ we denote the
    minimal set of object constants of signature $\Sigma_H$ of
    $\mathcal{D}_H$ closed under the following rules:
    \begin{enumerate}
    \item Object constants occurring in $a^H$ are in $relObCon_H(T)$;
    \item If $f(x_1,\ldots, x_n) = y$ belongs to $\sigma_1$ or
      $\sigma_2$, but not both, then $x_1, \ldots, x_n, y$ are in
      $relObCon_H(T)$;
    \item If body $B$ of an executability condition of $a^H$ contains
      an occurrence of a term $f(x_1,\dots,x_n)$ and $f(x_1,\dots,x_n)
      = y \in \sigma_1$ then $~x_1,\dots,x_n,y~$ are in
      $relObCon_H(T)$.
    \end{enumerate}
    Constants from $relObCon_H(T)$ are said to be \emph{relevant} to
    $T$. }
\end{defn2}
%\hrule
%\vspace{1em}
\noindent
In Example~\ref{ex:logician-example}, consider transition $T=\langle
\sigma_1, grasp(rob_1, cup_1), \sigma_2 \rangle$ such that
$loc(rob_1)= kitchen$ and $loc(cup_1)= kitchen$ are in $\sigma_1$.
Then, $relObCon_H(T)$ consists of $rob_1$ of sort $robot$ and $cup_1$
of sort $object$ (based on the first rule above), and $kitchen$ of
sort $place$ (based on the third rule above and fourth axiom in
Statement~\ref{eqn:logician-executability} in
Example~\ref{ex:logician-example}).  For more details, see
Example~\ref{ex:zoom2}.

\medskip
\noindent
Now we are ready for the first step of the construction of
$\mathcal{D}_{LR}(T)$. Object constants of the signature $\Sigma_H(T)$
of the new system description $\mathcal{D}_H(T)$ are those of
$relObCon_H(T)$.  Basic sorts of $\Sigma_H(T)$ are non-empty
intersections of basic sorts of $\Sigma_H$ with $relObCon_H(T)$.  The
domain attributes and actions of $\Sigma_H(T)$ are those of $\Sigma_H$
restricted to the basic sorts of $\Sigma_H(T)$, and the axioms of
$\mathcal{D}_H(T)$ are restrictions of axioms of $\mathcal{D}_H$ to
$\Sigma_H(T)$. It is easy to show that the system descriptions
$\mathcal{D}_H$ and $\mathcal{D}_H(T)$ satisfy the following
requirement---for any transition $T = \langle \sigma_1, a^H,
\sigma_2\rangle$ of transition diagram $\tau_H$ corresponding to
system description $\mathcal{D}_H$, there exists a transition $\langle
\sigma_1(T), a^H, \sigma_2(T)\rangle$ in transition diagram
$\tau_H(T)$ corresponding to system description $\mathcal{D}_H(T)$,
where $\sigma_1(T)$ and $\sigma_2(T)$ are obtained by restricting
$\sigma_1$ and $\sigma_2$ (respectively) to the signature
$\Sigma_H(T)$. % Notice, for instance,
% that for the states of $\mathcal{D}_H$ corresponding to transition
% $T$, removing atoms irrelevant to $T$ results in states of
% $\mathcal{D}_H(T)$, with action $a^H\in T$ causing the transition
% between them.

In the second step, the zoomed system description
$\mathcal{D}_{LR}(T)$ is constructed by refining the system
description $\mathcal{D}_H(T)$. Unlike the description of refinement
in Section~\ref{sec:arch-refine}, which requires the designer to
supply domain-specific information, we do not need any additional
input from the designer for refining $\mathcal{D}_H(T)$ and can
automate the entire zoom operation. We now provide a formal definition
of the zoomed system description.
%\vspace{1em}
%\hrule
\begin{defn2}\label{def:zoom}[Zoomed system description]\\
  {\rm For a coarse-resolution transition $T$, system description
    $\mathcal{D}_{LR}(T)$ with signature $\Sigma_{LR}(T)$ is said to
    be the \emph{zoomed fine-resolution system description} if:
    \begin{enumerate}
    \item Basic sorts of $\Sigma_{LR}(T)$ are those of
      $\mathcal{D}_{LR}$ that are components of the basic sorts of
      $\mathcal{D}_H(T)$.
      % They form a hierarchy based on the subclass relation inherited
      % from $\mathcal{D}_{LR}$.
    
    \item Functions of $\Sigma_{LR}(T)$ are those of
      $\mathcal{D}_{LR}$ restricted to the basic sorts of
      $\Sigma_{LR}(T)$.

    \item Actions of $\Sigma_{LR}(T)$ are those of $\mathcal{D}_{LR}$
      restricted to the basic sorts of $\Sigma_{LR}(T)$.

    \item Axioms of $\mathcal{D}_{LR}(T)$ are those of
      $\mathcal{D}_{LR}$ restricted to the signature $\Sigma_{LR}(T)$.
    \end{enumerate}
  }
\end{defn2}
%\hrule
%\vspace{1em}
\noindent
Consider $T=\langle \sigma_1, move(rob_1, kitchen), \sigma_2 \rangle$
such that $loc(rob_1)= of\!\!fice \in \sigma_1$.  The basic sorts of
$\Sigma_{LR}(T)$ include $robot_L^z=\{rob_1\}$, $place_L^{z} =
\{of\!\!fice, kitchen\}$ and $place_L^{*z} = \{c_i : c_i\in
kitchen\cup of\!\!fice\}$. Functions of $\Sigma_{LR}(T)$ include
$loc^*(rob_1)$ taking values from $place_L^{*z}$, $loc(rob_1)$ taking
values from $place_L^z$, $range(loc^*(rob_1), place_L^{*z})$, statics
$next\_to^*(place_L^{*z}, place_L^{*z})$ and $next\_to(place_L^z,
place_L^z)$, properly restricted functions related to testing the
values of fluent terms etc. The actions include $move^*(rob_1, c_i)$
and $test_{loc^*}(rob_1, rob_1, c_i)$, where $c_i$ are individual
elements of $place_L^{*z}$.  Finally, restricting the axioms of
$\mathcal{D}_{LR}$ to the signature $\Sigma_{LR}(T)$ removes causal
laws for $grasp$ and $put\_down$, and the state constraint encoded by
Statement~\ref{eqn:logician-constraint}(a) in $\mathcal{D}_{LR}$.
Furthermore, in the causal law and executability condition for
$move^*$, we only consider cells in the $kitchen$ or the $of\!\!fice$.

\medskip
\noindent
Based on Definition~\ref{def:strong-refinement} and
Proposition~\ref{prop:strong-refinement}, it is easy to show that the
system descriptions $\mathcal{D}_H(T)$ and $\mathcal{D}_{LR}(T)$
satisfy the following requirement---for any transition $\langle
\sigma_1(T), a^H, \sigma_2(T)\rangle$ in transition diagram
$\tau_H(T)$ of system description $\mathcal{D}_H(T)$, where
$\sigma_1(T)$ and $\sigma_2(T)$ are obtained by restricting states
$\sigma_1$ and $\sigma_2$ (respectively) of $\mathcal{D}_H$ to
signature $\Sigma_H(T)$, there exists a path in $\tau_{LR}(T)$ between
every refinement $\delta_1^z$ of $\sigma_1(T)$ and a refinement
$\delta_2^z$ of $\sigma_2(T)$. We now provide two examples of
constructing the zoomed system description. In Example~\ref{ex:zoom1},
the coarse-resolution action corresponds to a robot grasping a cup. In
Example~\ref{ex:zoom2}, we consider the coarse-resolution action of
the robot moving from one room to another, and demonstrate the
benefits of zooming when additional functions are included in the
system description.

%\smallskip
\noindent
\begin{example2}\label{ex:zoom1}[First example of  zoom]\\
  {\rm As an illustrative example of zoom, consider the transition
    $T=\langle \sigma_1, grasp(rob_1, cup_1), \sigma_2 \rangle$ such
    that $(loc(rob_1)= kitchen) \in \sigma_1$. Based on
    Definition~\ref{def:relevant}, $relObCon_H(T)$ consists of $rob_1$
    of sort $robot$ and $cup_1$ of sort $cup$, and $kitchen$ of sort
    $place$---basic sorts of $\Sigma_H(T)$ are intersections of these
    sorts with those of $\Sigma_H$. The domain attributes and actions
    of $\Sigma_H(T)$ are restricted to these basic sorts, and axioms
    of $\mathcal{D}_H(T)$ are those of $\mathcal{D}_H$ restricted to
    $\Sigma_H(T)$. Now, the signature $\Sigma_{LR}(T)$ of the zoomed
    system description $\mathcal{D}_{LR}(T)$ has the following:
    \begin{itemize}
    \item Basic sorts $robot_L^z=\{rob_1\}$, $place_L^{z} =
      \{kitchen\}$, $place_L^{*z} = \{c_i : c_i\in kitchen\}$, and
      $object_L^z=\{cup_1\}$.

    \item Functions that consist of (a) basic non-knowledge fluents
      $loc^*(robot_L^z)$ and $loc^*(object_L^z)$ that take values from
      $place_L^{*z}$, $loc(robot_L^z)$ and $loc(object_L^z)$ that take
      values from $place_L^{z}$, and $in\_hand(robot_L^z,
      object_L^z)$; (b) $range(loc^*(robot_L^z), place_L^{*z})$; (c)
      statics $next\_to^*(place_L^{*z}, place_L^{*z})$ and
      $next\_to(place_L^z, place_L^z)$; (d) knowledge fluents
      restricted to the basic sorts and fluents.

    \item Actions such as (a) $move^*(robot_L^z, place_L^{*z})$; (b)
      $grasp(robot_L^z, object_L^z)$; (c) $putdown(robot_L^z,
      object_L^z)$; (d) knowledge-producing actions
      $test_{loc^*}(robot_L^z, robot_L^z, place_L^{*z})$ and
      $test_{loc^*}(robot_L^z, object_L^z, place_L^{*z})$.
    \end{itemize}

    \medskip
    \noindent
    The axioms of $\mathcal{D}_{LR}(T)$ are those of
    $\mathcal{D}_{LR}$ restricted to the signature $\Sigma_{LR}(T)$.
    These axioms include:
    \begin{align*}
      &move^*(rob_1, c_j)~~\mathbf{causes}~~loc^*(rob_1)~=~\{C :
      range(loc^*(rob_1), C)\}\\
      &grasp(rob_1, cup_1)~~\mathbf{causes}~~in\_hand(rob_1,
      cup_1)=\{true, false\} \\
      &test_{loc^*}(rob_1, rob_1, c_j)~~{\bf causes }
      ~~observed_{loc^*}(rob_1, rob_1, c_j)= \{true, false\}
      ~~{\bf if}~~ loc^*(rob_1) = c_j\\
      &test_{loc^*}(rob_1, cup_1, c_j)~~{\bf causes }~~
      observed_{loc^*}(rob_1, cup_1, c_j)= \{true, false\}
      ~~{\bf if}~~ loc^*(cup_1) = c_j\\
      &\mathbf{impossible}~~move^*(rob_1, c_j) ~~\mathbf{if} ~~
      loc^*(rob_1)=c_i,~\neg next\_to^*(c_j, c_i)\\
      &\mathbf{impossible}~~grasp(rob_1, cup_1) ~~\mathbf{if} ~~
      loc^*(rob_1)=c_i,~loc^*(cup_1)=c_j,~c_i\neq c_j
    \end{align*}
    where $range(loc^*(rob_1), C)$ may hold for $C\in \{c_i, c_j,
    c_k\}$, cells within the range of the robot's current location
    ($c_i$) and elements of sort $place_L^{*z}$. The states of
    $\tau_{LR}(T)$ include atoms such as $loc^*(rob_1)=c_i$ and
    $loc^*(cup_1)=c_j$, where $c_i, c_j \in place_L^{*z}$,
    $in\_hand(rob_1, cup_1)$, $observed_{loc^*}(rob_1, rob_1, c_i) =
    true$, and $next\_to^*(c_i,c_j)$.  Actions include $move^*(rob_1,
    c_i)$, $grasp(rob_1, cup_1)$, $putdown(rob_1, cup_1)$,
    $test_{loc^*}(rob_1, rob_1, c_i)$ and $test_{loc^*}(rob_1, cup_1,
    c_i)$.  }
\end{example2}

\begin{example2}\label{ex:zoom2}[Second example of  zoom]\\
  {\rm As another example of zoom, consider the transition $T=\langle
    \sigma_1, move(rob_1, kitchen), \sigma_2 \rangle$ such that
    $loc(rob_1)= of\!\!fice \in \sigma_1$. In addition to the
    description in Example~\ref{ex:logician-example}, the domain
    includes (a) boolean fluent $broken(robot)$; and (b) fluent
    $color(robot)$ taking a value from a set of colors---there is also
    an executability condition:
    \begin{align*}
      \mathbf{impossible}~move(Rb, Pl)~~\mathbf{if}~~broken(Rb)
    \end{align*}
    Intuitively, $color(Rb)$ and $broken(Rb)$, where $Rb \not= rob_1$,
    are not relevant to $a^H$, but $broken(rob_1)$ is relevant.
    Specifically, based on Definition~\ref{def:relevant},
    $relObCon_H(T)$ consists of $rob_1$ of sort $robot$, and
    $\{kitchen, of\!\!fice\}$ of sort $place$---basic sorts of
    $\Sigma_H(T)$ are intersections of these sorts with those of
    $\Sigma_H$. Similar to Example~\ref{ex:zoom1}, the domain
    attributes and actions of signature $\Sigma_H(T)$ are restricted
    to these basic sorts, and axioms of $\mathcal{D}_H(T)$ are those
    of $\mathcal{D}_H$ restricted to $\Sigma_H(T)$, e.g., they only
    include suitably ground instances of
    Statement~\ref{eqn:logician-causal}(a),
    Statement~\ref{eqn:logician-constraint}(b), and
    Statement~\ref{eqn:logician-executability}(a-c).  The signature
    $\Sigma_{LR}(T)$ of $\mathcal{D}_{LR}(T)$ has the following:
    \begin{itemize}
    \item Basic sorts $robot_L^z=\{rob_1\}$, $place_L^{z} =
      \{of\!\!fice, kitchen\}$ and $place_L^{*z} = \{c_i : c_i\in
      kitchen\cup of\!\!fice\}$.

    \item Functions (a) fluents $loc(robot_L^z)$ and
      $loc^*(robot_L^z)$ taking values from $place_L^z$ and
      $place_L^{*z}$ respectively, and $range(loc^*(robot_L^z),
      place_L^{*z})$; (b) statics such as $next\_to^*(place_L^{*z},
      place_L^{*z})$ and $next\_to(place_L^z, place_L^z)$; (c) fluent
      $broken(robot_L^z)$; and (d) $observed_{loc^*}(robot_L^z,
      robot_L^z, place_L^z)$ and other relevant knowledge fluents.

    \item Actions that include (a) $move^*(robot_L^z, place_L^{*z})$;
      and (b) $test_{loc^*}(robot_L^z, robot_L^z, place_L^{*z})$.
    \end{itemize}

    \medskip
    \noindent
    The axioms of $\mathcal{D}_{LR}(T)$ are those of
    $\mathcal{D}_{LR}$ restricted to $\Sigma_{LR}(T)$, e.g., they
    include:
    \begin{align*}
      &move^*(rob_1, c_j)~~\mathbf{causes}~~loc^*(rob_1)~=~\{C :
      range(loc^*(rob_1), C)\}\\
      &test_{loc^*}(rob_1, rob_1, c_j)~~{\bf causes }~~
      observed_{loc^*}(rob_1, rob_1, c_j)= \{true, false\}
      ~~{\bf if}~~ loc^*(rob_1) = c_j\\
      &\mathbf{impossible}~~move^*(rob_1, c_j) ~~\mathbf{if} ~~
      loc^*(rob_1)=c_i,~\neg next\_to^*(c_j, c_i)\\
      &\mathbf{impossible}~~move^*(rob_1, c_j) ~~\mathbf{if} ~~
      broken(rob_1)
    \end{align*}
    where $range(loc^*(rob_1), C)$ may hold for $C\in \{c_i, c_j,
    c_k\}$, which are within the range of the robot's current location
    ($c_i$), and are elements of $place_L^{*z}$. Assuming the robot is
    not broken, each state of $\tau_{LR}(T)$ thus includes an atom of
    the form $loc^*(rob_1)=c_i$, where $c_i$ is a cell in the
    $kitchen$ or the $of\!\!fice$, $\neg broken(rob_1)$, direct
    observations of this atom, e.g., $observed_{loc^*}(rob_1, rob_1,
    c_i) = true$, and statics such as $next\_to^*(c_i,c_j)$ etc.
    Specific actions include $move^*(rob_1, c_i)$ and
    $test_{loc^*}(rob_1, rob_1, c_i)$.
    
    As an extension to this example, if $rob_1$ is holding textbook
    $tb_1$ before executing $a^H = move(rob_1, kitchen)$, i.e.,
    $in\_hand(rob_1, tb_1) \in \sigma_1$, then $\Sigma_H(T)$ also
    includes $tb_1$ of sort $textbook$, and $\Sigma_{LR}(T)$ includes
    $object_L^z =\{tb_1\}$. The functions of $\mathcal{D}_{LR}(T)$
    include basic fluent $in\_hand(robot_L^z, object_L^z)$ and the
    corresponding knowledge fluents, and the actions and axioms are
    suitably restricted.  }
\end{example2}

\medskip
\noindent
In Examples~\ref{ex:zoom1} and~\ref{ex:zoom2}, probabilities are
assigned to the outcomes of actions based on the statistics collected
earlier (see Definition~\ref{def:learned-probs} in
Section~\ref{sec:arch-random}). For instance, if action $move(rob_1,
c_1)$ is executed, the probabilities of the possible outcomes of this
action may be:
\begin{align*}
  &P( loc^*(rob_1)= c_1 ) = 0.85\\
  &P( loc^*(rob_1)= Cl~|~range(loc^*(rob_1), Cl), Cl \neq c_1 ) =
  \frac{0.15}{|Cl|};~~\textrm{where}~~Cl=\{Cl: range(loc^*(rob_1),
  Cl),\, Cl\neq c_1\}
\end{align*}
Similarly, if the robot has to search for a textbook $cup_1$ once it
reaches the $kitchen$, and if a $test$ action is executed to determine
the location of a textbook $cup_1$ in cell $c_i$ in the $kitchen$, the
probabilities of the outcomes may be:
\begin{align*}
  P\Big( observed_{loc^*}(rob_1, cup_1, c_i) = true~\Big|~loc^*(cup_1) = c_i \Big) = 0.9 \\
  P\Big( observed_{loc^*}(rob_1, cup_1, c_i) =
  false~\Big|~loc^*(cup_1) = c_i \Big) = 0.1
\end{align*}
Also, when the execution of an action changes the value of a fluent
that is its indirect consequence, the probabilities are computed by
marginalizing the related fluents. For instance, the probability of a
cup being in a particular cell is computed by considering the
probability of the robot being in the cell and holding the cup, and
the probability of the cup being in the cell but not in the robot's
grasp.
% \commentm{Fix error in description}
% \begin{align*}
%   P(loc^*(cup_1) = c) =& P(loc^*(cup_1) = c~|~ in\_hand(rob_1,
%   cup_1))\cdot P(in\_hand(rob_1, cup_1)) +\\
%   &P(loc^*(cup_1) = c~|~ \lnot in\_hand(rob_1, cup_1))\cdot P(\lnot
%   in\_hand(rob_1, cup_1))
% \end{align*}

Given $\mathcal{D}_{LR}(T)$ and the probabilistic information, the
robot now has to execute a sequence of concrete actions that implement
the desired transition $T=\langle \sigma_1,a^H,\sigma_2\rangle$. For
instance, a robot searching for $cup_1$ in the $kitchen$ can check
cells in the $kitchen$ for $cup_1$ until either the cell location of
$cup_1$ is determined with high probability (e.g., $\ge 0.9$), or all
cells are examined without locating $cup_1$. In the former case, the
probabilistic belief can be elevated to a fully certain statement, and
the robot reasons about the action outcome and observations to infer
that $cup_1$ is in the $kitchen$, whereas the robot infers that
$cup_1$ is not in the $kitchen$ in the latter case. Such a
probabilistic implementation of an abstract action as a sequence of
concrete actions is accomplished by constructing and solving a POMDP,
and repeatedly invoking the corresponding policy to choose actions
until termination, as described below.

%%%%%-----------------------------------------------------------------------------
%%%%%-----------------------------------------------------------------------------
\section{POMDP Construction and Probabilistic Execution}
\label{sec:arch-pomdp}
In this section, we describe the construction of a POMDP
$\mathcal{P}_o(T)$ as a representation of the zoomed system description
$D_{LR}(T)$ and the learned probabilities of action outcomes
(Section~\ref{sec:arch-random}), and the use of $\mathcal{P}_o(T)$ for
the fine-resolution implementation of transition $T=\langle
\sigma_1,a^H,\sigma_2\rangle$ of $\tau_H$.  First,
Section~\ref{sec:arch-pomdp-overview} summarizes the use of a POMDP to
compute a policy for selecting one or more concrete actions that
implement any given abstract action $a^{H}$.
Section~\ref{sec:arch-pomdp-construct} then describes the steps of the
POMDP construction in more detail.

%%%%%-----------------------------------------------------------------------------
\subsection{POMDP overview}
\label{sec:arch-pomdp-overview}
A POMDP is described by a tuple $\mathcal{P}_o = \langle A^P, S^P,
b_0^P, Z^P, T^P, O^P, R^P \rangle$ for specific goal state(s). This
formulation of a POMDP builds on the standard
formulation~\cite{kaelbling:AI98}. Since the states and observations
of a POMDP are different from the definitions of these terms as used
in this paper, we begin by introducing some terminology.

We refer to each state represented by the POMDP as a \emph{p-state}.
% Before defining the tuple's elements, we first acknowledge that the
% define the notion of a \emph{p-state} and an \emph{observation} of
% the POMDP.
Recall that each state $\delta_x$ of the fine-resolution system
description $\mathcal{D}_{LR}(T)$ contains atoms formed of statics,
non-knowledge fluent terms and knowledge fluent terms.  There is a
many-to-one correspondence between states of $\mathcal{D}_{LR}(T)$,
and the p-states and observations of the POMDP $\mathcal{P}_o(T)$
constructed from $\mathcal{D}_{LR}(T)$. We provide the following
definition to avoid confusion as to whether we are referring to
$\mathcal{P}_o(T)$ or $\mathcal{D}_{LR}(T)$.

\begin{defn2}\label{def:pomdp-pstate-obs}[P-states and observations of
  POMDP $\mathcal{P}_o(T)$]\\
  {\rm Let $\mathcal{P}_o(T)$ be a POMDP constructed from the zoomed
    fine-resolution system description $\mathcal{D}_{LR}(T)$.
    \begin{itemize}
    \item Each p-state $s$ of $\mathcal{P}_o(T)$ is a projection of
      states of $\mathcal{D}_{LR}(T)$ on the set of atoms of the form
      $f(t) = y$, where $f(t)$ is a basic non-knowledge
      fine-resolution fluent term, or a special p-state called the
      \emph{terminal p-state}. % We write:
%       \begin{align*}
%         s = Proj(\mathbf{\delta}, \mathcal{P}_o_{pbf}) = \{ f(t)=y :
%         f(t) \in \mathbf{\delta} \land f(t) \in \mathcal{P}_o_{pbf} \}
%       \end{align*}

    \item Each observation $z$ of $\mathcal{P}_o(T)$ is a projection
      of states of $\mathcal{D}_{LR}(T)$ on the set of atoms of basic
      knowledge fluent terms corresponding to the robot's observation
      of the possible values of fine-resolution fluent terms such as
      $observed_{f^*}(robot, x, y) = outcome$, where $y$ is a possible
      value of the fluent term $f^*(x)$. For simplicity, we use the
      observation \emph{none} to replace all instances that have
      $undet$ as the $outcome$. % We write:
%       \begin{align*}
%         z = Proj(\mathbf{\delta}, \mathcal{P}_o_{kbf}) = \{ f(t)=y :
%         f(t) \in \mathbf{\delta} \land f(t) \in \mathcal{P}_o_{kbf} \}
%       \end{align*}
    \end{itemize}
    In other words, the p-states (observations) of $\mathcal{P}_o(T)$
    are obtained by dropping the atoms formed of knowledge
    (non-knowledge) fluent terms and statics in the states of
    $\mathcal{D}_{LR}(T)$. }
\end{defn2}

% \item $S^P$: set of p-states $s_x \cup s_w$, where $s_x = Proj(
%   \delta_x, \mathcal{P}_o_{pbf} )$.
% \item $Z^P$: set of observations $z_x = Proj ( \delta_x,
%   \mathcal{P}_o_{kbf} )$

\noindent
We can now define the elements of a POMDP tuple:
\begin{itemize}
\item $A^P$: set of concrete, fine-resolution actions available to the
  robot.

\item $S^P$: set of p-states to be considered for probabilistic
  implementation of $a^{H}$. % A p-state is a projection of states of
%   $\mathcal{D}_{LR}(T)$ on the set of atoms of the form $f(t) = y$,
%   where $f(t)$ is a basic non-knowledge fine-resolution fluent term,
%   or a special p-state called the \emph{terminal p-state}. We use the
%   term ``p-state'' to differentiate between the states represented by
%   the POMDP and the definition of state we use in this paper.

\item $b_0^P$: initial \emph{belief state}, where a belief state is a
  probability distribution over $S^P$.

\item $Z^P$: set of observations. % An observation is a projection of
%   states of $\mathcal{D}_{LR}(T)$ on the set of atoms of basic
%   knowledge fluent terms corresponding to the robot's observation of
%   the value of a fine-resolution fluent term, e.g.,
%   $directly\_observed(robot, f(t), y) = outcome$, where $y$ is a
%   possible outcome of the fluent term $f(t)$. For simplicity, we use
%   the observation \emph{none} to replace all instances that have
%   $undet$ as the $outcome$.

\item $T^P: S^P\times A^P\times S^P \to [0, 1]$, the transition
  function, which defines the probability of each transition from one
  p-state to another when particular actions are executed. As
  described later, impossible state transitions are not included in
  $T^P$.

\item $O^P: S^P\times A^P\times Z^P \to [0, 1]$, the observation
  function, which defines the probability of obtaining particular
  observations when particular actions are executed in particular
  p-states. As described later, one valid state-action-observation
  combinations are included in $O^P$.

\item $R^P: S^P\times A^P \times S^P \to \Re$, the reward
  specification, which encodes the relative immediate reward (i.e.,
  numerical value) of taking specific actions in specific p-states.
\end{itemize}
The p-states are considered to be \emph{partially observable} because
they cannot be observed with complete certainty, and the POMDP reasons
with probability distributions over the p-states, called \emph{belief
  states}. In this formulation, the belief state is a sufficient
statistic that implicitly captures all the information in the history
of observations and actions.

% is only based on a system description and does not include any
% history of observations and actions---in a standard POMDP
% formulation, the current p-state is assumed to be the result of all
% information obtained in previous time steps, i.e., the belief state
% is assumed to implicitly include the history of observations and
% actions.

The use of a POMDP has two phases (1) policy computation; and (2)
policy execution. The first phase computes \emph{policy} $\pi^P:
B^P\to A^P$ that maps belief states to actions, using an algorithm
that maximizes the utility (i.e., expected cumulative discounted
reward) over a planning horizon---we use a point-based approximate
solver that only computes beliefs at a few samples points in the
belief space~\cite{Ong:ijrr2010}. In the second phase, the computed
policy is used to repeatedly choose an action in the current belief
state, updating the belief state after executing the action and
receiving an observation. This belief revision is based on Bayesian
updates:
%\vspace{-0.5em}
\begin{align}
  \label{eqn:belief-update}
  b_{t+1}^P(s_{t+1}^P) \propto O^P(s_{t+1}^P, a_{t+1}^P,
  o_{t+1}^P)\sum_{s_t^P}\{ T^P(s_t^P,a_{t+1}^P,s_{t+1}^P)\cdot
  b_t^P(s_t^P)\}
  %\vspace{-1em}
\end{align}
where $b_{t+1}^P$, $s_{t+1}$, $a_{t+1}$ and $o_{t+1}$ are the belief
state, p-state, action and observation (respectively) at time $t+1$.
Equation~\ref{eqn:belief-update} says that $b_{t+1}^P$ is proportional
to the product of the terms on the right hand side. The belief update
continues until policy execution is terminated. In our case, policy
execution terminates when doing so has a higher (expected) utility
than continuing to execute the policy. This happens when either the
belief in a specific p-state is very high (e.g., $\ge 0.8$), or none
of the p-states have a high probability associated with them after
invoking the policy several times---the latter case is interpreted as
a failure to execute the coarse-resolution action under consideration.
% We will use ``POMDP-1'' to refer to the process of constructing a
% POMDP, computing the policy, and using this policy to implement the
% desired abstract action.

%%%%%-----------------------------------------------------------------------------
\subsection{POMDP construction}
\label{sec:arch-pomdp-construct}
Next, we describe the construction of POMDP $\mathcal{P}_o(T)$ for the
fine-resolution probabilistic implementation of coarse-resolution
transition $T = \langle \sigma_1, a^{H}, \sigma_2\rangle\in \tau_H$,
using $\mathcal{D}_{LR}(T)$ and the statistics collected in the
training phase as described in Section~\ref{sec:arch-random}. We
illustrate these steps using examples based on the domain described in
Example~\ref{ex:illus-example}, including the example described in
Appendix~\ref{sec:appendix-pomdp}.

\medskip
\noindent
\textbf{\underline{Actions:}} the set $A^P$ of actions of
$\mathcal{P}_o(T)$ consists of concrete actions from the signature of
$\mathcal{D}_{LR}(T)$ and new \emph{terminal actions} that terminate
policy execution. We use a single terminal action---if $A^P$ is to
include domain-specific terminal actions, it is the designer's
responsibility to specify them. For the discussion below, it will be
useful to partition $A^P$ into three subsets (1) $A_1^P$, actions that
cause a transition between p-states; (2) $A_2^P$, knowledge-producing
actions for testing the values of fluents; and (3) $A_3^P$, terminal
actions that terminate policy execution. The example in
Appendix~\ref{sec:appendix-pomdp} includes (a) actions from $A_1^P$
that move the robot to specific cells, e.g., \stt{move-0} and
\stt{move-1} cause robot to move to cell $0$ and $1$ respectively, and
the action $grasp(rob_1, tb_1)$; (b) $test_{loc^*}$ actions from
$A_2^P$ to check if the robot or target object ($tb_1$) are in
specific cells; and (c) action $finish$ from $A_3^P$ that terminates
policy execution.

\medskip
\noindent
\textbf{\underline{P-states, initial belief state and observations:}}
the following steps are used to construct $S^P$, $Z^P$ and $b_0^P$.
\begin{enumerate}
\item 
  Construct ASP program $\Pi_c(\mathcal{D}_{LR}(T)) \cup Q$.  Here,
  $\Pi_c(\mathcal{D}_{LR}(T))$ is constructed as described in
  Definition~\ref{def:ald-state} (Section~\ref{sec:arch-ald}), and $Q$
  is a collection of (a) atoms formed by statics; and (b) disjunctions
  of atoms formed by basic fluent terms. Each disjunction is of the
  form $\{f(t) = y_1~\lor~\ldots~\lor~f(t)=y_n\}$, where
  $\{y_1,\ldots,y_n\}$ are possible values of basic fluent term
  $f(t)$. Observe that if $AQ$ is an answer set of $Q$, then there is
  a state $\delta$ of $\mathcal{D}_{LR}(T)$ such that
  $AQ=\delta^{nd}$; also, for every state $\delta$ of
  $\mathcal{D}_{LR}(T)$, there is an answer set $AQ$ of $Q$ such that
  $AQ=\delta^{nd}$. It can be shown that $AS$ is an answer set of
  $\Pi_c(\mathcal{D}_{LR}(T)) \cup Q$ iff it is an answer set of
  $\Pi_c(\mathcal{D}_{LR}(T)) \cup AQ$ where $AQ$ is an answer set of
  $Q$. This statement follows from the definition of answer set and
  the splitting set theorem~\cite{balduccini:lpnmr09}.

%   \commentm{Answer sets of this Q corresponds to all possible sets of
%     $\sigma^nd$ for Defn. 2. It can be shown that AS is an answer set
%     of ...}

\item Compute answer set(s) of ASP program $\Pi_c(\mathcal{D}_{LR}(T))
  \cup Q$. Based on the observation in Step-1 above, and the
  well-foundedness of $\mathcal{D}_{LR}(T)$, it is easy to show that
  each answer set is unique and is a state of $\mathcal{D}_{LR}(T)$.

\item From each answer set, extract all atoms of the form $f(t) = y$,
  where $f(t)$ is a basic non-knowledge fine-resolution fluent term,
  to obtain an element of $S^P$. Basic fluent terms corresponding to a
  coarse-resolution domain attribute, e.g., room location of the robot,
  are not represented probabilistically and thus not included in
  $S^P$. We refer to such a projection of a state $\delta$ of
  $\mathcal{D}_{LR}(T)$ as the \emph{p-state defined by} $\delta$.
  Also include in $S^P$ an ``absorbing'' terminal p-state $absb$ that
  is reached when a terminal action from $ A_3^P$ is executed.

\item From each answer set, extract all atoms formed by basic
  knowledge fluent terms corresponding to the robot sensing a
  fine-resolution fluent term's value, to obtain elements of $Z^P$,
  e.g., $directly\_observed(robot, f(t), y) = outcome$. We refer to
  such a projection of a state $\delta$ of $\mathcal{D}_{LR}(T)$ as an
  \emph{observation defined by} $\delta$. As described earlier, for
  simplicity, observation $none$ replaces all instances in $Z^P$ that
  have $undet$ as the $outcome$.

\item In general, the initial belief state $b_0^P$ is a uniform
  distribution, i.e., all p-states are considered to be equally
  likely. This does not prevent the designer from using other priors,
  but these priors would have to be derived from sources of knowledge
  external to our architecture.
\end{enumerate}
In the example in Appendix~\ref{sec:appendix-pomdp}, abstract action
$grasp(rob_1, tb_1)$ has to be executed in the $of\!\!fice$. To do so,
the robot has to move and find $tb_1$ in the $of\!\!fice$.
Example~\ref{ex:zoom1} above contains the corresponding
$\mathcal{D}_{LR}(T)$. Here, $Q$ includes (a) atoms formed by statics,
e.g., $next\_to^*(c_1, c_2)$ where $c_1$ and $c_2$ are neighboring
cells in the $of\!\!fice$; and (b) disjunctions such as
$\{loc^*(rob_1) = c_1~\lor~\ldots~\lor~loc^*(rob_1)=c_n\}$ and
$\{loc^*(tb_1) = c_1~\lor~\ldots~\lor~loc^*(tb_1)=c_n\}$, where
$\{c_1, \ldots, c_n\} \in of\!\!fice$. In Step 3, p-states such as
$\{loc^*(rob_1) = c_1,~loc^*(tb_1) = c_1,~\neg in\_hand(rob_1,
tb_1)\}$ are extracted from the answer sets.  In Step 4, observations
such as $observed_{loc^*}(rob_1, rob_1, c_1) = true$ and
$observed_{loc^*}(rob_1, tb_1, c_1) = false$ are extracted from the
answer sets. Finally, the initial belief state $b_0^P$ is set as a
uniform distribution (Step 5).

\medskip
\noindent 
\textbf{\underline{Transition function and observation function:}}
next, we consider the construction of $T^P$ and $O^P$ from
$\mathcal{D}_{LR}(T)$ and the statistics collected in the initial
training phase (see Section~\ref{sec:arch-random}).

A transition between p-states of $\mathcal{P}_o(T)$ is
defined as $\langle s_i, a, s_j \rangle \in T^P$ iff there is an
action $a \in A_1^P$ and a transition $\langle \delta_x, a, \delta_y
\rangle$ of $\mathcal{D}_{LR}(T)$ such that $s_i$ and $s_j$ are
p-states defined by $\delta_x$ and $\delta_y$ respectively. The
probability of $\langle s_i, a, s_j \rangle \in T^P$ equals the
probability of $\langle \delta_x, a,\delta_y \rangle$. In a similar
manner, $\langle s_i, a, z_j \rangle \in O^P$ iff there is an action
$a \in A_2^P$ and a transition $\langle \delta_x, a, \delta_y \rangle$
of $\mathcal{D}_{LR}(T)$ such that $s_i$ and $z_j$ are a p-state and
an observation defined by $\delta_x$ and $\delta_y$ respectively.  The
probability of $\langle s_i, a, z_j \rangle \in O^P$ equals the
probability of $\langle \delta_x, a, \delta_y \rangle$.

\smallskip
\noindent
First, we augment $\mathcal{D}_{LR}(T)$ with causal laws for proper
termination:
\begin{align*}
  finish~~\mathbf{causes}~~absb\\
  \mathbf{impossible}~~A^P~~\mathbf{if}~~absb
\end{align*}
Next, we note that actions in $A_1^P$ cause p-state transitions but
provide no observations, while actions in $A_2^P$ do not cause p-state
changes but provide observations, and terminal actions in $A_3^P$
cause transition to the absorbing state and provide no observations.
To use state of the art POMDP solvers, we need to represent $T^P$ and
$O^P$ as a collection of tables, one for each action. Specifically,
$T_a^P[s_i, s_j]=p$ iff $\langle s_i, a, s_j \rangle \in T^P$ and its
probability is $p$. Also, $O_a^P[s_i, z_j]=p$ iff $\langle s_i, a, z_j
\rangle \in O^P$ and its probability is $p$.
Algorithm~\ref{alg:pomdpTO} describes the construction of $T^P$ and
$O^P$.

\begin{algorithm}[htb]
  \caption{Constructing POMDP transition function $T^P$ and
    observation function $O^P$}
  \label{alg:pomdpTO}
  
  \Indm

  \KwIn{$S^P$, $A^P$, $Z^P$, $\mathcal{D}_{LR}(T)$; transition
    probabilities for actions $\in A_1^P$; observation probabilities
    for actions $\in A_2^P$.}

  \KwOut{POMDP transition function $T^P$ and observation function
    $O^P$.}
  
  \Indp \BlankLine \SetNoFillComment
  
  Initialize $T^P$ as $|S^P|\times |S^P|$ identity matrix for each
  action.

  Initialize $O^P$ as $|S^P|\times |Z^P|$ matrix of zeros for each
  action.
  
  \BlankLine

  \tcp*[h]{Handle special cases}

  \For{each $a_j\in A_3^P$} {
    $T_{a_j}^P(*, absb) = 1$ 
  
    $O_{a_j}^P(*, none) = 1$ 
  }

  \For{each action $a_j\in A_1^P$} {
    $O_{a_j}^P(*, none) = 1$ 
  }
  
  \For{each $a_j\in A^P$} {$O_{a_j}^P(absb, none) = 1$}
  
  \BlankLine
  \tcp*[h]{Handle normal transitions}

  \For{each p-state $s_i \in S^P$} {
 
    \BlankLine

    \tcp*[h]{Construct and set probabilities of p-state transitions}

    Construct ASP program $\Pi(\mathcal{D}_{LR}(T), s_i,
    Disj(A_1^P))$.
      
    Compute answer sets $\mathbf{AS}$ of ASP program.

    From each $AS\in \mathbf{AS}$, extract p-state transition $\langle
    s_i, a_k, s_j \rangle$, and set the probability of $T_{a_k}^P[s_i,
    s_j]$.
 
    \BlankLine

    \tcp*[h]{Construct and set probabilities of observations}

    Construct ASP program $\Pi(\mathcal{D}_{LR}(T), s_i,
    Disj(A_2^P))$.
      
    Compute answer sets $\mathbf{AS}$ of ASP program.

    From each $AS\in \mathbf{AS}$, extract triple $\langle s_i, a_k,
    z_j \rangle$, and set value of $O_{a_k}^P[s_i, z_j]$.  }
  
  \Return $T^P$ and $O^P$.
\end{algorithm}

\medskip
\noindent
Some specific steps of Algorithm~\ref{alg:pomdpTO} are elaborated
below.
\begin{itemize}
\item After initialization, Lines 3--12 of Algorithm~\ref{alg:pomdpTO}
  handle special cases. For instance, any terminal action will cause a
  transition to the terminal p-state and provide no observations
  (Lines 4-5).

\item An ASP program of the form $\Pi(\mathcal{D}_{LR}(T), s_i,
  Disj(A))$ (Lines 12, 15) is defined as $\Pi(\mathcal{D}_{LR}(T))\cup
  val(s_i, 0)\cup Disj(A)$. Here, $Disj(A)$ is a disjunction of the
  form $\{occurs(a_1, 0)\lor\ldots\lor occurs(a_n, 0)\}$, where
  $\{a_1,\ldots, a_n\}\in A$. Lines 14-16 construct and compute answer
  sets of such a program to identify all possible p-state transitions
  as a result of actions in $A_1^P$. Then, Lines 17-19 construct and
  compute answer set of such a program to identify possible
  observations as a result of actions in $A_2^P$.

\item Line 16 extracts a statement of the form $occurs(a_k \in A_1^P,
  0)$, and p-state $s_j\in S^P$, from each answer set $AS$, to obtain
  p-state transition $\langle s_i, a_k, s_j\rangle$.  As stated
  earlier, a p-state is extracted from an answer set by extracting
  atoms formed by basic non-knowledge fluent terms.

\item Line 19 extracts a statement of the form $occurs(a_j \in A_2^P,
  0)$, and observation $z_j\in Z^P$, from each answer set $AS$, to
  obtain triple $\langle s_i, a_k, z_j \rangle$. As described earlier,
  an observation is extracted from an answer set by extracting atoms
  formed by basic knowledge fluent terms.

\item Probabilities of p-state transitions are set (Line 16) based on
  the corresponding physical state transition probabilities (first
  type of transition in Definition~\ref{def:learned-probs} in
  Section~\ref{sec:arch-random}). Probabilities of observations are
  set (Line 19) based on the knowledge state transition probabilities
  (second type of transition in Definition~\ref{def:learned-probs}).
\end{itemize}
In the example in Appendix~\ref{sec:appendix-pomdp}, a robot in the
$of\!\!fice$ has to pick up textbook $tb_1$ believed to be in the
$of\!\!fice$. This example assumes that a $move$ action from one cell
to a neighboring cell succeeds with probability $0.95$---with
probability $0.05$ the robot remains in its current cell. It is also
assumed that with probability $0.95$ the robot observes (does not
observe) the textbook when it exists (does not exist) in the cell the
robot is currently in. The corresponding $T^P$ and $O^P$, constructed
for this example, are shown in Appendix~\ref{sec:appendix-pomdp}.

\medskip
\noindent
The correctness of the approach used to extract p-state transitions
and observations, in Lines 16, 19 of Algorithm~\ref{alg:pomdpTO}, is
based on the following propositions.
%\vspace{1em}
%\hrule
\begin{prop2}\label{prop:pomdpT}[Extracting p-state transitions
  from answer sets]
  {\rm
    \begin{itemize}
    \item If $\langle s_i, a, s_j \rangle \in T^P$ then there is an
      answer set $AS$ of program $\Pi(\mathcal{D}_{LR}(T), s_i,
      Disj(A_1^P))$ such that $s_j = \{f(\bar{x})=y : f(\bar{x})=y \in
      AS~~\textrm{and } f \textrm{ is basic}\}$.
    \item For every answer set $AS$ of program
      $\Pi(\mathcal{D}_{LR}(T), s_i, Disj(A_1^P))$ and $s_j =
      \{f(\bar{x})=y : f(\bar{x})=y \in AS~~\textrm{and } f \textrm{
        is basic}\}$, $\langle s_i, a, s_j \rangle \in T^P$.
    \end{itemize} 
  }
\end{prop2}
%\vspace{1pt}
%\hrule
\begin{prop2}\label{prop:pomdpO}[Extracting observations from answer
  sets]
  {\rm 
    \begin{itemize}
    \item If $\langle s_i, a, z_j \rangle \in O^P$ then there is an
      answer set $AS$ of program $\Pi(\mathcal{D}_{LR}(T), s_i,
      Disj(A_2^P))$ such that $z_j = \{f(\bar{x})=y : f(\bar{x})=y \in
      AS~~\textrm{and } f \textrm{ is basic}\}$.
    \item For every answer set $AS$ of program
      $\Pi(\mathcal{D}_{LR}(T), s_i, Disj(A_1^P))$ and $z_j =
      \{f(\bar{x})=y : f(\bar{x})=y \in AS~~\textrm{and } f \textrm{
        is basic}\}$, $\langle s_i, a, z_j \rangle \in O^P$.
    \end{itemize} 
  }
\end{prop2}
%\vspace{1pt}
%\hrule
\medskip
\noindent
Proposition~\ref{prop:pomdpT} says that a transition between p-states
is in $\mathcal{P}_o(T)$ \emph{iff} a matching transition is in
$\mathcal{D}_{LR}(T)$, and that for any state transition in
$\mathcal{D}_{LR}(T)$ a matching p-state transition is in
$\mathcal{P}_o(T)$. Proposition~\ref{prop:pomdpO} makes a similar
statement about observations of $\mathcal{P}_o(T)$. These propositions
are true by construction, and they help establish that every state
transition in $\mathcal{D}_{LR}(T)$ can be achieved by a sequence of
actions and observations in $\mathcal{P}_o(T)$.

\begin{algorithm}[tb]
  \caption{Construction of POMDP reward function $R^P$}
  \label{alg:pomdpR}
  \Indm 

  \KwIn{$S^P$, $A^P$, and $T^P$; statistics regarding accuracy and
    time taken to execute non-terminal actions.}

  \KwOut{Reward function $R^P$.}

  \Indp
  \BlankLine
  \SetNoFillComment
  \tcp*[h]{Consider each possible p-state transition}

  \For{each $(s, a, s')\in S^P\times A^P\times S^P$ with $T^P(s, a, s')\neq 0$} {
    \BlankLine
    \tcp*[h]{Consider terminal actions first}

    \uIf{$a \in A_3^P$}{ 
      \uIf{$s'$ is a goal p-state} { 
        $R^P(s, a, s') = $ large positive value.
      }
      \Else {
        $R^P(s, a, s') =$ large negative value.
      }
    }

    \BlankLine
    \tcp*[h]{Rewards are costs for non-terminal actions}

    \Else{
      Set $R^P(s, a, s')$ based on relative computational effort and
      accuracy.
    }
  }
  
  \Return $R^P$
\end{algorithm}

\medskip
\noindent
\textbf{\underline{Reward specification:}} the reward function $R^P$
assigns a real-valued reward to each p-state transition, as described
in Algorithm~\ref{alg:pomdpR}. Specifically, for any state transition
with a non-zero probability in $T^P$:
\begin{enumerate}
\item If it involves a terminal action from $A_3^P$, the reward is a
  large positive (negative) value if this action is chosen after
  (before) achieving the goal p-state.
\item If it involves non-terminal actions, reward is a real-valued
  cost (i.e., negative reward) of action execution.
\end{enumerate}
% Each non-terminal action (by design) only influences one basic fluent.
Here, any p-state $s\in S^P$ defined by state $\delta$ of
$\mathcal{D}_{LR}(T)$ that is a refinement of $\sigma_2$ in transition
$T=\langle \sigma_1, a^{H}, \sigma_2\rangle$ is a goal p-state. In
Appendix~\ref{sec:appendix-pomdp}, we assign large positive reward
($100$) for executing $finish$ when textbook $tb_1$ is in the robot's
grasp, and large negative reward ($-100$) for terminating before
$tb_1$ has been grasped (Lines 3-7, Algorithm~\ref{alg:pomdpR}). We
assign a fixed cost ($-1$) for all other (i.e., non-terminal) actions
(Line 9). When necessary, this cost can be a heuristic function of
relative computational effort and accuracy, using domain expertise and
statistics collected experimentally, e.g., we can set $R^P(*, shape,
*)=-1$ and $R^P(*, color, *)=-2$ because statistics indicate that the
knowledge-producing action that determines an object's color takes
twice as much time as the action that determines the object's shape.
Although we do not do so in our example, it is also possible to assign
high cost (i.e., large negative reward) to transitions that should be
avoided or are dangerous, e.g., actions that take a wheeled robot near
a flight of stairs. The reward function, in turn, influences the (a)
rate of convergence during policy computation; and (b) accuracy of
results during policy execution. Appendix~\ref{sec:appendix-pomdp}
describes the reward function for a specific example.

\medskip
\noindent
\textbf{\underline{Computational complexity and efficiency:}}
% The randomized fine-resolution transition diagram and the
% probabilities of action outcomes may be used with other formulations
% for probabilistic reasoning~\cite{Koller:gmbook10}, including
% probabilistic extensions of ASP~\cite{baral:TPLP09,lee:aaaisss15},
% which can be used for computing a sequence of actions to achieve a
% given goal.
Let us consider the complexity of solving POMDPs and or our approach
to construct the POMDPs. For exact algorithms (i.e., algorithms that
solve POMDPs optimally), the complexity class of infinite-horizon
stochastic-transition POMDPs with boolean rewards is known to be
\emph{EXPTIME}; for polynomial time-bounded POMDPs, the complexity
class improves to \emph{PSPACE}~\cite{littman:thesis96}.  Approximate
belief-point approaches, which we employ here, are more efficient. In
these the complexity of one backup (i.e., one step of belief update)
across all belief points is given by~\cite{shani:JAAMAS13}:
\begin{align}
  \label{eqn:pomdp-approx}
  O(|A^P| \times |Z^P| \times |V^P| \times |S^P|^2 + |B^P| \times
  |A^P| \times |S^P| \times |Z^P|)
\end{align}
where $B^P$ is the set of belief points. This compares favorably with
the complexity of one backup across all $\alpha$-vectors\footnote{The
  $\alpha$-vectors are hyperplanes computed in belief space and used
  to select the appropriate action to be executed in any given belief
  state.}  for exact algorithms, which is~\cite{shani:JAAMAS13}:
\begin{align}
  \label{eqn:pomdp-exact}
  O(|A^P| \times |Z^P| \times |V^P| \times |S^P|^2 + |A^P| \times
  |S^P| \times |V^P|^{|Z^P|})
\end{align}
where $V^P$ is the set of $\alpha$-vectors. For more details about the
complexity of POMDP solvers, please see~\cite{shani:JAAMAS13}.

Even the (approximate) belief point algorithms are susceptible to
problem size, with the best solvers able to tackle problems with a few
hundred p-states (i.e., $|S^P| \simeq 100$) if both the transition and
observation functions are stochastic, as they are here\footnote{There
  are solvers, such as POMCP, which work on very large state spaces,
  but which have not had demonstrable results on problems that show
  scaling with both stochastic transitions and
  observations~\cite{silver:nips10}.}. Thus, there is an advantage in
reducing the problem to be tackled by the solver. In a POMDP created
from a relational representation, such as employed here, this is
particularly critical. In general, if we have $m$ fluents, each with
an average of $k$ values, $|S^P|=k^m$. In our approach, domain
knowledge and prior information (e.g., defaults encoded in ASP program
at coarse-resolution) remove a proportion of atoms formed of fluent
literals from consideration during zooming. If we model the remaining
proportion of fluent literals as $0 < \beta < 1$, then clearly
$|S^P|=k^{\beta m}$. As indicated by Equation~\ref{eqn:pomdp-approx},
this reduction can provide significant computational benefits,
especially in more complex domains where many more fluents are likely
to be irrelevant to any given transition, e.g., if only two out of a
$100$ atoms are relevant, $|S^P| = k^{0.02 m}$.

For specific tasks such as path planning, it may also be possible to
use specific heuristic or probabilistic algorithms that are more
computationally efficient than a POMDP.  However, POMDPs provide a (a)
principled and quantifiable trade-off between accuracy and
computational efficiency in the presence of uncertainty in both
sensing and actuation; and (b) near-optimal solution if the POMDP's
components are modeled correctly. The computational efficiency of
POMDPs can also be improved by incorporating hierarchical
decompositions, or by dividing the state estimation problem into
sub-problems that model actions and observations influencing one
fluent independent of those influencing other fluents---we have
pursued such options in other work~\cite{zhang:TRO13}. These
approaches are not always possible, e.g., when a robot is holding a
textbook in hand, the robot's location and the textbook's location are
not independent.  Instead, in our architecture, we preserve such
constraints and construct a POMDP for the relevant part of the domain
to significantly reduce the computational complexity of solving the
POMDP. Furthermore, many of the POMDPs required for a given domain can
be precomputed, solved and reused. For instance, if the robot has
constructed a POMDP for locating a textbook in a room, the POMDP for
locating a different book (or even a different object) in the same
room may only differ in the values of some transition probabilities,
observation probabilities, and rewards. This similarity between tasks
may not hold in non-stationary domains, in which the elements of the
POMDP tuple (e.g., set of p-states) and the collected statistics
(e.g., transition probabilities) may need to be revised over time.

Our algorithms for constructing the POMDP $\mathcal{P}_o(T)$ for a
specific coarse-resolution transition have two key steps: (1)
construction of matrices that represent the functions for transition,
observation and reward; and (2) computing answer sets of specific ASP
programs to identify valid transitions, observations etc. The first
step is polynomial in the size of $S^P$ and $Z^P$ ($|S^P|$ is usually
bigger than $|Z^P|$). The second step, which involves grounding the
domain attributes and then computing possible answer sets, can (in the
worst case) be exponential in the size of (ground)
atoms~\cite{gebser:aspbook12}\footnote{In many modern ASP solvers
  based on SAT algorithms, the exponential factor is a small number
  greater than 1.  We can also use solvers that do incremental
  grounding~\cite{gebser:lpnmr15}.}.  Recall that we only consider
object constants relevant to the transition under consideration (see
Section~\ref{sec:arch-zoom} on zooming). This, in conjunction with the
fact that we reuse POMDPs when possible (as described above), makes
the construction of $\mathcal{P}_o(T)$ computationally efficient.

\medskip
\noindent
\underline{\textbf{Computational error:}} Although the
outcomes of POMDP policy execution are non-deterministic, following an
optimal policy produced by an exact POMDP solver is most likely (among
all such possible policies) to take the robot to a goal p-state if the
following conditions hold:
\begin{itemize}
\item The coarse-resolution transition diagram $\tau_H$ of the domain
  has been constructed correctly;
\item The statistics collected in the initial training phase
  (Section~\ref{sec:arch-random}) correctly model the domain dynamics;
  and
\item The reward function is constructed to suitably reward desired
  behavior.
\end{itemize}
This statement is based on existing
literature~\cite{kaelbling:AI98,littman:thesis96,Sondik:thesis71}. We
use an approximate POMDP solver for computational efficiency, and an
exact belief update (Equation~\ref{eqn:belief-update}), which provides
a bound on the regret (i.e., loss in value) achieved by following the
computed policy in comparison with the optimal
policy~\cite{Ong:ijrr2010}. \emph{We can thus only claim that the
  outcomes of executing of our policy are approximately correct with
  high probability}. We can also provide a bound on the margin of
error~\cite{Ong:ijrr2010}, i.e. the probability that at least one
incorrect commitment is made to the history. For instance, if the
posterior probability associated with a statement $observed_{f}(R, X,
Y)$ in the fine-resolution representation is $p$, the probability of
error in the corresponding commitment made to the history
$\mathcal{H}$ (in the coarse-resolution representation) based on this
statement is $(1-p)$. If a series of statements with probabilities
$p_i$ are used to arrive at a conclusion that is committed to
$\mathcal{H}$, $(1-\prod_i p_i)$ is the corresponding probability that
at least one erroneous commitment has been made. If a later commitment
$j$ is based on a prior belief influenced by previous commitment $i$
then $p_j$ is a conditional probability, conditioned on that previous
commitment.

\begin{figure}[tb]
  \begin{center}
    \includegraphics[width=0.65\columnwidth]{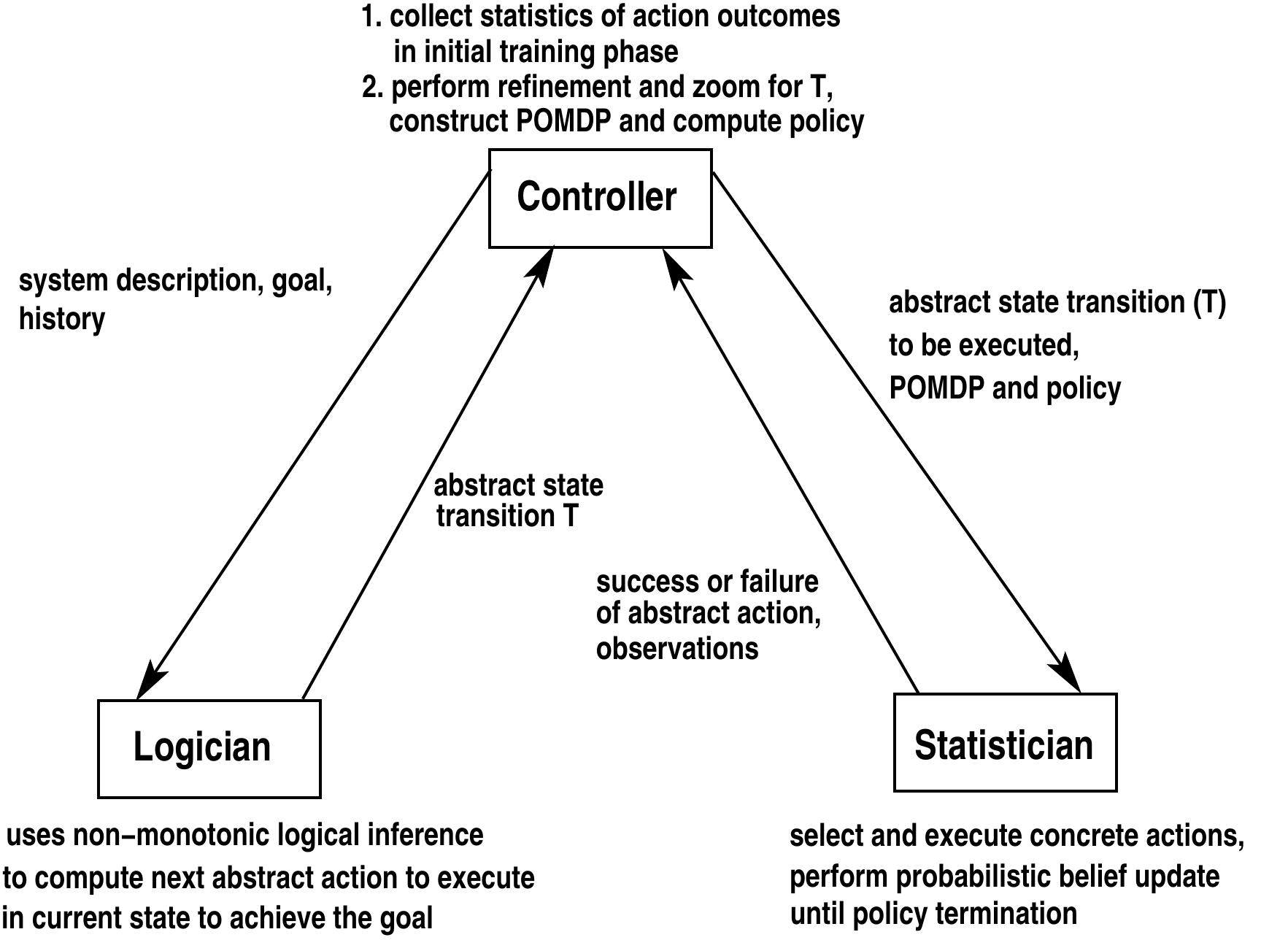}
  \end{center}
  \vspace{-1em}
  \caption{The proposed architecture can be viewed as a
    \emph{logician} and a \emph{statistician} communicating through a
    \emph{controller}.  The architecture combines the complementary
    strengths of declarative programming and probabilistic models.}
  \label{fig:arch-overview}
  %\vspace{-1em}
\end{figure}

%%%%%-----------------------------------------------------------------------------
%%%%%-----------------------------------------------------------------------------
%\vspace*{-0.5em}
\section{Reasoning System and Control Loop of Architecture}
%\vspace*{-0.75em}
\label{sec:arch-overall}
Recall that the fifth step of the design methodology (see
Section~\ref{sec:methodology}) is to: 
\medskip
\noindent
\begin{center}
  \fbox{%
    \parbox{0.95\textwidth}{%
      5. Provide the components described above, together with any
      desired goal, to a reasoning system that directs the robot
      towards achieving this goal. } }
\end{center}

% done := false.

% while not done do

%    Find a plan A1,...An from Sigma to a final state
%    If no plan exist return failure

%    i := 1; possible := true;
%    While possible do 
%        possible :- execute(Ai, Sigma)  % executes Ai and updates Sigma
%        if i=n done := true

\begin{algorithm}[tbh]
  \caption{Control loop}
  \label{alg:control-loop}
  %\DontPrintSemicolon
  %\dontprintsemicolon
  
  \KwIn{coarse-resolution system description $\mathcal{D}_H$ and
    history $\mathcal{H}$; its randomized refinement
    $\mathcal{D}_{LR}$; coarse-resolution description of goal $G$;
    coarse-resolution initial state $\sigma$.  }
  
  \KwOut{Return success/true, indicating that with high probability
    the robot is in a state satisfying $G$; return failure/false,
    indicating that with high probability achieving $G$ is
    impossible.}
  
  \SetKwProg{Func}{Function}{}{}
  
  %\Indp
  \BlankLine
  \Func{Main()} {
    
    done := false

    currState := $\sigma$

    \While{$\lnot$ done} { 
      
      Logician extracts a plan of abstract actions
      $a_1^{H},\ldots,a_n^{H}$ for $G$ from the answer set of $Plan :=
      \Pi^n(\mathcal{D}_H, \mathcal{H}) \cup Classical\_plan \cup DPC$
      (see Proposition~\ref{prop:plan-reduce})
      
      \If{no plan found} {
        \Return failure
      }
      
      done := Implement($a_1^{H},\ldots,a_n^{H}, currState, \mathcal{H}$)
    }
    \Return success
  }
  
  \BlankLine

  \tcp{Function below implements plan of abstract actions; records
    successful execution of action and observations in $\mathcal{H}$
    and sets $currState$ to current state of the system; returns
    true/success if entire sequence is implemented, otherwise returns
    false/failure}

  \Func{Implement($a_1^{H},\ldots,a_n^{H}, currState, \mathcal{H}$) } {

    i := 1

    executable := true

    \While{$(i \le n)$ $\land$  executable} {
%    \tcp*[h]{Implement next coarse-resolution transition}
      currAction := $a_i^H$
      
      executable := Implement($currState, currAction, \mathcal{H}$)
      
      \uIf{executable} {
        i := i+1
      }       
      
    } % end of while(executable)...
    \Return executable
  }% end of implement(sequence)
  
\end{algorithm}
%\SetNlSty{textbf}{(}{)}
%\SetNlSty{textbf}{}{}

\begin{algorithm}[tbh]
  %\BlankLine
  \setcounter{AlgoLine}{20}

  \SetKwProg{Func}{Function}{}{}

  \tcp*[h]{Implement particular coarse-resolution action at
    fine-resolution; return failure/false if action cannot be executed
    in current state, or if the fine-resolution execution terminates
    without implementing this action; otherwise, update
    coarse-resolution state $currState$ and history $\mathcal{H}$, and
    return success/true}

  \Func{Implement($currState, currAction, \mathcal{H}$) } {

    Controller extracts $T = \langle currState, currAction,
    \sigma'\rangle$ from answer set of $\Pi(\mathcal{D}_H, currAction,
    currState)$

    \BlankLine

    \tcp*[h]{Exit if the coarse-resolution action can no longer be
      executed}
    
    \uIf{no answer set} { 
      
      \Return failure
    } % check AS of transition
        
    \BlankLine
    
    \tcp*[h]{Zoom, construct and solve POMDP relevant to $T$}

    Controller zooms to $\mathcal{D}_{LR}(T)$, the part of
    $\mathcal{D}_{LR}$ relevant to transition $T$ and constructs a
    POMDP $\mathcal{P}_o(T)$.

    Controller solves $\mathcal{P}_o(T)$ to compute a policy.

    \BlankLine
    \tcp*[h]{Execute POMDP policy until termination}
    
    executed := false
      
    \While{$\lnot$ executed} {
        
      Statistician selects action using policy, executes action,
      obtains observation, and updates belief state
          
      \uIf{terminal action executed} {
        
        Statistician communicates observations to the controller
        
        executed := true
        
      }
    } % while (possible)
    
    \BlankLine
    
    \tcp*[h]{Fine-resolution inference, update $\mathcal{H}$ and
      $currState$}

    Controller performs fine-resolution inference, adds
    coarse-resolution outcomes and observations to $\mathcal{H}$
    
    $currState := $ current coarse-resolution state

    \uIf{$currState \not= \sigma'$} {
      \Return failure
    }
    \Return success
    
    %\Return done
  } % end of Implement(transition) function
\end{algorithm}

\medskip
\noindent
Algorithm~\ref{alg:control-loop} describes the reasoning system and
overall control loop of our architecture for building intelligent
robots. For this description, we (once again) view a robot as
consisting of a logician and a statistician, who communicate through a
controller, as described in Section~\ref{sec:intro} and shown in
Figure~\ref{fig:arch-overview}.  For any given goal $G$, the logician
takes as input the system description $\mathcal{D}_H$ that corresponds
to a coarse-resolution transition diagram $\tau_H$, recorded history
$\mathcal{H}$ with initial state defaults (see
Example~\ref{ex:defaults}), and the current coarse-resolution state
$\sigma$. Diagnostics and planning to achieve $G$ are reduced to
computing answer sets of the corresponding CR-Prolog program $Plan =
\Pi^n(\mathcal{D}_H, \mathcal{H}) \cup Classical\_plan \cup DPC$ (Line
4, also see Proposition~\ref{prop:plan-reduce}). If no such answer set
is found, the control loop terminates reporting failure (Lines 5-7).
If a plan exists, each coarse-resolution action $a_i^H, i\in[1, n]$ in
the plan is implemented one after the other until either one of the
actions can no longer be executed, or the entire sequence is
implemented successfully (Lines 11-20).

To implement a given coarse-resolution action $a_i^H$ in current state
$currState$, the controller first checks whether the corresponding
transition is feasible; if not, the implementation of this action (and
thus the entire coarse-resolution plan) is terminated early (Lines
23-24). If the transition is feasible, the controller zooms to the
relevant part of the randomized, fine-resolution system description
$\mathcal{D}_{LR}(T)$, constructs the corresponding POMDP
$\mathcal{P}_o(T)$, and solves it to obtain a policy (Lines 25-26).
The statistician repeatedly invokes this policy to select an action,
execute the action, obtain an observation, and update the belief
state, until a terminal action is executed (Lines 28-33). The action
outcomes are communicated to the controller, which performs
fine-resolution inference, updates coarse-resolution history
$\mathcal{H}$, and updates $currState$ to be the current
coarse-resolution state. Note that the fine-resolution implementation
of a coarse-resolution action succeeds iff the desired transition is
achieved (Lines 36-38).

\medskip
\noindent
Notice that executing Algorithm~\ref{alg:control-loop} involves:
\begin{enumerate}
\item Applying the planning and diagnostics algorithm discussed in
  Section~\ref{sec:arch-hl-reason} for planning with $\tau_H$ and
  $\mathcal{H}$;
\item For any given coarse-resolution transition $T$, automatically
  constructing $\mathcal{D}_{LR}(T)$ by zooming, as described in
  Section~\ref{sec:arch-refzoomrand}; and
\item Constructing a POMDP from $\mathcal{D}_{LR}(T)$, solving it, and
  using the corresponding policy to execute a sequence of
  fine-resolution actions implementing $T$ until termination, as
  discussed in Section~\ref{sec:arch-pomdp}.
\end{enumerate}
It is not difficult to show that the algorithm satisfies the
specifications. Consider the behavior of the algorithm when the
algorithm receives the appropriate input and there is a state
satisfying the assigned goal.  In this case, when the control loop is
completed, with high probability (or equivalently, with a low margin
of error) the robot will be in a state satisfying the goal.  Also, if
the goal cannot be achieved, the robot will (with high probability)
report failure in achieving this goal. The control loop thus results
in \emph{correct} behavior of the robot.

The execution of fine-resolution actions based on probabilistic models
of uncertainty in perception and actuation (e.g., Line 29,
Algorithm~\ref{alg:control-loop}) is supported by probabilistic state
estimation algorithms that process inputs from sensors and actuators.
For instance, the robot builds a map of the domain and estimates its
position in the map using a \emph{Particle Filter} algorithm for
Simultaneous Localization and Mapping (SLAM)~\cite{thrun:robotics05}.
This algorithm represents the true underlying probability distribution
over the possible states using samples drawn from a proposal
distribution. Samples more likely to represent the true state,
determined based on the degree of match between the expected and
actual sensor observations of domain landmarks, are assigned higher
(relative) weights and re-sampled to incrementally converge to the
true distribution. Implementations of the particle filtering algorithm
are used widely in the robotics literature to track multiple
hypotheses of system state. A similar algorithm is used to estimate
the pose of the robot's arm. On the physical robot, other algorithms
used to process specific sensor inputs. For instance, we use existing
implementations of algorithms to process camera images, which are the
primary source of information to identify specific domain objects. The
robot also uses an existing implementation of a SLAM algorithm to
build a domain map and localize itself in the map. These algorithms
are summarized in Section~\ref{sec:exp}, when we discuss experiments
on physical robots.

%%%%%-----------------------------------------------------------------------------
%%%%%-----------------------------------------------------------------------------
\section{Experimental Setup and Results}
\label{sec:exp}
This section describes the experimental setup and results of
evaluating the architecture's capabilities.

%%%%%-----------------------------------------------------------------------------
\subsection{Experimental setup}
\label{sec:exp-setup}
The proposed architecture was evaluated in simulation and on a
physical robot. As stated in Section~\ref{sec:arch-pomdp}, statistics
of action execution, e.g., observed outcomes of all actions and
computation time for knowledge producing actions, are collected in an
initial training phase. These statistics are used by the controller to
compute the relative utility of different actions, and the
probabilities of obtaining different action outcomes and observations.
The simulator uses these statistics to simulate the robot's movement
and perception. In addition, the simulator represents objects using
probabilistic functions of features extracted from images, with the
corresponding models being acquired in an initial training
phase---see~\cite{zhang:TRO13} for more details about such models.

In each experimental trial, the robot's goal was to find and move
specific objects to specific places---the robot's location, the target
object, and locations of domain objects were chosen randomly. An
action sequence extracted from an answer set of the ASP program
provides a plan comprising abstract actions, each of which is executed
probabilistically. Our refinement-based architecture ``REBA'' was
compared with: (1) POMDP-1, which constructs a POMDP from the
fine-resolution description (and computed statistics), computes the
policy, and uses this policy to implement the desired abstract action;
and (2) POMDP-2, which revises POMDP-1 by assigning specific
probability values to default statements to bias the initial belief.
The performance measures were: (a) \emph{success}, the fraction (or
$\%$) of trials in which the robot achieved the assigned goals; (b)
\emph{planning time}, the time taken to compute a plan to achieve the
assigned goal; and (c) the average \emph{number of actions} that were
executed to achieve the desired goal. We evaluate the following three
key hypotheses:
\begin{enumerate}
\item[\underline{\textbf{H1}}] REBA simplifies design in comparison
  with architectures based on purely probabilistic reasoning and
  increases confidence in the correctness of the robot's behavior;
\item[\underline{\textbf{H2}}] REBA achieves the assigned goals more reliably and
  efficiently than POMDP-1; and 
\item[\underline{\textbf{H3}}] Our representation for defaults
  improves reliability and efficiency in comparison with not using
  defaults or assigning specific probability values to defaults.
\end{enumerate}
We examine the first hypothesis qualitatively in the context of some
execution traces grounded in the illustrative domain described in
Example~\ref{ex:illus-example} (Section~\ref{sec:exp-traces}). We then
discuss the quantitative results corresponding to the experimental
evaluation of the other two hypotheses in simulation and on physical
robots (Section~\ref{sec:exp-results}).

%%%%%-----------------------------------------------------------------------------
\subsection{Execution traces}
\label{sec:exp-traces}
The following (example) execution traces illustrate some of the key
capabilities of the proposed architecture. 

\begin{execexample}\label{exec:defaults}[Planning with default knowledge]\\
  {\rm Consider the scenario in which a robot is assisting with a
    meeting in the $of\!\!fice$, i.e., $loc(rob_1, of\!\!fice)$, and
    is assigned a goal state that contains:
    $$loc(cup_1, of\!\!fice)$$
    where the robot's goal is to move coffee cup $cup_1$ to the
    $of\!\!fice$.
    \begin{itemize}
    \item The plan of abstract actions, as created by the logician,
      is:
      \begin{align*}
        &move(rob_1, kitchen), ~~grasp(rob_1, cup_1)\\
        &move(rob_1, of\!\!fice), ~~putdown(rob_1, cup_1)
      \end{align*}
      \noindent
      Note that this plan uses initial state default knowledge that
      $kitchenware$ are usually found in the $kitchen$. Each abstract
      action in this plan is executed by computing and executing a
      sequence of concrete actions.

    \item To implement $move(rob_1, kitchen)$, the controller
      constructs $\mathcal{D}_{LR}(T)$ by zooming to the part of
      $\mathcal{D}_{LR}$ relevant to this action. For instance, only
      cells in the $kitchen$ and the $of\!\!fice$ are possible
      locations of $rob_1$, and $move$ is the only action that can
      change the physical state, in the fine-resolution
      representation.

    \item $\mathcal{D}_{LR}(T)$ is used to construct and solve a POMDP
      to obtain an action selection policy, which is provided to the
      statistician. The statistician repeatedly invokes this policy to
      select actions (until a terminal action is selected) that are
      executed by the robot. In the context of
      Figure~\ref{fig:refine-gridmap}, assume that the robot moved
      from cell $c_1\in of\!\!fice$ to $c_5\in kitchen$ (through cell
      $c_2\in of\!\!fice$) with high probability.

    \item The direct observation from the POMDP,
      $observed_{loc^*}(rob_1, rob_1, c_5)=true$, is used by the
      controller for inference in $\mathcal{D}_{LR}(T)$ and
      $\mathcal{D}_L$, e.g., to produce $observed_{loc}(rob_1, rob_1,
      kitchen)$. The controller adds this information to the
      coarse-resolution history $\mathcal{H}$ of the logician, e.g.,
      $obs(rob_1, loc(rob_1)=kitchen, 1)$. Since the first abstract
      action has had the expected outcome, the logician sends the next
      abstract action in the plan, $grasp(rob_1, cup_1)$ to the
      controller for implementation.
   
    \item A similar sequence of steps is performed for each abstract
      action in the plan, e.g., to grasp $cup_1$, the robot locates
      the coffee cup in the $kitchen$ and then picks it up. Subsequent
      actions cause $rob_1$ to move $cup_1$ to the $of\!\!fice$, and
      put $cup_1$ down to achieve the assigned goal.
    \end{itemize}
  }
\end{execexample}

\begin{execexample}\label{exec:planfail}[Planning with unexpected failure]\\
  {\rm Consider the scenario in which a robot in the $of\!\!fice$ is
    assigned the goal of fetching textbook $tb_1$, i.e., the initial
    state includes $loc(rob_1, of\!\!fice)$, and the goal state
    includes:
    $$loc(tb_1, of\!\!fice)$$
    The coarse-resolution system description $\mathcal{D}_H$ and
    history $\mathcal{H}$, along with the goal, are passed on to the
    logician.
    \begin{itemize}
    \item The plan of abstract actions, as created by the logician,
      is:
      \begin{align*}
        &move(rob_1, main\_library), ~~grasp(rob_1, tb_1)\\
        &move(rob_1, of\!\!fice), ~~putdown(rob_1, tb_1)
      \end{align*}
      This plan uses the default knowledge that textbooks are
      typically in the $main\_library$ (Statement~\ref{def:d1}).  Each
      abstract action in this plan is executed by computing and
      executing a sequence of concrete actions.

    \item Assume that $loc(rob_1, main\_library)$, i.e., that the
      robot is in the $main\_library$ after successfully executing the
      first abstract action. To execute the $grasp(rob_1, tb_1)$
      action, the controller constructs $\mathcal{D}_{LR}(T)$ by
      zooming to the part of $\mathcal{D}_{LR}$ relevant to this
      action. For instance, only cells in the $main\_library$ are
      possible locations of $rob_1$ and $tb_1$ in the fine-resolution
      representation.

    \item $\mathcal{D}_{LR}(T)$ is used to construct and solve a POMDP
      to obtain an action selection policy, which is provided to the
      statistician. The statistician repeatedly invokes this policy to
      select actions (until a terminal action is selected) that are
      executed by the robot. In the context of
      Figure~\ref{fig:refine-gridmap}, if $r_2$ is the
      $main\_library$, the robot may move to and search for $tb_1$ in
      each cell in $r_2$, starting from its current location.

    \item The robot unfortunately does not find $tb_1$ in any cells of
      the $main\_library$ in the second step. These observations from
      the POMDP, i.e., $observed_{loc^*}(rob_1, tb_1, c_i) = false$
      for each $c_i\in main\_library$, are used by the controller for
      inference in $\mathcal{D}_{LR}(T)$ and $\mathcal{D}_L$. This
      inference produces $observed_{loc}(rob_1, tb_1, main\_library) =
      false$ and other observations, which (in turn) results in
      suitable statements being added by the controller to the
      coarse-resolution history $\mathcal{H}$, e.g., $obs(rob_1,
      loc(tb_1)\neq main\_library, 2)$.

    \item The inconsistency caused by the observation is resolved by
      the logician using a CR rule, and the new plan is created based
      on the second initial state default that a textbook not in the
      $main\_library$ is typically in the $aux\_library$
      (Statement~\ref{def:d2}):
      \begin{align*}
        &move(rob_1, aux\_library), ~~grasp(rob_1, tb_1)\\
        &move(rob_1, of\!\!fice), ~~putdown(rob_1, tb_1)
      \end{align*}

    \item This time, the robot is able to successfully execute each
      abstract action in the plan, i.e., it is able to move to the
      $aux\_library$, find $tb_1$ and grasp it, move back to the
      $of\!\!fice$, and put $tb_1$ down to achieve the assigned goal.
      
    \end{itemize}
  }
\end{execexample}
\noindent
Both these examples illustrate key advantages provided by the formal
definitions, e.g., of the different system descriptions and the tight
coupling between them, which are part of our architecture:
\begin{enumerate}
\item Once the designer has provided the domain-specific information,
  e.g., for refinement or computing probabilities of action outcomes,
  planning, diagnostics, and execution for any given goal can be
  automated.

\item Attention is automatically directed to the relevant knowledge at
  the appropriate resolution. For instance, reasoning by the logician
  (statistician) is restricted to a coarse-resolution (zoomed
  fine-resolution) system description.  It is thus easier to
  understand, and to fix errors in, the observed behavior, in
  comparison with architectures that consider all the available
  knowledge or only support probabilistic reasoning.

\item There is smooth transfer of control and relevant knowledge
  between components of the architecture, and confidence in the
  correctness of the robot's behavior. Also, the proposed methodology
  supports the use of this architecture on different robots in
  different domains, e.g., Section~\ref{sec:exp-results} describes the
  use of this architecture on robots in two different indoor domains.
\end{enumerate}

\noindent
Next, we describe the experimental evaluation of the hypotheses H2 and
H3 in simulation and on a mobile robot.

%%%%%-----------------------------------------------------------------------------
\subsection{Experimental results}
\label{sec:exp-results}
\begin{figure}[tb]
  \begin{center}%\hspace*{-1em}
    \includegraphics[width=0.85\columnwidth]{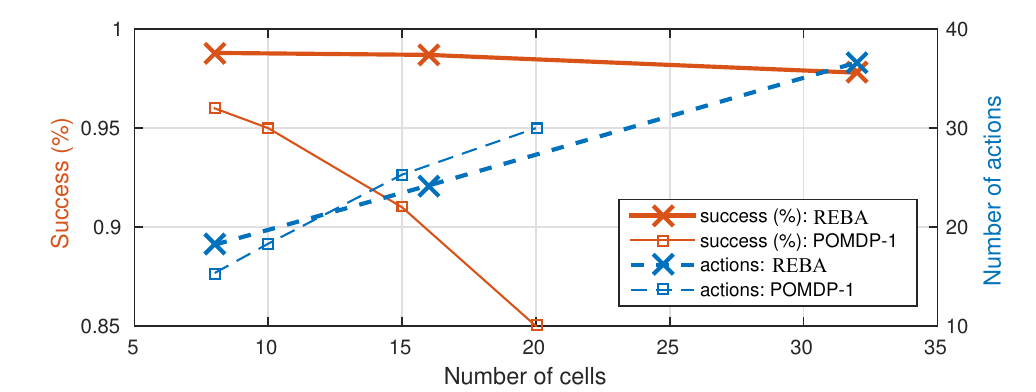}
  \end{center}
  \vspace{-1.5em}
  \caption{Ability to successfully achieve the assigned goal, and the
    number of actions executed before termination, as a function of
    the number of cells in the domain. REBA significantly increases
    accuracy and reduces the number of actions executed, in comparison
    with POMDP-1, as the number of cells in the domain increases.}
  \label{fig:accuracy}
\end{figure}

To evaluate hypothesis H2, we first compared REBA with POMDP-1 in a set
of trials in which the robot's initial position is known but the
position of the object to be moved is unknown. The solver used in
POMDP-1 was evaluated with different fixed amounts of time for
computing action policies.  Figure~\ref{fig:accuracy} summarizes the
results; each point is the average of $1000$ trials, and we set (for
ease of interpretation) each room to have four cells. The
brown-colored plots in Figure~\ref{fig:accuracy} represent the ability
to successfully achieve the assigned goal (y-axis on the left), as a
function of the number of cells in the domain. The blue-colored plots
show the number of actions executed before termination. For the plots
corresponding to POMDP-1, the number of actions the robot is allowed
to execute before it has to terminate is set to $50$.  We note that REBA
significantly improves the robot's ability to achieve the assigned
goal in comparison with POMDP-1. As the number of cells (i.e., size of
the domain) increases, it becomes computationally difficult to
generate good policies with POMDP-1. The robot needs a greater number
of actions to achieve the goal and there is a loss in accuracy if the
limit on the number of actions the robot can execute before
termination is reduced. While using POMDP-1, any incorrect
observations (e.g., incorrect sightings of objects) significantly
impacts the ability to complete the trials.  REBA, on the other hand,
directs the robot's attention to relevant regions of the domain (e.g.,
specific rooms), and it is thus able to recover from errors and
operate efficiently.

\begin{figure*}[tb]
  \begin{center}
    \subfigure[0.33\textwidth][Using all knowledge] {
      \includegraphics[width=0.33\textwidth]{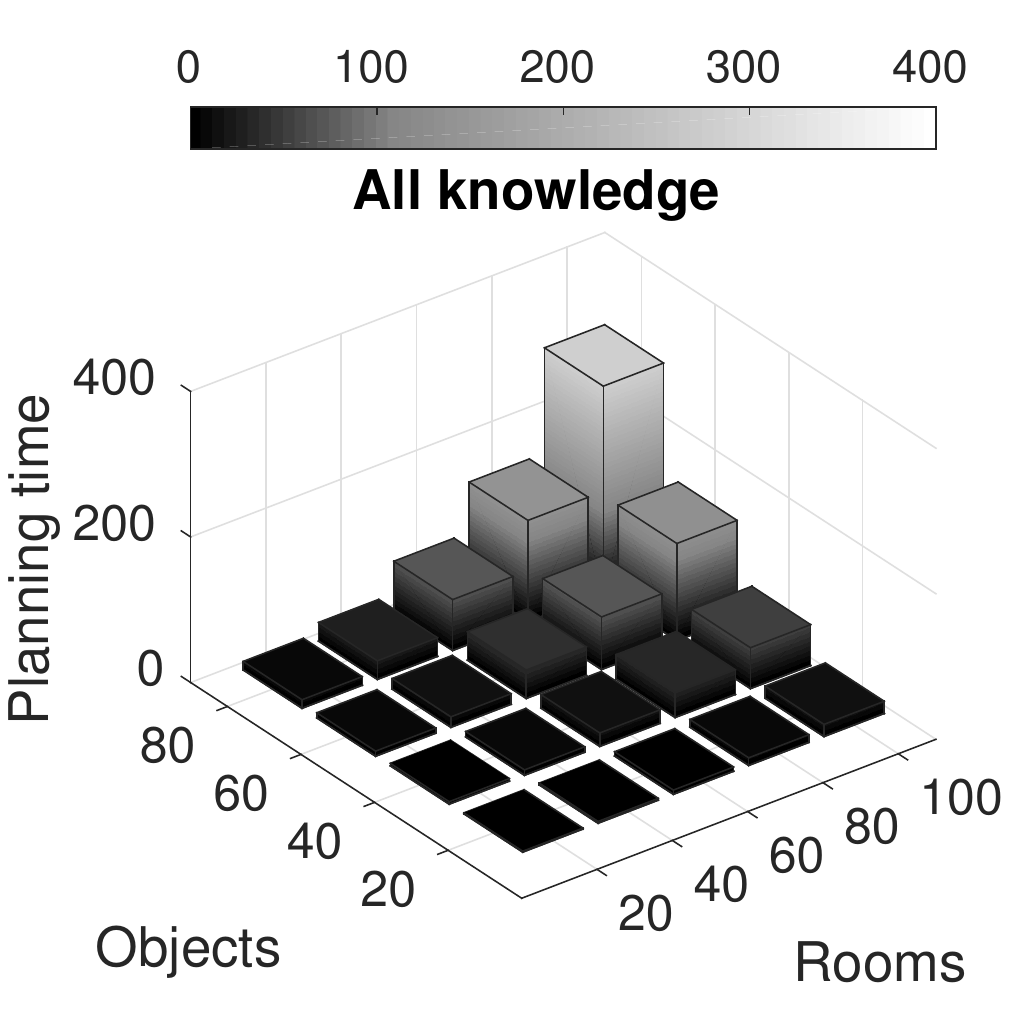}
      \label{fig:room-object-all}
    }%\hspace{-0.1in}
    \subfigure[0.33\textwidth][Using relevant knowledge] {
      \includegraphics[width=0.33\textwidth]{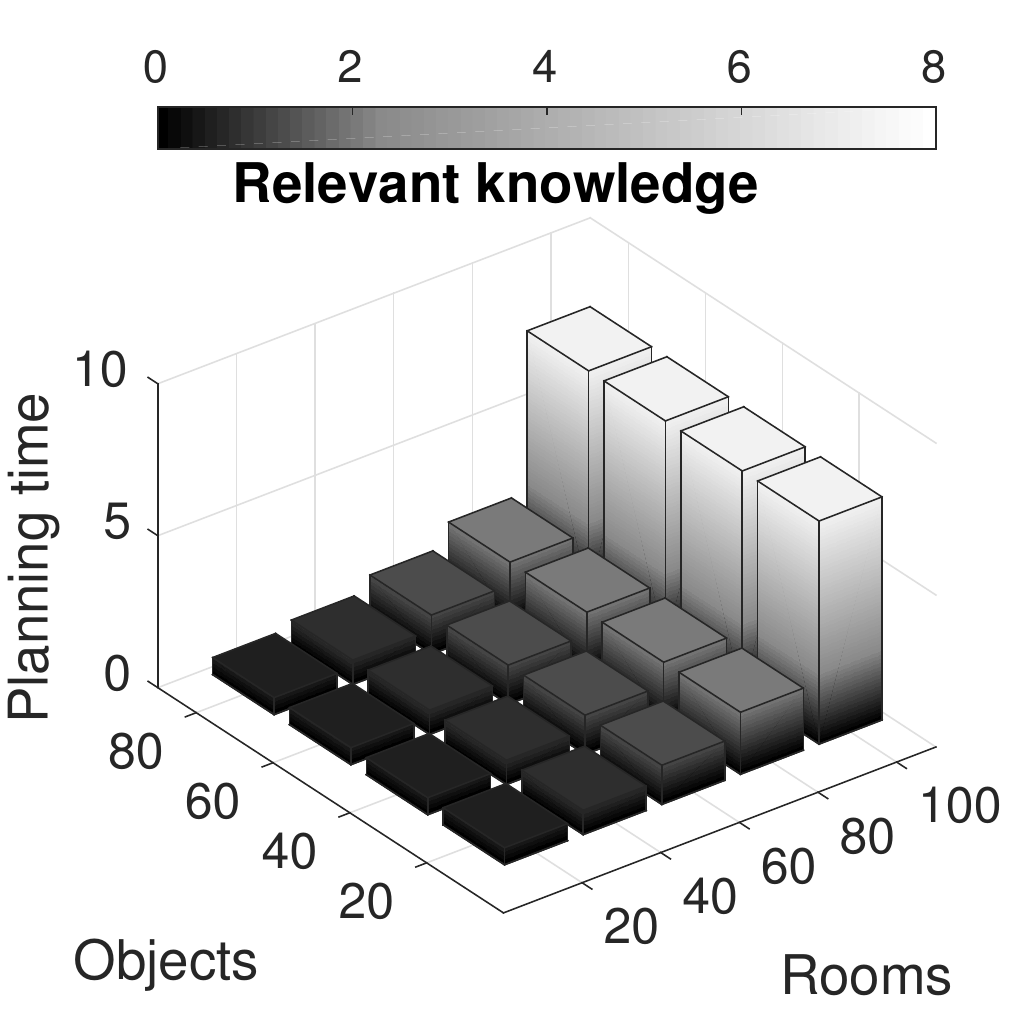}
      \label{fig:room-object-rel}
    }%\hspace{-0.1in}
    \subfigure[0.33\textwidth][Using some knowledge] {
      \includegraphics[width=0.33\textwidth]{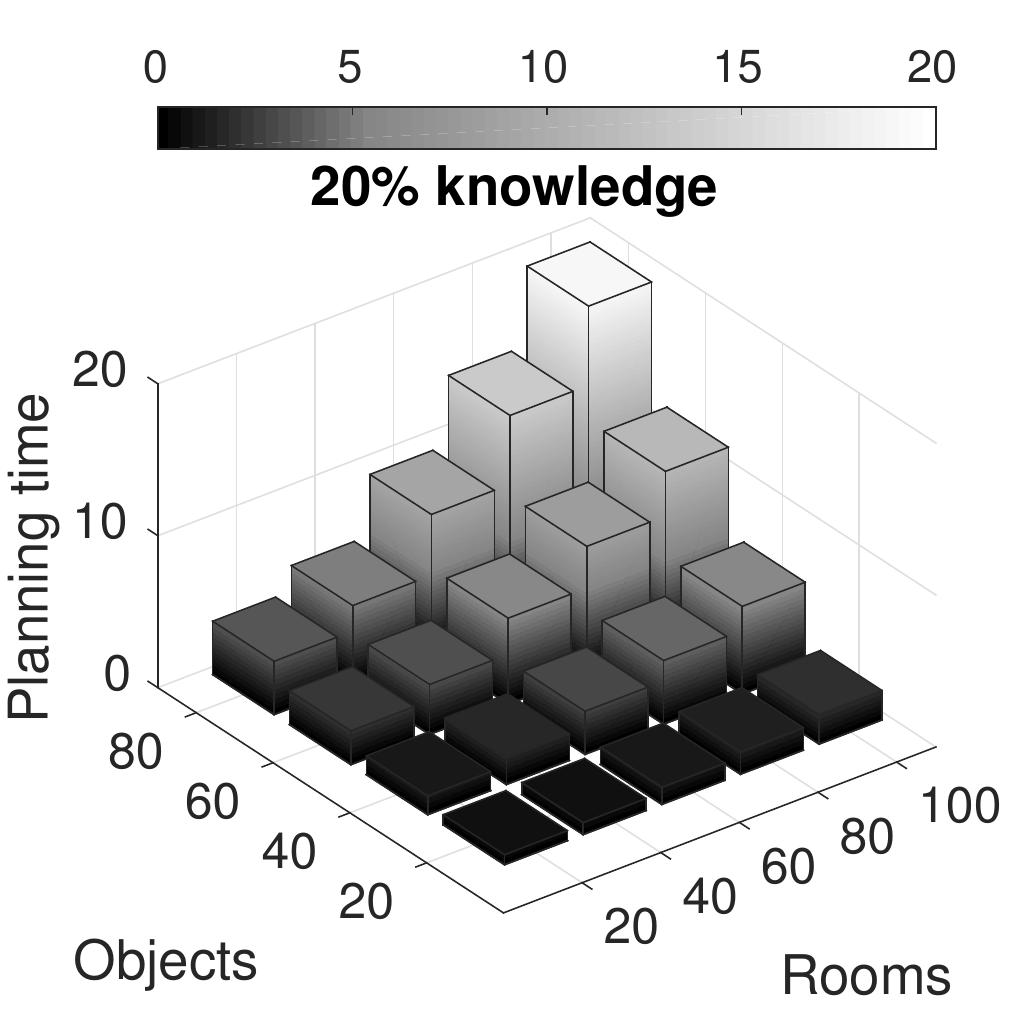}
      \label{fig:room-object-some}
    }
  \end{center}
  \vspace{-2em}
  \caption{Planning time as a function of the number of rooms and the
    number of objects in the domain---REBA only uses relevant
    knowledge for reasoning, and is thus able to scale to larger
    number of rooms and objects.}
  \label{fig:room-object}
\end{figure*}

% \begin{figure}[tbch]
%   \begin{center}%\hspace*{-1.7em}
%     \includegraphics[width=0.75\columnwidth]{rm_obj}
%   \end{center}
%   \vspace{-2em}
%   \caption{Planning time as a function of the number of rooms and
%     the number of objects in the domain---REBA scales to larger
%     number of rooms and objects.}
%   \label{fig:room-object}
% \end{figure}

Next, we evaluated the time taken by REBA to generate a plan as the size
of the domain increases. We characterize domain size based on the
number of rooms and the number of objects in the domain. We conducted
three sets of experiments in which the robot reasons with: (1) all
available knowledge of domain objects and rooms; (2) only knowledge
relevant to the assigned goal---e.g., if the robot knows an object's
default location, it need not reason about other objects and rooms in
the domain to locate this object; and (3) relevant knowledge and
knowledge of an additional $20\%$ of randomly selected domain objects
and rooms.
Figures~\ref{fig:room-object-all}-\ref{fig:room-object-some} summarize
these results. We observe that using just the knowledge relevant to
the goal to be accomplished significantly reduces the planning time.
REBA supports the identification of such knowledge based on the
refinement and zooming operations described in
Section~\ref{sec:arch-refzoomrand}.  As a result, robots equipped with
REBA will be able to generate appropriate plans for domains with a large
number of rooms and objects. Furthermore, if we only use a
probabilistic approach (POMDP-1), it soon becomes computationally
intractable to generate a plan for domains with many objects and
rooms. These results are not shown in Figure~\ref{fig:room-object},
but they are documented in prior papers evaluating just the
probabilistic component of the proposed
architecture~\cite{Sridharan:AIJ10,zhang:TRO13}.

\begin{figure}[tb]
  \begin{center}%\hspace*{-1.5em}
    \includegraphics[width=0.61\columnwidth]{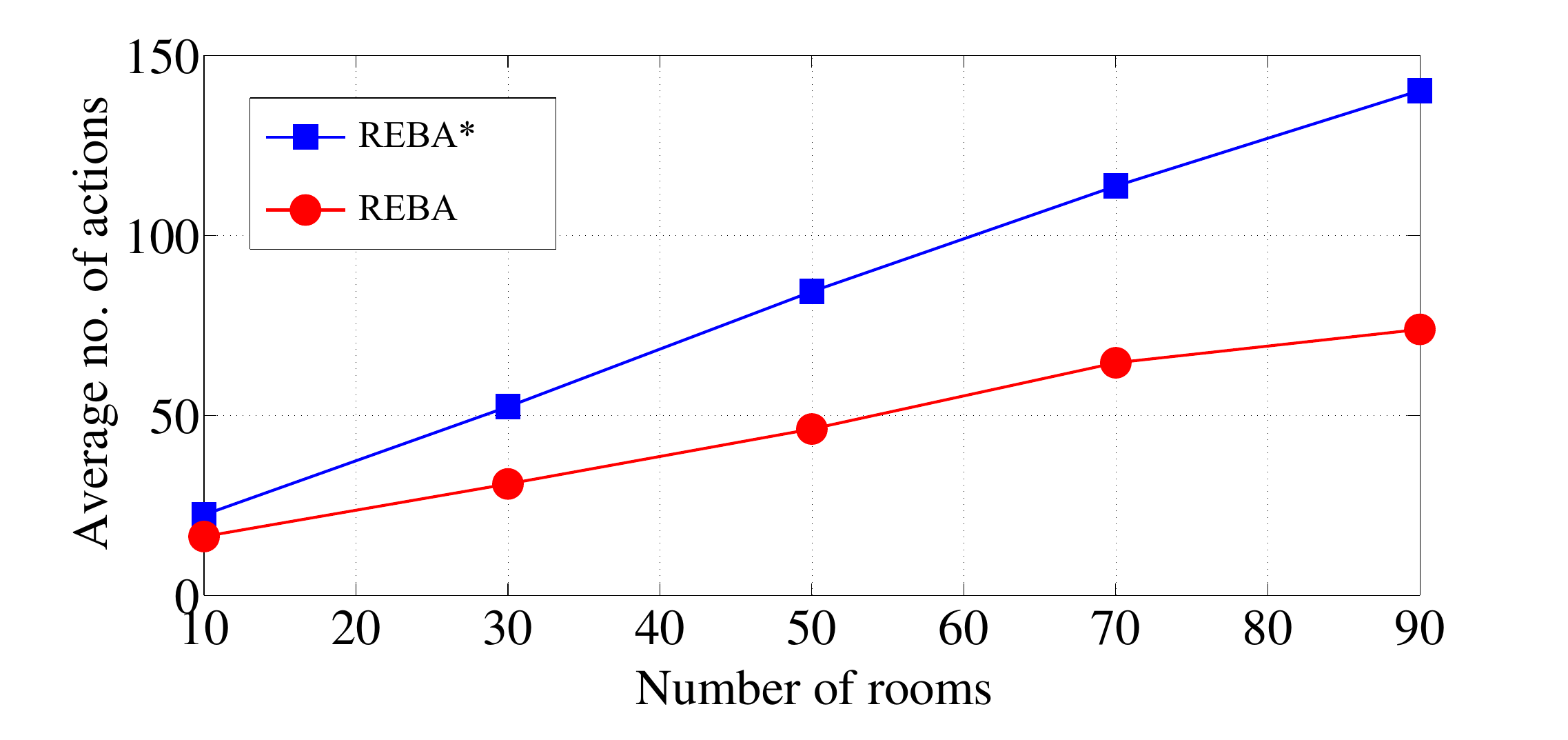}
  \end{center}
  \vspace{-2em}
  \caption{Effect of using default knowledge---principled
    representation of defaults significantly reduces the number of
    actions (and thus time) for achieving assigned goal.}
  \label{fig:default}
\end{figure}

\begin{figure}[tb]
  \begin{center}%\hspace*{-1.5em}
    \includegraphics[width=0.75\columnwidth]{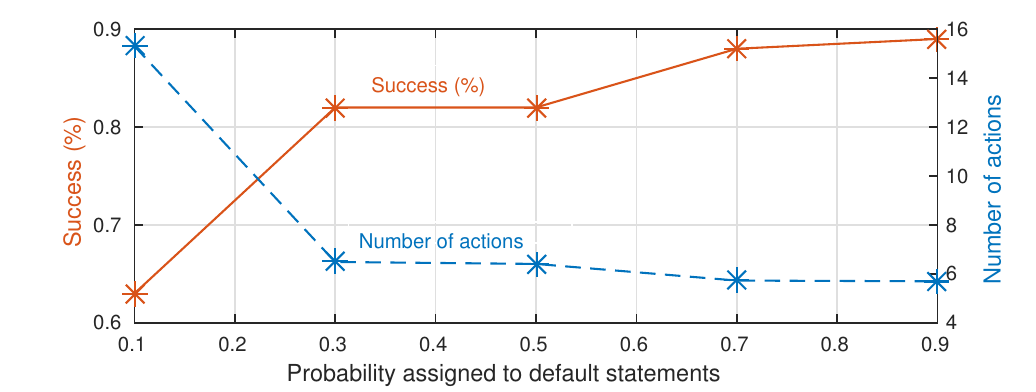}
  \end{center}
  \vspace{-2em}
  \caption{Ability to achieve goals, and number of actions executed,
    using only POMDPs, when different probability values are assigned
    to default statements and the ground truth locations of objects
    perfectly matches the default locations. The number of actions
    decreases and success (\%) increases as the probability value
    increases.  }
  \label{fig:default-match}
\end{figure}

\begin{figure}[tb]
  \begin{center}%\hspace*{-1.5em}
    \includegraphics[width=0.75\columnwidth]{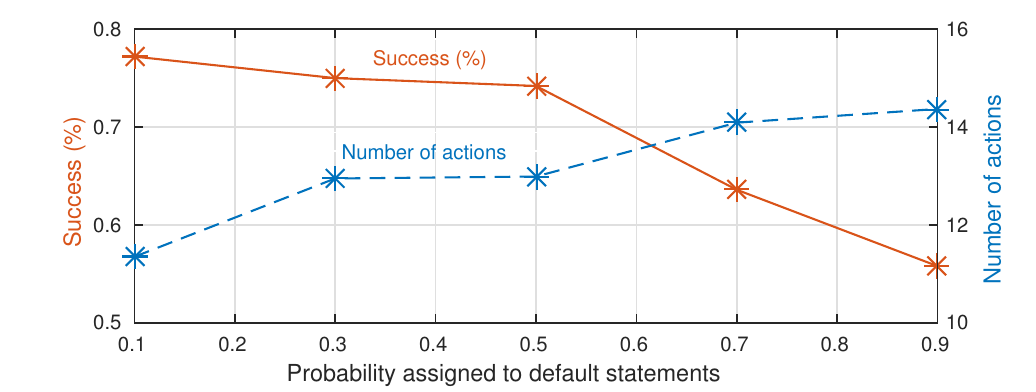}
  \end{center}
  \vspace{-2em}
  \caption{Ability to achieve goals, and number of actions executed,
    using only POMDPs, when different probability values are assigned
    to default statements and the ground truth locations of objects
    never matches the default locations. The number of actions
    increases and success (\%) decreases as the probability value
    increases.}
  \label{fig:default-mismatch}
\end{figure}

To evaluate hypothesis H3, i.e., to evaluate the effect of our
representation and use of default knowledge on reliability and
computational efficiency of decision making, we first conducted trials
in which REBA was compared with $\mathcal{REBA}^*$, a version that
does not include any default knowledge, e.g., when the robot is asked
to fetch a textbook, there is no prior knowledge regarding the
location of textbooks, and the robot explores the closest location
first.  Figure~\ref{fig:default} summarizes the average number of
actions executed per trial as a function of the number of rooms in the
domain---each sample point in this figure is the average of $10000$
trials. The goal in each trial is (as before) to move a specific
object to a specific place. We observe that our (proposed)
representation and use of default knowledge significantly reduces the
number of actions (and thus time) required to achieve the assigned
goal.

Next REBA was compared with POMDP-2, a version of POMDP-1 that assigns
specific probability values to default knowledge (e.g., ``textbooks
are in the library with probability $0.9$'') and suitably revises the
initial belief state. The goal (once again) was to find and move
objects to specific locations, and we measured the ability to
successfully achieve the assigned goal and the number of actions
executed before termination.
Figures~\ref{fig:default-match}-\ref{fig:default-mismatch} summarize
the corresponding results under two extreme cases representing a
perfect match (mismatch) between the default locations and ground
truth locations of objects. In Figure~\ref{fig:default-match}, the
ground truth locations of target objects (unknown to the robot) match
the default locations of the objects, i.e., there are no exceptions to
the default statements. We observe that as the probability assigned to
the default statement increases, the number of actions executed by the
robot decreases and the fraction of trials completed successfully
increases. However, for larger values along the x-axis, the difference
in the robot's performance for two different values of the probability
(assigned to defaults) is not that significant. In
Figure~\ref{fig:default-match}, the ground truth locations of the
target objects never match the default locations of the objects, i.e.,
unknown to the robot, all trials correspond to exceptions to the
default knowledge. In this case, the robot executes many more actions
before termination and succeeds in a smaller fraction of trials as the
probability value assigned to default statements increases. We also
repeated these experimental trials after varying the extent to which
the ground truth locations of objects matched their default locations.
We noticed that when the probability assigned to default statements
accurately reflects the ground truth, the number of trials in which
the robot successfully achieves the goal increases and approaches the
performance obtained with REBA. However, recall that computing the
probabilities of default statements accurately takes a lot of time and
effort. Also, these probabilities may change over time and the robot's
ability to achieve the assigned goals may be sensitive to these
changes, making it difficult to predict the robot's behavior with
confidence. In addition, it is all the more challenging to accurately
represent and efficiently use probabilistic information about
prioritized defaults (e.g., Example~\ref{ex:defaults}). In general, we
observed that the effect of assigning a probability value to defaults
is arbitrary depending on factors such as (a) the numerical value
chosen; and (b) the degree of match between ground truth and the
default information.  For instance, \emph{if a large probability is
  assigned to the default that books are typically in the library, but
  the book the robot has to move is an exception to the default (e.g.,
  a cookbook), it takes significantly longer for POMDP-2 to recover
  from the initial belief}.  REBA, on the other hand, supports elegant
representation of, and reasoning with, defaults and exceptions to
these defaults.

% 1. How easy is it to get probabilities for the defaults?
% 2. Given good probabilities what can we say about the efficiency?
% 3. What happens with the performance if the probability undergoes
% slight changes?
% 4. Use of defaults allows us to predict robot's behavior (not found in
% A it'll go to B). Can we do this without defaults?

% \begin{figure}[tbc]
%   \begin{center}
%     \includegraphics[width=0.7\textwidth]{map}
%   \end{center}
%   \vspace{-2em}
%   \caption{Subset of the map of the second floor of our department;
%     specific places are labeled as shown, and used during planning to
%     achieve the assigned goals.}
%   \label{fig:map}
% \end{figure}

\paragraph{\underline{Robot Experiments:}}
In addition to the trials in simulated domains, we implemented and
evaluated REBA with POMDP-1 on physical robots using the Robot Operating
System (ROS). We conducted experimental trials with two robot
platforms (see Figure~\ref{fig:map-robot}) in variants of the domain
described in Example~\ref{ex:illus-example}. Visual object recognition
is based on learned object models that consist of appearance-based and
contextual visual cues~\cite{li:icar13}. Since, in each trial, the
robot's initial location and the target object(s) are chosen randomly,
it is difficult to compute a meaningful estimate of variance, and
statistical significance is established through paired trials. In each
paired trial, for each approach being compared (e.g., REBA or POMDP-1),
the target object(s), the robot's initial location, and the location
of domain objects are the same, and the robot has the same initial
domain knowledge.

First, we conducted $50$ trials on two floors of our Computer Science
department building. This domain includes places in addition to those
included in our illustrative example, e.g., Figure~\ref{fig:map} shows
a subset of the domain map of the third floor of the building, and
Figure~\ref{fig:robot1} shows the \emph{Peoplebot} wheeled robot
platform used in these trials.  The robot is equipped with a stereo
camera, laser range finder, microphone, speaker, and a laptop running
Ubuntu Linux that performs all the processing.  The domain maps are
learned and revised by the robot using laser range finder data and the
existing ROS implementation of a SLAM
algorithm~\cite{dissanayake:TRA01}. This robot has a manipulator arm
that can be moved to reachable 3D locations relative to the robot.
However, since robot manipulation is not a focus of this work, once
the robot is next to the desired object, it extends its gripper and
asks for the object to be placed in it. For experimental trials on the
third floor, we considered $15$ rooms, which includes faculty offices,
research labs, common areas and a corridor. To make it feasible to use
POMDP-1 in such large domains, we used our prior work on a
hierarchical decomposition of POMDPs for visual sensing and
information processing that supports automatic belief propagation
across the levels of the hierarchy and model generation in each level
of the hierarchy~\cite{Sridharan:AIJ10,zhang:TRO13}. The experiments
included paired trials, e.g., over $15$ trials (each), POMDP-1 takes
$1.64$ as much time as REBA (on average) to move specific objects to
specific places. For these paired trials, this $39\%$ reduction in
execution time provided by REBA is statistically significant:
\emph{p-value} $= 0.0023$ at the $95\%$ significance level.

Consider a trial in which the robot's objective is to bring a specific
textbook to the place named \emph{study\_corner}. The robot uses
default knowledge to create a plan of abstract actions that causes the
robot to move to and search for the textbook in the
\emph{main\_library}. When the robot does not find this textbook in
the \emph{main\_library} after searching using a suitable POMDP
policy, replanning by the logician causes the robot to investigate the
\emph{aux\_library}. The robot finds the desired textbook in the
\emph{aux\_library} and moves it to the target location. A video of
such an experimental trial can be viewed online at
\url{http://youtu.be/8zL4R8te6wg}

To explore the applicability of REBA in different domains, we also
conducted $40$ experimental trials using the \emph{Turtlebot} wheeled
robot platform in Figure~\ref{fig:robot2} in a variant of the
illustrative domain in Example~\ref{ex:illus-example}. This domain had
three rooms in the Electrical Engineering department building arranged
to mimic a robot operating as a robot butler, with additional objects
(e.g., tables, chairs, food items etc).  The robot was equipped with a
Kinect (RGB-D) sensor, a laser range finder, and a laptop running
Ubuntu Linux that performs all the processing. As before, the robot
used the ROS implementation of a SLAM algorithm, and a hierarchical
decomposition of POMDPs for POMDP-1.  This robot did not have a
manipulator arm---once it reached a location next to the location of
the desired object, it asks for the object to be placed on it. The
experiments included paired trials, e.g., in $15$ paired trials,
POMDP-1 takes $2.3$ as much time as REBA (on average) to move specific
objects to specific places---this reduction in execution time by REBA
is statistically significant at the $95\%$ significance level.

Consider a trial in which the robot's goal was to fetch a bag of
crisps for a human. The robot uses default knowledge about the
location of the bag of crisps (e.g., that they are usually in the
$kitchen$), to create a plan of abstract actions to achieve this goal.
Execution of this plan causes the robot to first move to the $kitchen$
and search for the bag of crisps. The robot finds the bag of crisps,
asks for the bag to be placed on it (since it has no manipulator), and
moves back to $table1$ in $lab1$ (the location of the human who wanted
the crisps), only to be told that it has brought a bag of chocolates
instead. The robot diagnoses the cause for this unexpected observation
as human error (i.e., that it was given incorrect bag in the $kitchen$
by a human). The robot then computes and executes a plan that has it
go back and fetch the correct bag (of crisps) this time. A video of
this trial can be viewed online at \url{https://vimeo.com/136990534}

% \paragraph{\underline{Threats to Validity:}}

%%%%%-----------------------------------------------------------------------------
%%%%%-----------------------------------------------------------------------------
%\vspace{-1em}
\section{Conclusions and Future Work}
%\vspace{-0.75em}
\label{sec:conclusion}
This paper described a refinement-based knowledge representation and
reasoning architecture (REBA) that combines the complementary
strengths of declarative programming and probabilistic graphical
models. The architecture is based on tightly-coupled transition
diagrams that represent domain knowledge, and the robot's abilities
and beliefs, at two levels of granularity. The architecture makes
the following key contributions:
\begin{itemize}
\item Action language $\mathcal{AL}_d$ is extended to support
  non-Boolean fluents and non-deterministic causal laws, and is used
  to describe the coarse-resolution and fine-resolution transition
  diagrams.

\item The notion of a history of a dynamic domain is extended to
  include default knowledge in the initial state, and a model of this
  history is defined. These definitions are used to define a notion of
  explanation of unexpected observations, and to provide an algorithm
  for coarse-resolution planning and diagnostics that translates
  history into a program of CR-Prolog, computes answer set of this
  program, and extracts plan and explanation (if needed) from the
  answer set.

\item A formal definition is provided of one transition diagram being
  a weak refinement of another transition diagram, and a
  fine-resolution diagram is defined as a weak refinement of the
  domain's coarse-resolution transition diagram.

\item A theory of observations is introduced and a formal definition
  is provided of one transition diagram being a strong refinement of
  another transition diagram. This theory of observations is combined
  with the weakly refined fine-resolution transition diagram to obtain
  a fine-resolution transition diagram that is a strong refinement of
  the coarse-resolution transition diagram.

\item The randomization of the fine-resolution transition diagram is
  defined, and an approach is described for experimental collection of
  statistics. These statistics are used to compute the probabilities
  of action outcomes and observations at the fine-resolution.

\item A formal definition is provided for zooming to a part of the
  randomized fine-resolution diagram relevant to any given
  coarse-resolution (abstract) transition. This definition is used to
  automate the zoom operation and implement each abstract action in
  the coarse-resolution plan.

\item An algorithm is provided for automatically constructing data
  structures appropriate for the fine-resolution probabilistic
  implementation of any given abstract action. This algorithm uses
  probabilistic models of the uncertainty in sensing and actuation,
  and the zoomed part of the fine-resolution transition diagram. The
  outcomes of the fine-resolution execution update the
  coarse-resolution history for subsequent reasoning.

\item Finally, and possibly one of the major contributions, is a
  general methodology for the design of software components of robots
  that are re-taskable and robust. This design methodology is based on
  Dijkstra's view of step-wise refinement of the specification of a
  program.
  % It simplifies the use of our architecture in other domains,
  % provides a path to predict the robot's behavior, and thus
  % increases confidence in the correctness of the robot's behavior.
\end{itemize}
In this paper, the domain representation for coarse-resolution
non-monotonic logical reasoning is translated to a CR-Prolog program,
and the representation for probabilistic reasoning is translated to a
POMDP. The key advantages of using this architecture are:
\begin{itemize}
\item It substantially simplifies the design process and increases
  confidence in the correctness of the robot's behavior. In
  particular:
  \begin{itemize}
  \item Step-wise refinement leads to clear separation of concerns and
    supports early testing of the different components of the
    architecture.

  \item The formal (i.e., mathematical) descriptions of the different
    components, and of the flow of control and information between the
    components, helps characterize the robot's behavior accurately and
    prove correctness of the algorithms.

  \item The domain-independent representations of part of the
    commonsense knowledge used by the robot, e.g., theory of
    observations, weak refinement and strong refinement, allows for
    the reuse of these representations on other robots and application
    domains.

  \item There is a single framework for inference, planning,
    diagnostics, and for a quantifiable trade off between accuracy and
    computational efficiency in the presence of probabilistic models
    of uncertainty in sensing and actuation.
  \end{itemize}
  
\item It significantly improves the computational efficiency and
  reliability of the robot's actions. In particular:
  \begin{itemize}
  \item The robot is able to reason reliably and efficiently with
    hierarchically-organized knowledge and beliefs. 

  \item Experimental results in simulation and on physical robots in
    different domains indicate the ability to reason at the
    sensorimotor level and the cognitive level with violation of
    defaults, noisy observations and unreliable actions.

  \item The tight coupling between representation and reasoning at
    different resolutions, based on concepts such as refinement and
    zooming, supports precise reasoning while still demonstrating the
    potential to scale to complex domains.
  \end{itemize}
\end{itemize}

% \commentm{Discuss benefits of the architecture, e.g., simplicity of
%   design, from slides. Also say that the methodology of design
%   advocated here is based on Dijkstra's view of step-wise refinement of
%   specification of a program}

\noindent
The proposed architecture opens up many directions for further
research, some of which relax the constraints imposed in the design of
our current architecture. First, we will further explore the tight
coupling between the transition diagrams, and between logical and
probabilistic reasoning, in dynamic domains. We have, for instance,
explored different resolutions for reasoning
probabilistically~\cite{colaco:acra15}, and investigated the
inference, planning and diagnostics capabilities of architectures that
reason at different resolutions~\cite{zhang:TRO15}.  However, we have
so far not explored non-stationary domains, a limiting constraint that
we seek to relax in future work. Second, our architecture has so far
focused on a single robot, although we have instantiated the
architecture in different domains. Another direction of further
research is to extend the architecture to enable collaboration between
a team of robots working towards a shared goal.  It is theoretically
possible to use our current architecture on multiple robots, but it
will open up challenging questions and choices regarding communication
(between robots) and propagation of beliefs held by different members
of the team. Third, the proposed architecture has focused on
representation and reasoning with incomplete knowledge, but a robot
collaborating with humans in a dynamic domain also needs to be able to
revise and augment its existing knowledge.  Preliminary work in this
direction, e.g., based on exploiting the complementary strengths of
relational reinforcement learning, active (inductive) learning, and
reasoning with commonsense knowledge, has provided promising
results~\cite{mohan:acs17,mohan:icaps17,mohan:acs18}, and we seek to
further explore this direction of work in the future. The long-term
objective is to better understand the coupling between non-monotonic
logical reasoning and probabilistic reasoning, and to use this
understanding to develop architectures that enable robots to assist
humans in complex domains.

%%%%%-----------------------------------------------------------------------------
%%%%%-----------------------------------------------------------------------------
%\vspace{-0.5em}
\section*{Acknowledgments}
% The authors wish to thank Pat Langley and Simon Parsons for their
% feedback regarding early versions of the architecture described in
% this paper. 
This work was supported in part by the U.S. Office of Naval Research
Science of Autonomy Awards N00014-13-1-0766 (Mohan Sridharan, Shiqi
Zhang) and N00014-17-1-2434 (Mohan Sridharan), the Asian Office of
Aerospace Research and Development award FA2386-16-1-4071 (Mohan
Sridharan), and the EC-funded Strands project FP7-IST-600623 (Jeremy
Wyatt).  Opinions and conclusions in this paper are those of the
authors.

%%%%%-----------------------------------------------------------------------------
%%%%%-----------------------------------------------------------------------------
\newpage
\appendix
\section{Proof of Proposition~\ref{prop:model-hist}}
\label{sec:appendix-prop-hist}
In this section, we examine Proposition~\ref{prop:model-hist}, which
states that

\medskip
\noindent {\em If $\mathcal{D}$ is a well-founded system description
  and $\mathcal{H}$ is its recorded history, then every sequence
  induced by an answer set of $\Pi(\mathcal{D}, \mathcal{H})$ is a
  model of $\mathcal{H}$.  }

\medskip
\noindent
To prove this proposition, we begin with some notation. Let $\sigma$
be a collection of literals, $\alpha_k = \langle a_0,\dots, a_k
\rangle$ be a (possibly empty) sequence of actions, $occurs(\alpha_k)
= \{occurs(a_i,i) : 0 \leq i \leq k\}$ and $\Pi(\mathcal{D}, \sigma,
\alpha_k)=_{def}\Pi^{k+1}(\mathcal{D}) \cup val(\sigma, 0) \cup
occurs(\alpha_k)$.

\begin{lemm2}\label{lem:m1}
  Let $A$ be an answer set of $\Pi(\mathcal{D},\mathcal{H})$. Then
  there exists a state $\sigma_0$ of ${\tau(\mathcal{D})}$ and a
  sequence of actions $\alpha_k= \langle a_0,\dots, a_k \rangle$ such
  that the set $B$ obtained from $A$ by removing literals formed by
  $obs$, $hpd$, $prefer$, $ab$ and $defined\_by\_default$ is an answer
  set of $\Pi(\mathcal{D},\sigma_0,\alpha_k)$.
\end{lemm2}

\noindent 
\begin{proof}[Proof of Lemma~\ref{lem:m1}]
  Let $R$ be a set of CR-rules. By $\alpha(R)$ we denote the set of
  regular ASP rules obtained by replacing $\rif$ by $\leftarrow$ in
  rules of $R$.  Let $\Pi$ be a program of CR-Prolog. By $\Pi^{reg}$
  we denote the set of all regular rules of $\Pi$. A
  cardinality-minimal set of CR-rules of $\Pi$ such that ASP program
  $\Pi(R) =_{def} \Pi^{reg}\cup \alpha(R)$ is consistent, i.e.  has an
  answer set, is called an abductive support of $\Pi$. $A$ is an
  answer set of $\Pi$ if it is an answer set of program $\Pi(R)$ for
  some abductive support $R$ of $\Pi$.  Note that this is a special
  case of a more general definition from~\cite{Gelfond:aibook14} where
  minimality of $R$ is determined by an arbitrary preference relation.
  Finally, recall that if $\sigma$ is a state of a transition diagram,
  $\sigma^{nd}$ is obtained from $\sigma$ by removing all atoms formed
  by defined fluents.

  \medskip
  \noindent 
  Now, let $A$ be an answer set of $\Pi =_{def}
  \Pi(\mathcal{D},\mathcal{H})$.  Then, by the CR-Prolog definition of
  answer sets, we have:
  \begin{enumerate}[label=(\arabic*),ref=(\arabic*)]
  \item $A$ is an answer set of an ASP program $\Pi(R) = \Pi^{reg}
    \cup \alpha(R)$ for some abductive support $R$ of $\Pi$.
    \label{eqn:prop1-lem1-1}
  \end{enumerate}
  Clearly, $\alpha(R)$ is (a possibly empty) collection of rules of
  the form: $ab(d(\bar{x})) \leftarrow val(body(d(\bar{x})),0)$. We
  will prove the existence of $\sigma_0$ and $\alpha_k$ by
  construction. Let:
  \begin{enumerate}[label=(\arabic*),ref=(\arabic*),resume]
  \item $\sigma_0 = \{f(\bar{x}) = y : val(f(\bar{x}),y,0) \in A$ or
    $f(\bar{x}) = y \in A\}$ \label{eqn:prop1-lem1-2}
  \end{enumerate}
  We will show that $\sigma_0$ is a state of ${\tau(\mathcal{D})}$,
  i.e., that:
  \begin{itemize}
  \item[(a)] $\sigma_0$ is an interpretation, i.e. for every
    $f(\bar{x})$ there is unique $y$ such that $f(\bar{x})=y \in
    \sigma_0$; and
  \item[(b)] $\sigma_0$ is the unique answer set of program
    $\Pi_c(\mathcal{D}) \cup \sigma_{0}^{nd}$.
  \end{itemize}
  To show (a) consider first an arbitrary basic fluent $f(\bar{x})$.
  Based on~\ref{eqn:prop1-lem1-2}, for every $y$, $f(\bar{x})=y \in
  \sigma_0$ iff $val(f(\bar{x}),y,0)\in A$.  Hence, we need to show
  that there is $y$ such that $val(f(\bar{x}),y,0)\in A$. There are
  two cases depending on whether or not the body of
  Statement~\ref{eqn:defined-by-default} is satisfied by $A$. In the
  former case the existence of $y$ such that $val(f,y,0)\in A$ is
  guaranteed by Statement~\ref{eqn:enc-default}; otherwise it follows
  from Statement~\ref{eqn:disj1}.  If $f(\bar{x})$ is static, we have
  that there is $y$ such that $f(\bar{x}) = y \in A$ by
  Statement~\ref{eqn:pid-static} of $\Pi(\mathcal{D})$.  If
  $f(\bar{x})$ is a defined fluent, its boolean value is included in
  $A$ by the axioms for the defined fluents
  (Statement~\ref{eqn:pid-defined}).  Uniqueness of the value assigned
  to $f(\bar{x})$ follows from Statement~\ref{eqn:pid-rule} and
  consistency of $A$.

  \medskip 
  \noindent 
  To show (b) it suffices to notice that since $A$ satisfies rules
  such as Statement~\ref{eqn:pid-constraint} in $\Pi(R)$, $\sigma_0$
  satisfies rules such as Statement~\ref{eqn:pic-state-constraint} in
  $\Pi_c(\mathcal{D})$, and hence $\Pi_c(\mathcal{D}) \cup
  \sigma_{0}^{nd}$ has an answer set.  By (a), $\sigma_0$ is complete
  and consistent and hence, by the definition of well-foundedness,
  this answer set is unique.

  \medskip
  \noindent
  Next, let:
  \begin{enumerate}[label=(\arabic*),ref=(\arabic*),resume]
  \item $\alpha_k=\langle a_0,\dots, a_k \rangle$ where $a_i = \{e_j:
    occurs(e_j,i) \in A\}$. \label{eqn:prop1-lem1-3}
  \end{enumerate}
  and let $S_0$ be a set of literals formed by $obs$ and $hpd$. Note
  that $S_0$ is a splitting set of program $\Pi(R)$.
  From~\ref{eqn:prop1-lem1-1} and the splitting set
  theorem~\cite{balduccini:lpnmr09} we have:
  \begin{enumerate}[label=(\arabic*),ref=(\arabic*),resume]
  \item $A_0$, obtained from $A$ by removing literals formed by $obs$
    and $hpd$, is an answer set of program $\Pi_{0}(R)$ obtained from
    $\Pi(R)$ by: \label{eqn:prop1-lem1-4}
    \begin{itemize}
    \item removing all atoms formed by $obs$ and $hpd$; 
    \item removing all rules whose bodies contain atoms formed of
      $obs(*,*,*)$\footnote{Recall that the ``*'' denotes a wild-card
        character.} or $hpd(*,*)$ that are not in $A$; and
    \item removing all occurrence of atoms $obs(*,*,*)$ or
      $hpd(*,*)$ from the remaining rules.
    \end{itemize}
  \end{enumerate} 
  Note that the only rules changed by this transformation belong to
  the encoding of $\mathcal{H}$.

  \medskip
  \noindent
  Next, if $\Pi_{0}^{\mathcal{H}}(R)$ denotes the program obtained
  from $\Pi_{0}(R)$ by removing all atoms formed by $occurs$ and all
  rules of $\Pi^{k+1}(\mathcal{D})$, then from~\ref{eqn:prop1-lem1-4}
  and the definition of $\Pi_{0}^{\mathcal{H}}(R)$ we have that:
  \begin{enumerate}[label=(\arabic*),ref=(\arabic*),resume]
  \item $A_0$ is an answer set of $\Pi_{0}(R) =
    \Pi^{k+1}(\mathcal{D})\cup \Pi_{0}^{\mathcal{H}}(R) \cup
    \{occurs(a,i) : occurs(a,i) \in A\}$. \label{eqn:prop1-lem1-5}
  \end{enumerate}

  \medskip
  \noindent
  Now let $S_1$ be the set of atoms formed by statics, $prefer$, $ab$,
  $defined\_by\_default$, and $val(*,*,0)$.  It is not difficult to
  check that $S_1$ is a splitting set of $\Pi_{0}(R)$. It divides the
  program into two parts:
  \begin{itemize}
  \item Program $Bot$ consisting of $\Pi_{0}^{\mathcal{H}}(R)$
    combined with the set $Zero$ of instances of axioms encoded in
    Statements~\ref{eqn:pid-constraint},~\ref{eqn:pid-defined},
    ~\ref{eqn:pid-static} and~\ref{eqn:pid-rule} from
    $\Pi^{k+1}(\mathcal{D})$ with the time-step variable set to $0$.
  \item Program $Top = (\Pi^{k+1}(\mathcal{D}) \setminus Zero) \cup
    \{occurs(a,i) : occurs(a,i) \in A\}$
  \end{itemize}
  So, by the splitting set theorem, we now have:
  \begin{enumerate}[label=(\arabic*),ref=(\arabic*),resume]
  \item $A_0$ is an answer set of program $B_0 \cup Top$ where $B_0$
    is an answer set of $Bot$. \label{eqn:prop1-lem1-6}
  \end{enumerate}

  \medskip
  \noindent
  Next, observe that:
  \begin{enumerate}[label=(\arabic*),ref=(\arabic*),resume]
  \item $B_0$ can be partitioned into $B_1$ and $B_2$ with $B_1$
    consisting of atoms of $A_0$ formed by $prefer$, $ab$, and
    $defined\_by\_default$ and $B_2$ consisting of atoms of $A_0$
    formed by statics and $val(*,*,0)$. \label{eqn:prop1-lem1-7}
  \end{enumerate}
  Using definition of answer sets for ASP program, it can be proved
  that for any two programs $\Pi_1$ and $\Pi_2$ whose signatures
  $\Sigma_1$ and $\Sigma_2$ are disjoint, $X$ is an answer set of a
  program $\Pi_1 \cup \Pi_2$ iff $X|_{\Sigma_1}$ and $X|_{\Sigma_2}$
  are answer sets of $\Pi_1$ and $\Pi_2$ respectively.  Hence, we have
  that:
  \begin{enumerate}[label=(\arabic*),ref=(\arabic*),resume]
  \item $B = A_0 \setminus B_1$ is an answer set of $B_2 \cup Top$.
    \label{eqn:prop1-lem1-8}
  \end{enumerate}
  From~\ref{eqn:prop1-lem1-4},~\ref{eqn:prop1-lem1-7}
  and~\ref{eqn:prop1-lem1-8} above, we have that $B$ is obtained from
  $A$ by removing literals formed by $obs$, $hpd$, $prefer$, $ab$ and
  $defined\_by\_default$.

  \medskip
  \noindent
  To show that $B$ is an answer set of
  $\Pi(\mathcal{D},\sigma_0,\alpha_k)$ we first demonstrate that:
  \begin{enumerate}[label=(\arabic*),ref=(\arabic*),resume]
  \item $B$ is an answer set of $\Pi_{1}^*=_{def} B_2 \cup
    \Pi^{k+1}(\mathcal{D}) \cup \{occurs(a,i) : occurs(a,i) \in A\} $.
    \label{eqn:prop1-lem1-9}
  \end{enumerate}
  By construction, we have that:
  \begin{enumerate}[label=(\arabic*),ref=(\arabic*),resume]
  \item $\Pi_{1}^* = B_2 \cup Top \cup Zero$.
    \label{eqn:prop1-lem1-10}
  \end{enumerate}
  To prove~\ref{eqn:prop1-lem1-9}, we will show that $B$ is an answer
  set of the reduct, $(\Pi_{1}^*)^{B}$ of $\Pi_{1}^*$ with respect to
  $B$ (note that this is the definition of answer set).
  % (i.e. $\subseteq$- minimal set satisfying 
  Based on the definition of the reduct and~\ref{eqn:prop1-lem1-10},
  we have:
  \begin{enumerate}[label=(\arabic*),ref=(\arabic*),resume]
  \item $(\Pi_{1}^*)^{B} = B_2 \cup Zero^{B} \cup Top^{B}$.
    \label{eqn:prop1-lem1-11}
  \end{enumerate}  
  From~\ref{eqn:prop1-lem1-8} and the definition of answer set, we
  have that $B$ is a $\subseteq$-minimal set satisfying $B_2 \cup
  Top^{B}$.  Then, based
  on~\ref{eqn:prop1-lem1-6}-\ref{eqn:prop1-lem1-8}, we have that $B_2$
  (and hence $B$) also satisfies $Zero^{B}$, and
  thus~\ref{eqn:prop1-lem1-9} holds.

  \medskip
  \noindent
  Then, based on~\ref{eqn:prop1-lem1-2},~\ref{eqn:prop1-lem1-6}
  and~\ref{eqn:prop1-lem1-7}, we have that $val(\sigma_0,0) = B_2$
  which, together with~\ref{eqn:prop1-lem1-9}, implies that:
  \begin{enumerate}[label=(\arabic*),ref=(\arabic*),resume]
  \item $B$ is an answer set of
    $\Pi^{k+1}(\mathcal{D},\sigma_0,\alpha_k)$.
  \end{enumerate}
  This completes the proof of Lemma~\ref{lem:m1}. 
\end{proof}

%%%%--------------------------------------

\begin{lemm2}\label{lem:m2}
  Let $B$ be an answer set of $\Pi(\mathcal{D},\sigma_0,\alpha_k)$,
  $B_k$ be obtained from $B$ by removing all literals containing
  time-step $k+1$, and $\Pi_{k}^{k+1}(\mathcal{D},\sigma_k,a_k)$ be
  $\Pi(\mathcal{D},\sigma_k,a_k)$ with time-steps $0$ and $1$ replaced
  by $k$ and $k+1$ respectively.  Then:
  \begin{s_itemize}
  \item $B_k$ is an answer set of
    $\Pi(\mathcal{D},\sigma_0,\alpha_{k-1})$.
  \item $B = B_{k} \cup U$ where $U$ is an answer set of
    $\Pi_{k}^{k+1}(\mathcal{D},\sigma_k,a_k)$.
  \end{s_itemize}
\end{lemm2}

\begin{proof}[Proof of Lemma~\ref{lem:m2}]
  Let $S$ be a set of literals of $\Pi(\mathcal{D},\sigma_0,\alpha_k)$
  not containing time step $k+1$. It is easy to check that $S$ is a
  splitting set of this program, which divides it into two parts:
  \begin{enumerate}[label=(\arabic*),ref=(\arabic*)]
  \item $Bot = \Pi(\mathcal{D},\sigma_0,\alpha_{k-1})$ and $Top =
    \Pi(\mathcal{D},\sigma_0,\alpha_k) \setminus
    Bot$.\label{eqn:prop1-lem2-1}
  \end{enumerate}
  By the splitting set theorem and definition of $B_k$, we have:
  \begin{enumerate}[label=(\arabic*),ref=(\arabic*),resume]
  \item $B_k$ is an answer set of
    $Bot=\Pi(\mathcal{D},\sigma_0,\alpha_{k-1})$.
    \label{eqn:prop1-lem2-2}
  \end{enumerate}
  and:
  \begin{enumerate}[label=(\arabic*),ref=(\arabic*),resume]
  \item $B$ is an answer set of the program $B_k \cup Top$.
    \label{eqn:prop1-lem2-3}
  \end{enumerate}
  By definition, $\sigma_k = \{f(\bar{x}) = y : val(f(\bar{x}),y,k)
  \in B\} \cup \{f(\bar{x}) = y : f(\bar{x}) = y \in B\}$, and hence,
  $val(\sigma_k,k)$ is a subset of $B$ and of $B_k$. Thus, we have:
  \begin{enumerate}[label=(\arabic*),ref=(\arabic*),resume]
  \item $B_k \cup Top = B_k \cup val(\sigma,k) \cup Top = B_k \cup
    \Pi_{k}^{k+1}(\mathcal{D},\sigma_k,a_k)$. \label{eqn:prop1-lem2-4}
  \end{enumerate}

  \medskip
  \noindent 
  Now let:
  \begin{enumerate}[label=(\arabic*),ref=(\arabic*),resume]
  \item $B_k = B_k^\prime \cup B_k^{\prime\prime}$
    \label{eqn:prop1-lem2-5}
  \end{enumerate}
  where $B_k^\prime$ consists of atoms of $B_k$ containing time-steps
  smaller than $k$ and $B_k^{\prime\prime} = B_k \setminus
  B_k^{\prime}$. Note that $B_k^{\prime\prime}$ consists of atoms of
  $B_k$ formed by statics and of those containing time-step $k$.
  From~\ref{eqn:prop1-lem2-4},~\ref{eqn:prop1-lem2-5}, and the
  definition of $\sigma_k$, we then have:
  \begin{enumerate}[label=(\arabic*),ref=(\arabic*),resume]
  \item $B_k \cup Top = B_{k}^\prime \cup
    \Pi_{k}^{k+1}(\mathcal{D},\sigma_k,a_k)$. \label{eqn:prop1-lem2-6}
  \end{enumerate}
  Based on~\ref{eqn:prop1-lem2-3} and~\ref{eqn:prop1-lem2-6}, we have:
  \begin{enumerate}[label=(\arabic*),ref=(\arabic*),resume]
  \item $B$ is an answer set of $B_{k}^\prime \cup
    \Pi_{k}^{k+1}(\mathcal{D},\sigma_k,a_k)$. \label{eqn:prop1-lem2-7}
  \end{enumerate}

  \medskip
  \noindent 
  Since, by construction, the signatures of $B_{k}^\prime$ and
  $\Pi_{k}^{k+1}(\mathcal{D},\sigma_k,a_k)$ are disjoint,
  from~\ref{eqn:prop1-lem2-7}, we have:
  \begin{enumerate}[label=(\arabic*),ref=(\arabic*),resume]
  \item $B = B_{k} \cup U$ where $U$ is an answer set of
    $\Pi_{k}^{k+1}(\mathcal{D},\sigma_k,a_k)$.
    \label{eqn:prop1-lem2-8}
  \end{enumerate}
  This completes the proof of Lemma~\ref{lem:m2}.
\end{proof}

\noindent
\begin{proof}[Proof of Proposition \ref{prop:model-hist}]
  Let $\mathcal{D}$ and $\mathcal{H}$ be as in the proposition, $A$ be
  an answer set of CR-Prolog program $\Pi(\mathcal{D}, \mathcal{H})$,
  and $M = \langle
  \sigma_0,a_0,\sigma_1,\dots,\sigma_{n},a_n,\sigma_{n+1}\rangle$ be a
  sequence induced by $A$. We will show that $M$ is a model of
  $\mathcal{H}$, i.e. $M$ is a path of transition diagram
  ${\tau(\mathcal{D})}$ (definition \ref{def:model}).

  \medskip\noindent
  The proposition will be an immediate consequence 
  of a more general statement:
  \begin{enumerate}[label=(\arabic*),ref=(\arabic*)]
  \item \emph{for every $0 \leq k \leq n+1$ $M_k = \langle
      \sigma_0,a_0,\sigma_1,\dots, \sigma_{k}\rangle$ is a path in
      ${\tau(\mathcal{D})}$}. \label{eqn:prop1-1}
  \end{enumerate}
  Before proceeding with inductive proof of~\ref{eqn:prop1-1}, let us
  notice that, by Lemma~\ref{lem:m1}:
  \begin{enumerate}[label=(\arabic*),ref=(\arabic*),resume]
  \item $M$ is induced by an answer set $B$ of an ASP program
    $\Pi(\mathcal{D},\sigma_0,\alpha_n)$ where $\sigma_0$ is a state
    and $B$ is obtained from $A$ by removing atoms formed by $obs$,
    $hpd$, $prefer$, $ab$ and $defined\_by\_default$.
    \label{eqn:prop1-2}
  \end{enumerate}       
  %\medskip
%  \noindent 
  We use induction on $k$. The base case is: $k=0$, i.e. $M_k =
  \langle \sigma_0 \rangle$.  Then~\ref{eqn:prop1-1} follows
  immediately from~\ref{eqn:prop1-2}.

  \medskip
  \noindent 
  Nex, consider the inductive step: let $k > 0$ and $M_k = \langle
  \sigma_0,a_0,\sigma_1,\dots, \sigma_{k-1},a_{k-1},
  \sigma_{k}\rangle$.  By inductive hypothesis:
  \begin{enumerate}[label=(\arabic*),ref=(\arabic*),resume]
  \item $M_k = \langle \sigma_0,a_0,\sigma_1,\dots,
    \sigma_{k-1}\rangle$ is a path in ${\tau(\mathcal{D})}$.
    \label{eqn:prop1-3}
  \end{enumerate}
  We need to show that $L=\langle \sigma_{k-1},a_{k-1},
  \sigma_{k}\rangle$ is a transition of ${\tau(\mathcal{D})}$. By
  Lemma~\ref{lem:m2}, we have:
  \begin{enumerate}[label=(\arabic*),ref=(\arabic*),resume]
  \item $L$ is induced by an answer set $U_0$ of
    $\Pi_{k}^{k+1}(\mathcal{D},\sigma_k,a_k)$. \label{eqn:prop1-4}
  \end{enumerate}
  Let $U$ be obtained from $U_0$ by replacing time-steps $k$ and $k+1$
  by $0$ and $1$ respectively. From~\ref{eqn:prop1-4} and the
  definition of $\Pi_{k}^{k+1}(\mathcal{D},\sigma_k,a_k)$, we have
  that:
  \begin{enumerate}[label=(\arabic*),ref=(\arabic*),resume]
  \item $L$ is induced by an answer set $U$ of
    $\Pi(\mathcal{D},\sigma_k,a_k)$. \label{eqn:prop1-5}
  \end{enumerate}
  From (3) we have that:
  \begin{enumerate}[label=(\arabic*),ref=(\arabic*),resume]
  \item $\sigma_k$ is a state. \label{eqn:prop1-6}
  \end{enumerate}
  To prove that $\sigma_{k+1}$ is a state we first show that
  $\sigma_{k+1}$ is an interpretation, i.e. for every $f(\bar{x})$
  there is unique $y$ such that $val(f(\bar{x}),y,1)\in U$.
  From~\ref{eqn:prop1-5} and~\ref{eqn:prop1-6}, we have that, for
  every $f(\bar{x})$ there is unique $y_1$ such that
  $val(f(\bar{x}),y,0)\in U$.  If the body of the inertia axiom for
  $f(\bar{x})$ is satisfied by $U$ then $val(f(\bar{x}),y_1,1)\in U$.
  Otherwise, the inertia axiom is defeated by
  Statement~\ref{eqn:pid-rule} and hence $val(f(\bar{x}),y_2,1)\in U$.
  Thus, we have that:
  \begin{enumerate}[label=(\arabic*),ref=(\arabic*),resume]
  \item $\sigma_{k+1}$ is an interpretation. \label{eqn:prop1-7}
  \end{enumerate}
%  \medskip
%  \noindent 
  The last step is to show that:
  \begin{enumerate}[label=(\arabic*),ref=(\arabic*),resume]
  \item $\sigma_{k+1}$ is the unique answer set of program
    $\Pi_c(\mathcal{D}) \cup \sigma_{k+1}^{nd}$. \label{eqn:prop1-8}
  \end{enumerate}
  To do that it suffices to notice that, since $U$ satisfies rules
  such as Statements~\ref{eqn:pid-defined} and
  \ref{eqn:pid-constraint} in $\Pi(\mathcal{D},\sigma_k,a_k)$,
  $\sigma_{k+1}$ satisfies rules such as
  Statements~\ref{eqn:pic-state-constraint} and
  \ref{eqn:pic-defined-fluent} in $\Pi_c(\mathcal{D})$, and hence
  $\Pi_c(\mathcal{D}) \cup \sigma_{k+1}^{nd}$ has an answer set. Based
  on~\ref{eqn:prop1-7}, $\sigma_{k+1}$ is complete and consistent and
  hence, by the definition of well-foundedness, this answer set is
  unique; this proves~\ref{eqn:prop1-8}. Then, based
  on~\ref{eqn:prop1-7} and~\ref{eqn:prop1-8}, and the definition of
  state, we have:
  \begin{enumerate}[label=(\arabic*),ref=(\arabic*),resume]
  \item $\sigma_{k+1}$ is a state. \label{eqn:prop1-9}
  \end{enumerate}
  Thus, based on~\ref{eqn:prop1-5}, ~\ref{eqn:prop1-6},
  ~\ref{eqn:prop1-9} and Definition~\ref{def:ald-trans}, we have that:
  \begin{enumerate}[label=(\arabic*),ref=(\arabic*),resume]
  \item $L$ is a transition. \label{eqn:prop1-10}
  \end{enumerate}
%  \medskip
%  \noindent 
  Next, based on~\ref{eqn:prop1-3}, the definition of $L$,
  and~\ref{eqn:prop1-10}: 
  \begin{enumerate}[label=(\arabic*),ref=(\arabic*),resume]
  \item $M_k$ is a path in ${\tau(\mathcal{D})}$. \label{eqn:prop1-11}
  \end{enumerate}
  This completes the proof of statement~\ref{eqn:prop1-1}. 
%  \medskip
%  \noindent
  Based on the definition of $M_k$, $M = M_{n+1}$, and based
  on~\ref{eqn:prop1-1}, $M$ is a path in ${\tau(\mathcal{D})}$.  Since
  $M$ is induced by $A$, based on Definition~\ref{def:model}, it is a
  model of $\mathcal{H}$.  This completes the proof of
  Proposition~\ref{prop:model-hist}.
\end{proof}

%%%%%-----------------------------------------------------------------------------
%%%%%-----------------------------------------------------------------------------
\vspace{-1em}
\section{Proof of Proposition~\ref{prop:plan-reduce}}
\label{sec:appendix-prop-plan}
In this section, we examine Proposition~\ref{prop:plan-reduce}, which
states that:

\medskip
\noindent {\em Let $\mathcal{P} = (\mathcal{D}, \mathcal{H}, h, G)$ be
  a planning problem with a well-founded, deterministic system
  description $\mathcal{D}$. A sequence $\langle a_0,\dots,a_{k-1}
  \rangle$ where $k < h$ is a solution of $\mathcal{P}$ iff there is
  an answer set $A$ of $Plan$ such that:
  \begin{enumerate}
  \item For any $n < i \leq n+k$, $occurs(a_i, i-1) \in A$,
  \item $A$ contains no other atoms of the form $occur(*, i)$ with $i \geq n$.
  \end{enumerate}
}

\medskip
\noindent
We begin by introducing some notation. Let $\Pi$ be an arbitrary
CR-Prolog program and $R$ be a collection of CR-rules from $\Pi$.
Similar to the terminology in~\cite{Gelfond:aibook14}, we use
$\Pi^{reg}$ to denote the collection of regular rules of $\Pi$ and
$\alpha(R)$ to denote the set of regular ASP rules obtained by
replacing $\rif$ by $\leftarrow$ in CR-rules of $R$. For completeness,
recall that for any program $\Pi$, we have $\Pi(R) = \Pi^{reg} \cup
\alpha(R)$. Also recall, from Section~\ref{sec:arch-hl-reason}, that:
\begin{align*}
  &Plan = Diag ~\cup~ Classical\_plan ~\cup~ \{DPC\}\\
  &Diag =_{def}\Pi^{n}(\mathcal{D}, \mathcal{H})\\
  &Classical\_plan = \Pi^{[n..n+h]}(\mathcal{D}) ~\cup~ goal(I)
  \leftarrow val(f(\bar{x}), y, I) ~\cup~ PM\\
  &\leftarrow Y = count\{X : ab(X)\}, Y > m~~~~~~~~~\textrm{\% DPC}
\end{align*}
where $n$ is the current step of $\mathcal{H}$, $m$ is the size of the
abductive support of $Diag$, and $PM$ is the planning module. We will
also need the following Lemma to prove
Proposition~\ref{prop:plan-reduce}.

\begin{lemm2}\label{lem:m3}
  {\rm For any set $R$ of CR-rules of $Diag$, $A$ is an answer set of
    ASP program $Plan(R)$ iff $A = A_0 \cup B_{0}$ where $A_0$ is an
    answer set of $Diag(R)$ satisfying $DPC$ and $B_{0}$ is an answer
    set of $Shifted\_plan =_{def} \{ val(f(\bar{x}),y,n) :
    val(f(\bar{x}),y,n)\in A_0\} \cup Classical\_plan$.  }
\end{lemm2}

\begin{proof}[Proof of Lemma~\ref{lem:m3}]
  Let $S_0$ be the set of literals of $Plan(R)$ not containing atoms
  with time steps greater than $n$ or atoms of the form $occurs(*, n)$
  and $hpd(*, n)$. It is easy to check that $S_0$ is a splitting set
  of $Plan(R)$ which splits the program into two parts, $Bot = Diag(R)
  \cup \{DPC\}$ and $Top=Classical\_plan$. By the splitting set
  theorem, $A$ is an answer set of $Plan(R)$ iff $A$ is an answer set
  of $A_0 \cup Top$ where $A_0$ is an answer set of $Bot$.  Clearly,
  $A_0 \cup Top = A_0 \cup Shifted\_plan$. Since $A_0$ is a collection
  of atoms, from the definition of answer set we have that $A=A_0 \cup
  B_0$ where $B_0$ is an answer set of $Shifted\_plan$.
\end{proof}

\noindent
Next, we turn to proving Proposition~\ref{prop:plan-reduce}.

\begin{proof}[Proof of Proposition~\ref{prop:plan-reduce}]
  Let $\mathcal{P}$ and $Plan$ be as in the proposition, $\sigma$ be a
  state and $\langle a_0,\dots,a_{k-1} \rangle$ with $k < h$ be a
  sequence of actions of $\mathcal{D}$.

  \medskip\noindent
  Based on Definition~\ref{def:planning}:

  \begin{enumerate}[label=(\arabic*),ref=(\arabic*)]
  \item $\langle a_0,\dots,a_{k-1} \rangle$ is a solution of
    $\mathcal{P}$ iff:\label{eqn:prop2-1}
    \begin{itemize}
    \item[(a)] there is a state $\sigma$ that is the current state of
      some model $M$ of $\mathcal{H}$; and \label{eqn:prop2-1a}
    \item[(b)] $\langle a_0,\dots,a_{k-1} \rangle$ is a solution of
      classical planning problem $\mathcal{P}_c =
      (\mathcal{D},\sigma,G)$ with horizon $h$. \label{eqn:prop2-1b}
    \end{itemize}
  \end{enumerate}
  Based on Definition~\ref{def:model} and the well-foundedness of
  $\mathcal{D}$, Statement~\ref{eqn:prop2-1a}(a) holds iff:
  \begin{enumerate}[label=(\arabic*),ref=(\arabic*),resume] % (2)
  \item $M$ is induced by some answer set $A_0$ of $Diag$, $n$ is the
    current step of history from $\mathcal{P}$, and:
    \begin{align*}
      \sigma = \{f(\bar{x})=y : val(f(\bar{x}),y,n)\in A_0\}
    \end{align*}
    \label{eqn:prop2-2}
  \end{enumerate}
  By the CR-Prolog definition of answer sets,
  Statement~\ref{eqn:prop2-2} holds iff:
  \begin{enumerate}[label=(\arabic*),ref=(\arabic*),resume] % (3)
  \item $A_0$ is an answer set of $Diag(R)$ for some abductive support
    $R$ of $Diag$ and $\sigma = \{f(\bar{x})=y :
    val(f(\bar{x}),y,n)\in A_0\}$ (since $A_0$ is an answer set of
    $Diag$ it satisfies $DPC$).  \label{eqn:prop2-3}
  \end{enumerate}

  \noindent 
  Based on Proposition $9.1.1$ from~\cite{Gelfond:aibook14},
  Statement~\ref{eqn:prop2-1}(b) holds iff:
  \begin{enumerate}[label=(\arabic*),ref=(\arabic*),resume] % (4)
  \item There is an answer set $S$ of ASP program $plan(\mathcal{P}_c,
    h)$ such that: \label{eqn:prop2-4}
    \begin{itemize}
    \item[(a)] For any $0 < i \leq k$, $occurs(a_i,i-1) \in S$; and
    \item[(b)] $S$ contains no other atoms formed by $occurs$.
    \end{itemize} 
  \end{enumerate}

  \medskip
  \noindent 
  Consider an ASP program:
  \begin{align*}
    Shifted\_plan =_{def} \{ val(f(\bar{x}),y,n) :
    val(f(\bar{x}),y,n)\in A_0\} ~\cup~ Classical\_plan
  \end{align*}
  It is easy to see that this program differs from
  $plan(\mathcal{P}_c, h)$ only in the domain of its time-step
  variables.  In the former case, such variables range over $[n,n+h]$
  while in the latter the range is $[0,h]$. The programs are
  isomorphic and hence Statement~\ref{eqn:prop2-4} holds for $S$ iff:
  \begin{enumerate}[label=(\arabic*),ref=(\arabic*),resume] % (5)
  \item $B_0$ obtained from $S$ by increasing all occurrences of time
    steps in atoms from $S$ by $n+1$ is an answer set of
    $Shifted\_plan$. Also:
    \begin{itemize}
    \item[(a)] For any $n < i \leq n+k$, $occurs(a_i,i-1) \in B_0$; and
    \item[(b)] $B_0$ contains no other atoms of the form
      $occurs(*, i)$ where $i \geq n$.
    \end{itemize}
    \label{eqn:prop2-5}
  \end{enumerate}
  Now we have that: 
  \begin{enumerate}[label=(\arabic*),ref=(\arabic*),resume] % (6)
  \item Statement~\ref{eqn:prop2-1} is true iff
    Statement~\ref{eqn:prop2-3} and Statement~\ref{eqn:prop2-5} are
    true.
    \label{eqn:prop2-6}
  \end{enumerate}
  \medskip
  \noindent 
  Let $A=A_0 \cup B_0$. Then, based on Lemma~\ref{lem:m3}, we have:
  \begin{enumerate}[label=(\arabic*),ref=(\arabic*),resume] % (7)
  \item Statements~\ref{eqn:prop2-3} and~\ref{eqn:prop2-5} are true
    iff $A$ is an answer set of $Plan(R)$.
    \label{eqn:prop2-7}
  \end{enumerate}
  Based on~\ref{eqn:prop2-7}, we have:
  \begin{enumerate}[label=(\arabic*),ref=(\arabic*),resume] % (8)
  \item Statement~\ref{eqn:prop2-1} is true iff $A$ is an answer set
    of $Plan(R)$. \label{eqn:prop2-8}
  \end{enumerate}
  However, since every answer set of $Plan$ must satisfy $DPC$,
  $Plan(R)$ has an answer set iff $R$ is an abductive support of
  $Plan$. Hence:
  \begin{enumerate}[label=(\arabic*),ref=(\arabic*),resume] % (9)
  \item $A$ is an answer set of $Plan(R)$ iff $A$ is an answer set of
    $Plan$. \label{eqn:prop2-9}
  \end{enumerate}
  From the construction of $A$, Statement~\ref{eqn:prop2-5}, and the
  fact that $A_0$ contains no atoms of the form $occurs(*, i)$ where
  $i \geq n$, we have that $A$ satisfies the conditions of the
  proposition. This completes the proof of
  Proposition~\ref{prop:plan-reduce}.
\end{proof}

%%%%%-----------------------------------------------------------------------------
%%%%%-----------------------------------------------------------------------------
\section{Proof of Proposition~\ref{prop:weak-refinement}}
\label{sec:appendix-prop-weak-refine}
In this section, we prove Proposition~\ref{prop:weak-refinement},
which states that:

\medskip
\noindent 
{\em Let $\mathcal{D}_H$ and $\mathcal{D}_{L, nobs}$ be coarse and
  fine resolution system descriptions from our running example.  Then
  $\tau_L$ is a weak refinement of $\tau_H$.  }

\medskip
\noindent
\begin{proof}[Proof of Proposition~\ref{prop:weak-refinement}]
  Definitions in $\mathcal{D}_H$ and $\mathcal{D}_{L, nobs}$ contain
  no dependency between defined domain functions and their negations.
  Both system descriptions are therefore weakly-acyclic and thus
  well-founded, which justifies the use of the following property in
  the proof. Let $\mathcal{D}$ over signature $\Sigma$ be a
  well-founded system description defining transition diagram $\tau$.
  Then, an interpretation $\delta$ of $\Sigma$ is a state of $\tau_H$
  iff:
  \begin{itemize}
  \item $\delta$ satisfies constraints of $\mathcal{D}$; and
  \item For every defined fluent $f$ of $\Sigma$, $f(\bar{u})\in
    \delta$ iff there is a rule from the definition of $f(\bar{u})$
    whose body is satisfied by the interpretation $\delta$.
  \end{itemize}
  For readability, we also repeat Definition~\ref{def:weak-refinement}
  of weak refinement of a transition diagram.  A transition diagram
  $\tau_{L, nobs}$ over $\Sigma_{L, nobs}$ is called a \emph{weak
    refinement} of $\tau_H$ if:
  \begin{enumerate}
  \item For every state $\sigma^\diamond$ of $\tau_{L, nobs}$, the
    collection $\sigma^\diamond |_{\Sigma_H}$ of atoms of
    $\sigma^\diamond$ formed by symbols from $\Sigma_H$ is a state of
    $\tau_H$.
  
  \item For every state $\sigma$ of $\tau_H$, there is a state
    $\sigma^\diamond$ of $\tau_{L, nobs}$ such that $\sigma^\diamond$
    is an extension of $\sigma$.

  \item For every transition $T = \langle \sigma_1,a^H,\sigma_2
    \rangle$ of $\tau_H$, if $\sigma^{\diamond}_1$ and
    $\sigma^{\diamond}_2$ are extensions of $\sigma_1$ and $\sigma_2$
    respectively, then there is a path $P$ in $\tau_{L, nobs}$ from
    $\sigma^{\diamond}_1$ to $\sigma^{\diamond}_2$ such that:
    \begin{itemize}
    \item actions of $P$ are concrete, i.e., directly executable by
      robots; and
    \item $P$ is \emph{pertinent} to $T$, i.e., all states of $P$ are
      extensions of $\sigma_1$ or $\sigma_2$.
    \end{itemize}
  \end{enumerate}

  \medskip
  \noindent 
  To prove the first clause of Definition~\ref{def:weak-refinement},
  let $\sigma^{\diamond}$ and $\sigma = \sigma^{\diamond}|_{\Sigma_H}$
  be as in the first clause. To prove that $\sigma$ is a state of
  $\mathcal{D}_H$, we show that it satisfies the clauses of the
  property described above. We start with the constraint in
  Statement~\ref{eqn:logician-constraint}(a) for a particular object
  $ob$:
  \begin{align*}
    loc(ob) = P~~\mathbf{if}~~ loc(rob_1)=P, ~in\_hand(rob_1, ob)
  \end{align*}
  
  \noindent
  Let: 
  \begin{enumerate}[label=(\roman*),ref=(\roman*)]
  \item \label{eqn:prop3-1-i} $(loc(rob_1)= P) \in \sigma$; and 
  \item \label{eqn:prop3-1-ii} $in\_hand(rob_1, ob) \in \sigma$.
  \end{enumerate}
  To show that $(loc(ob)= P) \in \sigma$ let $c_1$ be the value of
  $loc^*(rob_1)$ in $\sigma^{diamond}$, i.e:

  \begin{enumerate}[label=(\roman*),ref=(\roman*),resume]
  \item \label{eqn:prop3-1-iii} $(loc^*(rob_1)= c_1) \in
    \sigma^{\diamond}$
  \end{enumerate}
  Based on the bridge axiom in
  Statement~\ref{eqn:refine-bridge-specific}(a):
  \begin{align*}
    loc(rob_1)= P~~\mathbf{if}~~ loc^*(rob_1)=C, component(C, P)
  \end{align*}
  of $\mathcal{D}_{L,nobs}$ and conditions $(i)$ and $(iii)$, we have:

  \begin{enumerate}[label=(\roman*),ref=(\roman*),resume]
  \item \label{eqn:prop3-1-iv} $component(c_1, P)$
  \end{enumerate}
  Suppose this is not the case. Then, based on the definition of
  $place^*$, there is some place $P_2$ such that $component(c_1,P_2)
  \in \sigma^\diamond$.  This statement, together with $(iii)$ and the
  bridge axiom in Statement~\ref{eqn:refine-bridge-specific}(a) will
  entail $(loc(rob) = P_2)$, which contradicts condition $(i)$ above.

  \medskip
  \noindent 
  Next, the state constraint in
  Statement~\ref{eqn:refine-constraint-specific}(a):
  \begin{align*}
    loc^*(ob)= P ~~\mathbf{if}~~ loc^*(rob_1)=P, ~in\_hand(rob_1, ob)
  \end{align*}
  of $\mathcal{D}_{L,nobs}$, together with $(ii)$ and $(iii)$ imply:
  \begin{enumerate}[label=(\roman*),ref=(\roman*),resume]
  \item \label{eqn:prop3-1-v} $(loc^*(ob)=c_1) \in \sigma^\diamond$
  \end{enumerate}
  Then, the bridge axiom in
  Statement~\ref{eqn:refine-bridge-specific}(a), together with $(iv)$
  and $(v)$ imply that $(loc(ob)= P) \in \sigma$ and hence $\sigma$
  satisfies the first constraint of $\mathcal{D}_H$.

  \medskip
  \noindent 
  Next, consider the definition of the static $next\_to(P_1, P_2)$ in
  $\mathcal{D}_H$. It our example domain with four rooms (see
  Figure~\ref{fig:refine-gridmap}), $\mathcal{D}_H$ consists of
  statements such as:
  \begin{align*}
    &next\_to(r_1, r_2)\\
    &next\_to(r_1, r_3)\\
    &next\_to(r_2, r_4)\\
    &next\_to(r_3, r_4)\
  \end{align*}
  and the constraint:
  \begin{align*}
    next\_to(P_1, P_2) ~~\mathbf{if}~~ next\_to(P_2, P_1)
  \end{align*}
  In the fine-resolution system description $\mathcal{D}_{L, nobs}$,
  these statements are replaced by a collection of statements of the
  form $next\_to^*(c_i,c_j)$, state constraint in
  Statement~\ref{eqn:refine-constraint-specific}(b):
  \begin{align*}
    next\_to^*(C_1, C_2) ~~\mathbf{if}~~ next\_to^*(C_2, C_1)
  \end{align*}
  and a bridge axiom as described by
  Statement~\ref{eqn:refine-bridge-specific}(b):
  \begin{align*}
    next\_to(P_1,P_2) ~~\mathbf{if}~~ next\_to^*(C_1, C_2),
    component(C_1, P_1), component(C_1, P_2)
  \end{align*}
  The last axiom implies that $next\_to(r_i, r_j) \in \sigma$ iff
  $\sigma^\diamond$ indicates that there are two adjacent cells in the
  domain such that one of them is in $r_i$ and another is in $r_j$.
  This is the situation in our example domain, as shown in
  Figure~\ref{fig:refine-gridmap}.  
  % Finally, the Statement~\ref{eqn:coarse-}:
%   \begin{align*}
%     can\_be\_observed_{loc}(rob_1, Th, P) ~~\mathbf{if}~~ loc(rob_1) = P
%   \end{align*}
%   in $\mathcal{D}_H$ is replaced in $\mathcal{D}_{L, nobs}$ by:
%   \begin{align*}
%     can\_be\_observed_{loc^*}(rob_1, Th, C)~~\mathbf{if}~~
%     loc^*(rob_1) = C
%   \end{align*}
%   and the bridge axiom:
%   \begin{align*}
%     can\_be\_observed_{loc(Th)}(Rob,Room) ~~\mathbf{if} ~~
%     can\_be\_observed_{loc^*(Th)}(Rob,Cell), component(Cell,Room)
%   \end{align*}
%   We have already demonstrated that $loc(rob,room) \in \delta$ iff for
%   some component $cell$ of $room$, $loc(rob,cell) \in
%   \sigma^\diamond$.  using this, together with the constraint, we have
%   $can\_be\_observed_{loc^*(Th)}(Rob, Cell) \in \sigma^\diamond$.
%   Together with the bridge axiom this imply
%   $can\_be\_observed_{loc(Th)}(rob, room) \in \delta$.  
  This concludes the proof of the first clause of
  Definition~\ref{def:weak-refinement}.

  \medskip
  \noindent
  To prove clause 2 of Definition~\ref{def:weak-refinement}, consider
  a state $\sigma$ of $\tau_H$ and expand it to a state
  $\sigma^\diamond$ of $\tau_{L, nobs}$, and show that
  $\sigma^\diamond$ is a state of $\tau_{L, obs}$. We do so by
  construction by interpreting the fine-resolution domain functions of
  $\mathcal{D}_{L, nobs}$ such that it satisfies the bridge axioms,
  constraints and definitions of $\mathcal{D}_{L, nobs}$.  In our
  example domain, it is sufficient to map $loc^*(thing)$ to a cell $c$
  of room $r$ such that:
  \begin{itemize}
  \item if $loc^*(th)=c$ and $component(c, rm)$ are in
    $\sigma^\diamond$ then $loc(th)=rm \in \sigma$

  \item if $in\_hand(rob_1, ob) \in \sigma$ then the same cell is
    assigned to $rob_1$ and $ob$.
  \end{itemize}

  \medskip
  \noindent
  The definition of static $next\_to^*$ is the same for every state.
  It is symmetric and satisfies
  Statement~\ref{eqn:refine-bridge-specific}(b) describing the bridge
  axiom for $next\_to$. In other words, all state constraints and
  definitions of $\mathcal{D}_{L, nobs}$ are satisfied by
  $\sigma^\diamond$, which is thus a state of $\tau_{L, nobs}$.

  \medskip
  \noindent
  To prove the last clause of Definition~\ref{def:weak-refinement},
  consider a transition $T = \langle \sigma_1,move(rob, r_2), \sigma_2
  \rangle$ of $\tau_H$ and let $\sigma^{\diamond}_1$ and
  $\sigma^{\diamond}_2$ be states of $\tau_{L, nobs}$ expanding
  $\sigma_1$ and $\sigma_2$ respectively. Assume that the robot is in
  cell $c_1$ of room $r_1$ and that the robot's desired position in
  $\sigma^{\diamond}_2$ is $c_2$.  The required path $P$ then will
  consist of a sequence of moves of the form $move^*(rob_1, c_i)$
  which starts with robot being at $c_1$ and ends with it being at
  $c_2$.  Due to executability condition encoded in
  Statement~\ref{eqn:logician-executability}(b) for $move(rob_1, r_2)$
  rooms $r_1$ and $r_i$ are next to each other. Since our definition
  of $next\_to^*$ is such that the robot can always move to a
  neighboring cell and every two cells in rooms $r_1$ and $r_2$ are
  connected by paths which do not leave these rooms, clause 3 of
  Definition~\ref{def:weak-refinement} is satisfied. Thus, $\tau_{L,
    nobs}$ in our running example is a weak refinement of $\tau_H$.
\end{proof}

%%%%%-----------------------------------------------------------------------------
%%%%%-----------------------------------------------------------------------------
\section{Proof of Proposition~\ref{prop:strong-refinement}}
\label{sec:appendix-prop-strong-refine}
In this section, we prove Proposition~\ref{prop:strong-refinement},
which states that:

\medskip
\noindent {\em Let $\mathcal{D}_H$ and $\mathcal{D}_{L}$ be coarse and
  fine resolution system descriptions from our running example.  Then
  $\tau_L$ is a strong refinement of $\tau_H$.  }

\medskip
\noindent
\begin{proof}[Proof of Proposition~\ref{prop:strong-refinement}]

  For readability, we repeat the
  Definition~\ref{def:strong-refinement} of a strong refinement of a
  transition diagram. A transition diagram $\tau_L$ over $\Sigma_L$ is
  called a \emph{strong refinement} of $\tau_H$ if:
  \begin{enumerate}
  \item For every state $\sigma^\diamond$ of $\tau_L$, the collection
    $\sigma^\diamond |_{\Sigma_H}$ of atoms of $\sigma^\diamond$
    formed by symbols from $\Sigma_H$ is a state of $\tau_H$.
  
  \item For every state $\sigma$ of $\tau_H$, there is a state
    $\sigma^\diamond$ of $\tau_L$ such that $\sigma^\diamond$ is an
    extension of $\sigma$.

  \item For every transition $T = \langle \sigma_1,a^H,\sigma_2
    \rangle$ of $\tau_H$, if $\sigma^{\diamond}_1$ is an extension of
    $\sigma_1$, then for every observable fluent $f$ such that
    $observable_f(rob_1, \bar{x}, y)\in \sigma_2$, there is a path $P$
    in $\tau_L$ from $\sigma^{\diamond}_1$ to an extension
    $\sigma^{\diamond}_2$ of $\sigma_2$ such that:
    \begin{itemize}
    \item $P$ is \emph{pertinent} to $T$, i.e., all states of $P$ are
      extensions of $\sigma_1$ or $\sigma_2$;
    \item actions of $P$ are concrete, i.e., directly executable by
      robots; and
    \item $observed_f(rob_1, x, y)= true \in \sigma^{\diamond}_2$ iff
      $(f(x)=y) \in \sigma^{\diamond}_2$, and $observed_f(rob_1, x,
      y)= false \in \sigma^{\diamond}_2$ iff $(f(x)=y_1) \in
      \sigma^{\diamond}_2$ and $y_1 \not= y$.
    \end{itemize}
  \end{enumerate}

  \medskip
  \noindent
  The first two clauses of Definition~\ref{def:strong-refinement}
  follow immediately from the following observations:
  \begin{itemize}
  \item The states of $\tau_{L, nobs}$ and $\tau_L$ differ only by the
    knowledge functions. This follows immediately from the definition
    of a state and an application of the splitting set theorem.

  \item Both conditions are satisfied by the states of $\tau_{L,
      nobs}$; this follows from
    Proposition~\ref{prop:weak-refinement}.
  \end{itemize}
  To prove the third clause of Definition~\ref{def:strong-refinement},
  % we need to introduce some terminology.  We say that a path $P$ of
  % $\tau_L$ is \emph{good} if it is pertinent to $T$ and all of its
  % actions are concrete.  To prove the last clause of the proposition
  consider a transition $T=\langle \sigma_1, a^H, \sigma_2\rangle \in
  \tau_H$. There are two fluents $loc$ and $in\_hand$ that are
  observable in $\tau_H$. We start with the case in which the
  observable fluent is of the form:
  \begin{align*}
    loc(th) = rm
  \end{align*}
  Based on the third condition of the third clause of the proposition:
  \begin{enumerate}[label=(\arabic*),ref=(\arabic*)] %4-1
  \item $observable_{loc}(rob_1, th, rm) \in \sigma_2$
    \label{eqn:prop4-1}
  \end{enumerate}
  Based on the definition of $observable_{loc}$ for our example, this
  can happen only if:
  \begin{enumerate}[label=(\arabic*),ref=(\arabic*),resume] %4-2
  \item $loc(rob_1)=rm  \in \sigma_2$
    \label{eqn:prop4-2}
  \end{enumerate}
  
  \noindent
  Let $\delta^{\diamond}_0$ be a state of $\tau_L$ containing
  $\sigma_2$.  Then:
  \begin{enumerate}[label=(\arabic*),ref=(\arabic*),resume] %4-3
  \item $loc(rob_1) = rm  \in \delta^{\diamond}_0$
    \label{eqn:prop4-3}
  \end{enumerate}
  The value of $loc(th)$ in $\delta^{\diamond}_0$ is determined by the
  bridge axiom in Statement~\ref{eqn:refine-bridge-specific}(a) and
  hence Statement~\ref{eqn:prop4-3} holds iff for some cell $c_1$ of
  $rm$:
  \begin{enumerate}[label=(\arabic*),ref=(\arabic*),resume] %4-4
  \item $loc^*(rob_1)=c_1 \in \delta^{\diamond}_0$
    \label{eqn:prop4-4}
  \end{enumerate}
  Since by the definition of strong refinement, $\tau_L$ is also a
  weak refinement of $\tau_H$, Proposition~\ref{prop:weak-refinement}
  implies that there is a path $P_1$ of concrete action pertinent to
  $T$ from extension $\sigma^{\diamond}_1$ of $\sigma_1$ to
  $\delta^{\diamond}_0$. 

  \medskip
  \noindent
  There can be two possible cases:
  $$(i)~~loc(th) = rm \in \sigma_2$$
  $$(ii)~~loc(th) = rm \not \in \sigma_2$$
  
  \medskip
  \noindent
  In case (i), an argument similar to the one described above shows
  that there is a state $\delta^{\diamond}_1$ of $\tau_L$ containing
  $\sigma_2$ such that for some cell $c_2$ of room $rm$:
  \begin{enumerate}[label=(\arabic*),ref=(\arabic*),resume] %4-5
  \item $loc^*(th)=c_2 \in \delta^{\diamond}_1$
    \label{eqn:prop4-5}
  \end{enumerate}
  Now, let $P_2$ be the shortest sequence of the robot's moves from
  cell $c_1$ to cell $c_2$. Let $\delta^{\diamond}_2$ be the last
  state of this path. If at $\delta^{\diamond}_0$ the robot was
  already holding the thing $th$, then $P_2$ is empty. If the robot is
  not holding $th$, the moves of the robot do not change the location
  of $th$. Hence, we have:
  \begin{enumerate}[label=(\arabic*),ref=(\arabic*),resume] %4-6,7
  \item $loc^*(thing)=c_2 \in \delta^{\diamond}_2$ \label{eqn:prop4-6}
  \item $loc^*(rob)=c_2 \in \delta^{\diamond}_2$ \label{eqn:prop4-7}
  \end{enumerate}
  Statements~\ref{eqn:prop4-6} and~\ref{eqn:prop4-7}, together with
  the definition of $can\_be\_observed_{loc^*}$ imply that:
  \begin{enumerate}[label=(\arabic*),ref=(\arabic*),resume] %4-8
  \item $can\_be\_observed_{loc^*}(rob,thing,c_2) \in
    \delta^{\diamond}_2$ \label{eqn:prop4-8}
  \end{enumerate}

  \noindent
  The robot can now execute the knowledge-producing action
  $test_{loc^*}(rob_1, th, c_2)$, which moves the system into the
  state $\sigma^{\diamond}_2$.  Since this action does not change the
  values of physical fluents, locations of the robot and the thing
  remain unchanged. Now $observed_{loc}(rob_1, th, rm)\in
  \sigma^{\diamond}_2$ follows from Statements~\ref{eqn:refine-test}
  and~\ref{eqn:refine-indirect-obs}(a).  Notice that actions in the
  path $P$ defined as the concatenation of $P_1$, $P_2$ and $\langle
  \delta^{\diamond}_2, test_{loc^*}(rob_1, th, c_2),
  \sigma^{\diamond}_2 \rangle$ are concrete and relevant to $T$, and
  satisfies the conditions of the third clause of
  Definition~\ref{def:strong-refinement}.

  \medskip
  \noindent 
  In case (ii), i.e., with $loc(th) = rm \not\in \sigma_2$, let $P_1$
  be as before (i.e., a path of concrete action relevant to $T$ from
  $\sigma^{\diamond}_1$ to $\delta^{\diamond}_0$), $c_1, \ldots, c_n,
  c_1$ be a sequence visiting all the cells of $rm$, and $P$ be the
  concatenation of $P_1$ and the path $P_2$ of the form $\langle
  \delta^{\diamond}_i,move(rob_1, c_{i+1}), test_{loc^*}(rob_1, th,
  c_{i+1}), \delta^{\diamond}_{i+1} \rangle$.

  \medskip
  \noindent 
  Since every thing is assumed to have a location, $th$ is in some
  room, say $rm_1$ different from $rm$.  Since $loc(th)$ in determined
  by the bridge axiom in Statement~\ref{eqn:refine-bridge-specific}(a)
  and no grid cell can belong to two different rooms, there is some
  cell $c$ different from $c_1, \ldots, c_n$ such that $loc(th) = c$.
  Note that initially $observed_{loc}(rob_1, th, rm)$ and
  $observed_{loc^*}(rob_1, th, c)$ are $undet$ for every $c$ in $rm$.
  Since the thing $th$ is not in any cell of $rm$,
  $test_{loc^*}(rob_1, th, c)$ will return \emph{false} for every
  $c\in rm$. This means that
  Statement~\ref{eqn:refine-indirect-obs}(b) is not applicable, and
  Statement~\ref{eqn:refine-indirect-obs}(c) implies that
  $may\_be\_true_{loc}(rob,thing,rm)$ holds only until the robot
  reaches location $c_1$ and performs $test_{loc^*}(rob_1, th, c_1)$.
  In the resulting state, $\sigma^{\diamond}_2$, there is no component
  $c$ of $rm$ in which $observed_{loc^*}(rob_1, th, c)$ is $undet$.
  The value of the defined fluent $may\_be\_true_{loc}(rob,thing,rm)$
  is therefore \emph{false} in $\sigma^{\diamond}_2$. Based on
  Statement~\ref{eqn:refine-indirect-obs}(d), we conclude that
  $\sigma^{\diamond}_2$ contains $observed_{loc}(rob_1, th,
  rm)=false$.  Hence, the concatenation of $P_1$ and $P_2$ satisfies
  the conditions of the third clause of
  Definition~\ref{def:strong-refinement}.

  \medskip
  \noindent
  To complete the proof of Proposition~\ref{prop:strong-refinement},
  it only remains to notice that the desired path $P$ (of concrete
  actions relevant to $T$) corresponding to the observation of a
  fluent $in\_hand(rob_1, th)$ consists of just one action that tests
  if $in\_hand(rob_1, th) = true$; testing of a single value is
  sufficient due to Statement~\ref{eqn:refine-indirect-obs}(e).
\end{proof}

%%%%%-----------------------------------------------------------------------------
%%%%%-----------------------------------------------------------------------------
\section{POMDP Construction Example}
\label{sec:appendix-pomdp}
In this section, we illustrate the construction of a POMDP
$\mathcal{P}_o(T)$ for a specific coarse-resolution transition $T$
that needs to be implemented as a sequence of concrete actions whose
effects are modeled probabilistically.

\begin{example2}\label{ex:pomdp-model}[Example of POMDP construction]\\
  {\rm Consider abstract action $a^H = grasp(rob_1, tb_1)$, with the
    robot and textbook in the $of\!\!fice$, in the context of
    Example~\ref{ex:logician-example}. The corresponding zoomed system
    description $\mathcal{D}_{LR}(T)$ is in Example~\ref{ex:zoom1}.
    For ease of explanation, assume the following description of the
    transition function, observation function, and reward
    specification---these values would typically be computed by the
    robot in the initial training phase
    (Section~\ref{sec:arch-random}):
    \begin{itemize}
    \item Any move from a cell to a neighboring cell succeeds with
      probability $0.85$. Since there are only two cells in this room,
      the robot remains in the same cell if $move$ does not succeed.
    \item The $grasp$ action succeeds with probability $0.95$;
      otherwise it fails.
    \item If the thing being searched for in a cell exists in the
      cell, $0.95$ is the probability of successfully finding it.
    \item All non-terminal actions have unit cost. A correct answer
      receives a large positive reward ($100$), whereas an incorrect
      answer receives a large negative reward ($-100$).
    \end{itemize}
    The elements of the corresponding POMDP are described (below) in
    the format of the approximate POMDP solver used in our
    experiments~\cite{Ong:ijrr2010}. As described in
    Section~\ref{sec:arch-pomdp-construct}, please note that:
    \begin{itemize}
    \item Executing a terminal action causes a transition to a
      terminal state.
    \item Actions that change the p-state do not provide any
      observations.
    \item Knowledge-producing actions do not change the p-state.
    \item In any matrix corresponding to the transition function or
      observation function, the row and column entries (e.g., p-states
      or observations) are assumed to be in the order in which they
      appear at the top of the file.
    \end{itemize}
  }
\end{example2}

%\medskip
\begin{verbatim}
discount: 0.99

values: reward

% States, actions and observations as enumerated lists
states: robot-0-object-0-inhand robot-1-object-1-inhand robot-0-object-0-not-inhand
        robot-0-object-1-not-inhand robot-1-object-0-not-inhand 
        robot-1-object-1-not-inhand absb

actions: move-0 move-1 grasp test-robot-0 test-robot-1 test-object-0 test-object-1 
         test-inhand finish

observations: robot-found robot-not-found object-found object-not-found 
              inhand not-inhand none

% Transition function format.
% T : action : S x S' -> [0, 1]
% Probability of transition from first element of S to that of S' is
% in the top left corner of each matrix
T: move-0
1       0       0       0       0       0       0
0.85    0.15    0       0       0       0       0
0       0       1       0       0       0       0
0       0       0       1       0       0       0
0       0       0.85    0       0.15    0       0
0       0       0       0.85    0       0.15    0
0       0       0       0       0       0       1

T: move-1
0.15    0.85    0       0       0       0       0
0       1       0       0       0       0       0
0       0       0.15    0       0.85    0       0
0       0       0       0.15    0       0.85    0
0       0       0       0       1       0       0
0       0       0       0       0       1       0
0       0       0       0       0       0       1

T: grasp
1       0       0       0       0       0       0
0       1       0       0       0       0       0
0.95    0       0.05    0       0       0       0
0       0       0       1       0       0       0
0       0       0       0       1       0       0
0       0.95    0       0       0       0.05    0
0       0       0       0       0       0       1

T: test-robot-0
identity

T: test-robot-1
identity

T: test-object-0
identity

T: test-object-1
identity

T: test-inhand
identity

T: finish
uniform


% Observation function format(s) 
% O : action : s_i : z_i -> [0, 1] (or)
%            : S x Z -> [0, 1]
% In each matrix, first row provides probability of each possible
% observation in the first p-state in S
O: move-0 : * : none 1

O: move-1 : * : none 1

O: grasp : * : none 1

O: test-robot-0
0.95    0.05    0       0       0       0       0	
0.05    0.95    0       0       0       0       0	
0.95    0.05    0       0       0       0       0	
0.95    0.05    0       0       0       0       0	
0.05    0.95    0       0       0       0       0	
0.05    0.95    0       0       0       0       0	
0       0       0       0       0       0       1	

O: test-robot-1
0.05    0.95    0       0       0       0       0	
0.95    0.05    0       0       0       0       0	
0.05    0.95    0       0       0       0       0	
0.05    0.95    0       0       0       0       0	
0.95    0.05    0       0       0       0       0	
0.95    0.05    0       0       0       0       0	
0       0       0       0       0       0       1	

O: test-object-0
0       0       0.95    0.05    0       0       0	
0       0       0.05    0.95    0       0       0	
0       0       0.95    0.05    0       0       0	
0       0       0.05    0.95    0       0       0	
0       0       0.95    0.05    0       0       0	
0       0       0.05    0.95    0       0       0	
0       0       0       0       0       0       1	

O: test-object-1
0       0       0.05    0.95    0       0       0	
0       0       0.95    0.05    0       0       0	
0       0       0.05    0.95    0       0       0	
0       0       0.95    0.05    0       0       0	
0       0       0.05    0.95    0       0       0	
0       0       0.95    0.05    0       0       0	
0       0       0       0       0       0       1	

O: test-inhand
0       0       0       0       0.95    0.05    0	
0       0       0       0       0.95    0.05    0	
0       0       0       0       0.05    0.95    0	
0       0       0       0       0.05    0.95    0	
0       0       0       0       0.05    0.95    0	
0       0       0       0       0.05    0.95    0	
0       0       0       0       0       0       1	

O: finish : * : none 1


% Reward function format 
% R : action : s_i : s_i' : real value
R: finish	: robot-0-object-0-inhand	 : * : -100
R: finish	: robot-1-object-1-inhand	 : * : 100
R: finish	: robot-0-object-0-not-inhand	 : * : -100
R: finish	: robot-0-object-1-not-inhand	 : * : -100
R: finish	: robot-1-object-0-not-inhand	 : * : -100
R: finish	: robot-1-object-1-not-inhand	 : * : -100
R: move-0	: * : * : -1
R: move-1	: * : * : -1
R: grasp	: * : * : -1
R: test-robot-0	: * : * : -1
R: test-robot-1	: * : * : -1
R: test-object-0: * : * : -1
R: test-object-1: * : * : -1
R: test-inhand	: * : * : -1
\end{verbatim}

%%%%%-----------------------------------------------------------------------------
%%%%%-----------------------------------------------------------------------------

%%%%%-----------------------------------------------------------------------------
%%%%%-----------------------------------------------------------------------------

\newpage

\small
\bibliographystyle{plain}
\bibliography{references.bib}

%%%%%-----------------------------------------------------------------------------
%%%%%-----------------------------------------------------------------------------

\end{document}